\documentclass[10pt,twocolumn,letterpaper]{article}

\usepackage[pagenumbers]{cvpr} 

\usepackage{graphicx}
\usepackage{amsmath}
\usepackage{amssymb}
\usepackage{booktabs}
\usepackage{tabularx}
\usepackage{float}
\usepackage{colortbl,dcolumn}
\usepackage{xcolor}
\usepackage{hyphenat}
\usepackage{multirow}

\usepackage[pagebackref,breaklinks,colorlinks]{hyperref}

\usepackage[capitalize]{cleveref}
\crefname{section}{Sec.}{Secs.}
\Crefname{section}{Section}{Sections}
\Crefname{table}{Table}{Tables}
\crefname{table}{Tab.}{Tabs.}

\def\cvprPaperID{8170} 
\def\confName{CVPR}
\def\confYear{2022}

\begin{document}

\title{Unsupervised Domain Adaptation: A Reality Check}

\author{Kevin Musgrave\\
Cornell Tech\\
\and
Serge Belongie\\
University of Copenhagen\\
\and
Ser-Nam Lim\\
Meta AI
}
\maketitle

\begin{abstract}
Interest in unsupervised domain adaptation (UDA) has surged in recent years, resulting in a plethora of new algorithms. However, as is often the case in fast-moving fields, baseline algorithms are not tested to the extent that they should be. Furthermore, little attention has been paid to validation methods, i.e. the methods for estimating the accuracy of a model in the absence of target domain labels. This is despite the fact that validation methods are a crucial component of any UDA train/val pipeline. In this paper, we show via large-scale experimentation that 1) in the oracle setting, the difference in accuracy between UDA algorithms is smaller than previously thought, 2) state-of-the-art validation methods are not well-correlated with accuracy, and 3) differences between UDA algorithms are dwarfed by the drop in accuracy caused by validation methods. 
\end{abstract}

\section{Domain Adaptation Overview}
Imagine the following scenario: you have a model that accurately classifies \textit{photos} of animals, but you need the model to work on \textit{drawings} as well. You have a collection of animal drawings, but no corresponding labels, so standard supervised training is not possible. Luckily, you can use unsupervised domain adaptation (UDA) to solve this problem. The goal of UDA is to adapt a model trained on labeled source data $S$, for use on unlabeled target data $T$. More precisely, the $i$th sample of dataset $S$ is 
$$s_i=(\texttt{S.images}[i], \text{ } \texttt{S.labels}[i])$$
and the $i$th sample of dataset $T$ is: 
$$t_i=\texttt{T.images}[i]$$

\noindent Applications of UDA include semantic segmentation \cite{toldo2020unsupervised}, object detection \cite{oza2021unsupervised}, and natural language processing \cite{ramponi-plank-2020-neural}. There are also other types of domain adaptation, including semi-supervised \cite{saito2019semi}, multi-source \cite{peng2019moment}, partial \cite{panareda2017open, Saito_2018_ECCV, cao2018partial}, universal \cite{you2019universal}, and source-free \cite{liang2020shot}. In this paper, we focus on UDA for image classification, because it is well-studied and often used as a foundation for other domain adaptation subfields.

\subsection{Common themes in UDA}
In this section we provide a brief overview of ideas commonly used in UDA. See Table \ref{uda_algorithms_summary_table} for details of specific algorithms.
\begin{itemize}
    \item \textbf{Adversarial} methods use a GAN where the generator outputs feature vectors. The discriminator's goal is to correctly classify features as coming from the source or target domain, while the generator tries to minimize the discriminator's accuracy.
    \item \textbf{Feature distance losses} encourage source and target features to have similar distributions.
    \item \textbf{Maximum classifier discrepancy} \cite{saito2018maximum} methods use a generator and multiple classifiers in an adversarial setup. The classifiers' goal is to maximize the difference between their prediction vectors (i.e. after softmax) for the target domain data, while the generator's goal is to minimize this discrepancy.
    \item \textbf{Information maximization} methods use the entropy or mutual information of prediction vectors. 
    \item \textbf{SVD losses} apply singular value decomposition to the source and/or target features.
    \item \textbf{Image generation} methods use a decoder model to generate source/target -like images from feature vectors, usually as part of of an adversarial method. 
    \item \textbf{Pseudo labeling} methods generate labels for the unlabeled target domain data, to transform the problem from unsupervised to supervised. This is also known as self-supervised learning.
    \item \textbf{Mixup augmentations} create training data and features that are a blend between source and target domains.
\end{itemize}

\begin{table}[H]
\begin{tabularx}{\columnwidth}{>{\hsize=.6\hsize}X|>{\hsize=1.4\hsize}X}
 \toprule
\multicolumn{1}{c|}{\textbf{Algorithm}} & \multicolumn{1}{c}{\textbf{Highlight}} \\
\arrayrulecolor{gray} \midrule
\multicolumn{2}{c}{\textbf{Adversarial}} \\
\hline
DANN \cite{ganin2016domain} & Gradient reversal layer \\
\hline
DC \cite{tzeng2015simultaneous} & Uniform distribution loss  \\
\hline
ADDA \cite{tzeng2017adversarial} & Frozen source model \\
\hline
CDAN \cite{long2017conditional} & Randomized dot product for combining multiple features \\
\hline
VADA \cite{shu2018dirt} & Virtual adversarial training \\
\midrule
\multicolumn{2}{c}{\textbf{Feature distance losses}} \\
\hline
 MMD \cite{long2015learning} & Maximum mean discrepancy \\
 \hline
 CORAL \cite{sun2016return} & Covariance matrix alignment \\
 \hline
 JMMD \cite{long2017deep} & Joint MMD on multiple features \\
\midrule
\multicolumn{2}{c}{\textbf{Maximum classifier discrepancy}} \\
\hline
MCD \cite{saito2018maximum} & Discrepancy = L1 distance  \\
\hline
SWD \cite{lee2019sliced} & Discrepancy = sliced wasserstein \\
\hline
STAR \cite{lu2020stochastic} & Stochastic classifier layer \\
\midrule
\multicolumn{2}{c}{\textbf{Information maximization}} \\
\hline
ITL \cite{ICML2012Shi_566} & Maximize info of class predictions, minimize info of domain predictions. \\
\hline
MCC \cite{jin2020minimum} & Minimize class confusion via class correlations and entropy weighting \\
\hline
SENTRY \cite{prabhu2021sentry} & Min or max entropy, based on pseudo label + augmentation consistency \\
\midrule
\multicolumn{2}{c}{\textbf{SVD losses}} \\
\hline
BSP \cite{chen2019transferability} & Minimize singular values of features \\
\hline
BNM \cite{cui2020towards} & Max the sum of SVs of predictions \\
\midrule
\multicolumn{2}{c}{\textbf{Image generation}} \\
\hline
DRCN \cite{ghifary2016deep} & Reconstruct target images \\
\hline
GTA \cite{sankaranarayanan2018generate} & Generate source-like images from both source and target features \\
\midrule
\multicolumn{2}{c}{\textbf{Pseudo labeling}} \\
\hline
ATDA \cite{saito2017asymmetric} & Two source classifiers that create pseudo labels for the target classifier \\
\hline
ATDOC \cite{liang2021domain} & Pseudo labels from soft k-NN labels \\
\midrule
\multicolumn{2}{c}{\textbf{Mixup augmentations}} \\
\hline
DM-ADA \cite{xu2020adversarial} & Soft domain labels derived from image and feature domain mixup \\
\hline
DMRL \cite{wu2020dual} & Mixup using domain and class labels \\
\midrule
\multicolumn{2}{c}{\textbf{Other}} \\
\hline
RTN \cite{long2016unsupervised} & Residual connection between source and target logits \\
\hline
AFN \cite{xu2019larger} & Increase the L2 norm of features\\
\hline
DSBN \cite{chang2019domain} & Separate batchnorm layers for source and target domains \\
\hline
SymNets \cite{zhang2019domain} & Various operations on the concatenation of source and target predictions \\ 
\hline
GVB \cite{cui2020gradually} & Minimize L1 norm of bridge layers \\
\arrayrulecolor{black}\bottomrule
\end{tabularx}
\caption{Highlights of a selection of UDA algorithms}
\label{uda_algorithms_summary_table}
\end{table}

\subsection{Validation methods in UDA}\label{section_validation_methods_in_uda}
The assumption of UDA is that there are no target domain labels available, hence the name \textit{unsupervised} domain adaptation. This raises the question of how to evaluate models for the purpose of selecting algorithms and checkpoints, and tuning hyperparameters. Without labels, we cannot compute the accuracy of our model as we normally would. One potential workaround is to manually label a few of the target samples, and then use just those labeled samples to compute accuracy. However, if any labeled target data is available, we should use that data to train our model, because some labeled data is better than none. But now we are entering the realm of semi-supervised domain adaptation. To be unsupervised, we have to assume there are zero target labels available. Thus, the best we can do is to use methods that serve as a proxy to target domain accuracy. This subject has gotten little attention, so there are only a few methods that have been proposed in the literature:
\begin{itemize}
    \item \textbf{Reverse validation} \cite{reverseValidation, ganin2016domain} consists of two steps. First it trains a model via UDA on $S$ and $T$, and uses this model to create pseudo labels for $T$. Next, it trains a reverse model via UDA on $T$ and $S$, where $T$ is the pseudo labeled target data, and $S$ is the ``unlabeled" source data. The final score is the accuracy of the reverse model on $S$. One disadvantage of this approach is that it trains two models, doubling the required training time, but still producing only a single usable model.
    \item \textbf{Deep embedded validation (DEV)} \cite{pmlr-v97-you19a} computes the classification loss for every source validation sample, and weights each loss by the probability that the sample belongs to the target domain. (The probability comes from a domain classifier trained on source and target data.) The final score is obtained using the control variates method. One practical issue with DEV is that its scores are unbounded. This is because part of the calculation uses \texttt{1/var(weights)}, so if the domain classifier creates weights with small or zero variance, the score will be very large or NaN.  
    \item \textbf{Ensemble-based model selection (EMS)} \cite{robbiano2021adversarial} uses a linear regressor trained on 5 signals: target entropy, target diversity, Silhoutte \& Calinski-Harabasz scores on the target features, source accuracy, and time-consistent pseudo-labels. EMS differs from other methods because it requires a dataset of \{signal, ground truth accuracy\} pairs to train the regressor. These pairs have to be collected by training a model on a domain adaptation task that has labeled target data. After collecting the pairs and training the regressor, we still would not know if the regressor is accurate at predicting ground truth accuracy on our actual UDA task.
    \item \textbf{Soft neighborhood density (SND)} \cite{saito2021tune} computes the cosine similarity between all target features, converts each row of the similarity matrix into probabilities via temperature-scaled softmax, then returns the average entropy of the rows. High entropy means that each feature is close to many other features, which can indicate a well-clustered feature space. The caveat of SND is that it assumes the model has not mapped all target features into a single cluster. A single cluster would result in a high SND score, but low accuracy.
\end{itemize}

\noindent In addition to these real validation methods (a.k.a ``validators"), there is also the ``oracle'' method, which requires access to the ground truth target labels. Of course this cannot be used in reality, but it can be used in research experiments to find an algorithm's upperbound accuracy.

\section{Paper Meta Analysis}\label{section_meta_analysis}
To better understand the state of UDA research, we looked at 49 papers accepted at top conferences (CVPR, ECCV, ICCV, ICLR, ICML, NeurIPS, and AAAI) from 2015-2021. Our main goals were 1) to see how papers present the performance gap between state of the art (SOTA) and baseline results and 2) to see what validation methods are used. 

\subsection{SOTA-baseline performance gaps}

For each paper, we checked the results tables (if available) for Office31 \cite{saenko2010adapting} and OfficeHome \cite{venkateswara2017deep}, as they are among the most widely used datasets. Then for each transfer task, we compared the best performing algorithm with the two most commonly reported baselines: 1) ResNet50, which represents an ImageNet pretrained model that is finetuned on the source dataset (a.k.a source-only model), and 2) DANN, which is one of the seminal deep domain adaptation algorithms. Table \ref{meta_analysis_improvement_summary} summarizes our findings.

\begin{table}
\centering
\begin{tabular}{lrrrr}
\toprule
 & \multicolumn{2}{c}{Office31} & \multicolumn{2}{c}{OfficeHome} \\
{Year} & {Source-only} & {DANN} & {Source-only} & {DANN} \\
\midrule
2016 & - & 2.2 & - & - \\
2017 & 12.5 & 1.2 & - & 4.0 \\
2018 & 23.4 & 8.5 & 28.1 & 11.5 \\
2019 & 25.3 & 12.4 & 29.3 & 15.4 \\
2020 & 23.9 & 14.1 & 31.5 & 17.2 \\
2021 & 26.5 & 15.7 & 32.5 & 20.3 \\
\bottomrule
\end{tabular}
\caption{The largest average SOTA-baseline performance gap per year. For example, the 2021 OfficeHome/DANN value of 20.3 is the gap on the Product$\rightarrow$Art task, which is the task with the largest average SOTA-DANN gap for that year. Performance gap is measured as the absolute difference in accuracy.}
\label{meta_analysis_improvement_summary}
\end{table}

\subsection{Validation methods}
To determine what validation methods are used, we looked at both the papers and their official code repositories (repos) if available. Table \ref{meta_analysis_code_validation_methods} shows that most repos use the oracle method, regardless of what validator (if any) is mentioned in the corresponding paper.

\begin{table}
\centering
\resizebox{\columnwidth}{!}{\begin{tabular}{c c c || c} 
\toprule
Validator & \# Papers & \# Matches & \# Repos \\
\midrule
 full oracle & 0 & - & 30 \\
 subset oracle & 3 & 2 & 2 \\
 src accuracy & 0 & - & 1 \\
 src accuracy + loss & 2 & 0 & 0 \\
 consistency + oracle & 0 & - & 1 \\
 target entropy & 0 & - & 1 \\
 reverse validation & 2 & 0 & 0 \\
 IWCV \cite{sugiyama2007covariate} & 2 & 0 & 0 \\
 DEV & 2 & 0 & 0 \\
\bottomrule
\end{tabular}}
\caption{Validation methods in papers vs code. Out of 49 papers, 35 come with official repos. Of these 35 papers, 11 mention the validator that is used, and 2 use the same validator in both code and paper. 5 of the 6 papers that claim to use reverse validation, IWCV, or DEV, actually use oracle, and 1 uses target entropy.}
\label{meta_analysis_code_validation_methods}
\end{table}

\subsection{Discussion}
From the previous two sections, we can conclude that when using the oracle validator, the latest UDA algorithms can outperform baselines like DANN by over 20 points. However, there are two issues with this conclusion. First, it is uncommon for papers to re-implement baseline methods, so they may have been tested only a few times over the years. A re-implementation and a thorough hyperparameter search might yield surprising results. Second, the reported accuracies are obtained using the oracle validator. A non-oracle method will result in a non-optimal selection of models, hyperparameters, and algorithms, thus leading to a drop in accuracy. If the drop in accuracy is significant, it may render negligible the differences between algorithms in the oracle setting. In other words, the efficacy of the validator may be more important than the relatively subtle differences between algorithms.

With this in mind, we ran a large scale experiment to find out how UDA algorithms really stack up against each other, and how non-oracle validators affect accuracy.

\section{Experiment Methodology}\label{section_experiment_methodology}

In this section, we briefly describe our experiment setup\footnote{https://github.com/KevinMusgrave/pytorch-adapt}\textsuperscript{,}\footnote{ https://github.com/KevinMusgrave/powerful-benchmarker}. For more details about our methodology, please see the supplementary material.

\subsection{Datasets}
We ran experiments on 19 transfer tasks:
\begin{itemize}
    \item \textbf{MNIST}: 1 task between MNIST and MNISTM\cite{ganin2016domain}.
    \item \textbf{Office31} \cite{saenko2010adapting}: 6 tasks between 3 domains (Amazon, DSLR, Webcam).
    \item \textbf{OfficeHome} \cite{venkateswara2017deep}: 12 tasks between 4 domains (Art, Clipart, Product, Real).
\end{itemize}

\noindent MNIST and MNISTM are already divided into train/val splits, but Office31 and OfficeHome are not. So for each domain in these datasets, we created train/val splits with an 80/20 ratio per class (see Table \ref{split_usage}).

\begin{table}
\centering
\begin{tabular}{cccc}
\toprule
Step & Training & Validation & Testing \\
 \midrule
Finetuning & Source train & Source val & --- \\
\arrayrulecolor{gray}\hline
\begin{tabular}{@{}c@{}} UDA \end{tabular} &
 \begin{tabular}{@{}c@{}} Source train \\ Target train \end{tabular} &
 \begin{tabular}{@{}c@{}} Source train \\ Source val \\ Target train \end{tabular} &
 \begin{tabular}{@{}c@{}} Target val \end{tabular} \\
\arrayrulecolor{black}\bottomrule
\end{tabular}
\caption{How the four splits are used. The target train set is used during UDA validation because overfitting is unlikely to happen, due to the difficult unsupervised nature of the task. The source train/val sets may also be used, depending on the validator. The target val set is used for testing, and represents data that is seen for the first time during model deployment.}
\label{split_usage}
\end{table}

\subsection{Models}
For the MNIST$\rightarrow$MNISTM task, we used a LeNet-like model pretrained on MNIST as the trunk. For Office31 and OfficeHome, we used a ResNet50 \cite{He_2016_CVPR} pretrained \cite{rw2019timm} on ImageNet \cite{russakovsky2015imagenet}, and finetuned this model on every domain. For every task, we started each training run using the model finetuned on the source domain (i.e. the source-only model).

\subsection{Algorithms}
We evaluated algorithms from 20 papers\footnote{At the time of our experiments, the ATDOC paper had a typo. See https://github.com/KevinMusgrave/pytorch-adapt/issues/10.}, 12 of which are from 2018 or later. In addition to the DANN baseline mentioned in Section \ref{section_meta_analysis}, we also benchmarked minimum entropy (MinEnt) \cite{grandvalet2005semi}, information maximization (IM) \cite{ICML2012Shi_566}, and Information Theoretical Learning (ITL) \cite{ICML2012Shi_566}. All algorithms were implemented in PyTorch \cite{pytorchPaper}.

\subsection{Validation methods}
We ran experiments using four validation methods: oracle, IM, DEV \cite{pmlr-v97-you19a}, and SND \cite{saito2021tune}. The IM validator has the same definition as the IM UDA algorithm, but it uses the whole dataset rather than just a batch:

\begin{align}
IM = H(\frac{1}{N}\sum_{i=1}^{N}p_i) - \frac{1}{N}\sum_{i=1}^{N}H(p_i)
\end{align}

\noindent where $H$ is entropy, $p_i$ is the $i$th prediction vector, and $N$ is the size of the target dataset. 

IM has been used as part of UDA algorithms \cite{ICML2012Shi_566, liang2020shot}, but we are not aware of any paper that uses IM by itself as a general validation method. Robbiano et al \cite{robbiano2021adversarial} use IM as part of their EMS ensemble, and they also test the components (``diversity" and ``entropy") separately, but not the combination alone.

\subsection{Hyperparameter search}\label{section_hyperparameter_search}
In the oracle setting, we ran 100 steps of random hyperparameter search for each algorithm/task pair using Optuna \cite{optunaPaper}, and trained four additional models using the best settings. This full search was run using two different feature layers: the output of the trunk model (``FL0"), and the penultimate classifier layer (``FL6"). We also tried DANN with the softmax layer as features (``FL8").

For the non-oracle validators, we ran a similar hyperparameter search on 11 transfer tasks: MNIST, Office31, and four of the OfficeHome tasks (AP, CR, PA, and RC). We gathered 1.36 million datapoints, where a single datapoint consists of the validation score, source accuracy, and target accuracy collected from a validation step during training.

\section{Results}\label{section_results}

\subsection{Accuracy in the oracle setting}

Tables \ref{office31_oracle_results} and \ref{mnist_officehome_oracle_results} present results obtained using the oracle validator. Each table cell is the average of 5 runs using the best settings from all FL0 and FL6 experiments. Bold indicates the best value per column, and better values have a stronger green color. White cells have accuracy equal to or less than the source-only model. 

First note that our results are lower than typically reported. There are a few reasons for this:

\begin{itemize}
    \item Our training sets are 20\% smaller due to the creation of train/val splits. This has a big effect on Office31, which is already a small dataset.
    \item Our results are computed on the target validation set, which is never seen during training (see Table \ref{split_usage}). In contrast, papers usually report accuracy on the target training set because no validation set exists.
    \item Our results use macro-averaged accuracy instead of micro-averaged. This combined with the choice of evaluation split can have a non-trivial effect on accuracy as shown in Table \ref{difference_caused_by_splits}. 
\end{itemize}

\begin{figure}
    \centering
    \includegraphics[width=\columnwidth]{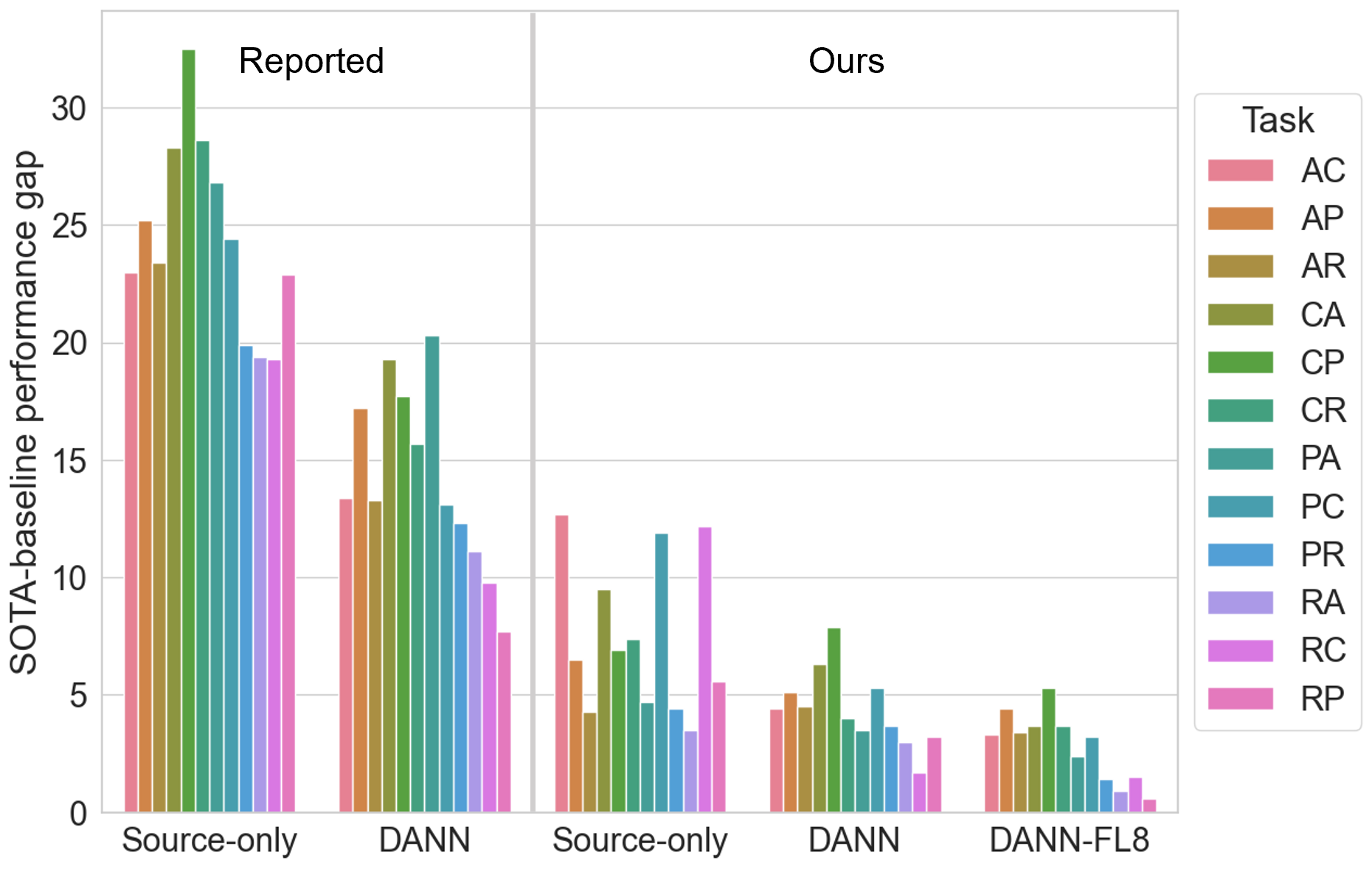}
    \caption{Performance gaps between SOTA and baseline algorithms (source-only and DANN) on OfficeHome tasks. The reported numbers are the average from 2021 papers.}
    \label{reported_vs_our_diffs}
\end{figure}

\noindent Next, we summarize the main takeaways of these results:

\begin{itemize}
    \item The source-only model is a strong baseline for Office31 and OfficeHome. In fact, there are many cases where UDA degrades performance, as indicated by the white table cells.
    \item MinEnt, IM, ITL, and DANN are strong UDA baselines for Office31 and OfficeHome, often outperforming more complicated methods like MCD, CDAN, VADA, SymNets, and ATDOC.
    \item The SOTA-baseline performance gap is much smaller than typically reported (Figure \ref{reported_vs_our_diffs} and Table \ref{reported_vs_our_diffs_table}).
    \item Some methods like DANN perform well on all three datasets. However, other methods perform poorly on MNIST, while scoring very highly on Office31 and OfficeHome, and vice versa. For example, MCC and BNM perform poorly on MNIST, but are the best on Office31 and OfficeHome. Likewise, STAR is among the best on MNIST, but among the lowest on OfficeHome.
\end{itemize}
\begin{table}
\centering
\begin{tabular}{ccc}
\toprule
 & Office31 & OfficeHome \\
 \midrule
Train Micro & 86.7 & 67.9 \\
Train Macro & 87.2 & 66.7 \\
Val Micro & 85.4 & 67.5 \\
Val Macro & 85.7 & 66.5 \\
\arrayrulecolor{black}\bottomrule
\end{tabular}
\caption{The accuracy on train/val splits, using micro and macro averaged accuracy. The values shown are the average of averages across transfer tasks, of all methods that outperform the source-only model. For example, the OfficeHome Val Macro number is the average of all green cells in the Avg column of Table \ref{mnist_officehome_oracle_results}.}
\label{difference_caused_by_splits}
\end{table}

\def\officethirtyoneAD#1{\ifdim#1pt>91.6pt\cellcolor{lime!100}\else\ifdim#1pt>90.1pt\cellcolor{lime!90}\else\ifdim#1pt>88.7pt\cellcolor{lime!80}\else\ifdim#1pt>87.2pt\cellcolor{lime!70}\else\ifdim#1pt>85.7pt\cellcolor{lime!60}\else\ifdim#1pt>84.2pt\cellcolor{lime!50}\else\ifdim#1pt>82.7pt\cellcolor{lime!40}\else\ifdim#1pt>81.3pt\cellcolor{lime!30}\else\ifdim#1pt>79.8pt\cellcolor{lime!20}\else\ifdim#1pt>78.3pt\cellcolor{lime!10}\else\cellcolor{lime!0}\fi\fi\fi\fi\fi\fi\fi\fi\fi\fi#1}

\def\officethirtyoneAW#1{\ifdim#1pt>92.2pt\cellcolor{lime!100}\else\ifdim#1pt>90.5pt\cellcolor{lime!90}\else\ifdim#1pt>88.9pt\cellcolor{lime!80}\else\ifdim#1pt>87.2pt\cellcolor{lime!70}\else\ifdim#1pt>85.6pt\cellcolor{lime!60}\else\ifdim#1pt>84.0pt\cellcolor{lime!50}\else\ifdim#1pt>82.3pt\cellcolor{lime!40}\else\ifdim#1pt>80.7pt\cellcolor{lime!30}\else\ifdim#1pt>79.0pt\cellcolor{lime!20}\else\ifdim#1pt>77.4pt\cellcolor{lime!10}\else\cellcolor{lime!0}\fi\fi\fi\fi\fi\fi\fi\fi\fi\fi#1}

\def\officethirtyoneDA#1{\ifdim#1pt>74.0pt\cellcolor{lime!100}\else\ifdim#1pt>73.5pt\cellcolor{lime!90}\else\ifdim#1pt>72.9pt\cellcolor{lime!80}\else\ifdim#1pt>72.4pt\cellcolor{lime!70}\else\ifdim#1pt>71.9pt\cellcolor{lime!60}\else\ifdim#1pt>71.4pt\cellcolor{lime!50}\else\ifdim#1pt>70.9pt\cellcolor{lime!40}\else\ifdim#1pt>70.3pt\cellcolor{lime!30}\else\ifdim#1pt>69.8pt\cellcolor{lime!20}\else\ifdim#1pt>69.3pt\cellcolor{lime!10}\else\cellcolor{lime!0}\fi\fi\fi\fi\fi\fi\fi\fi\fi\fi#1}

\def\officethirtyoneDW#1{\ifdim#1pt>97.0pt\cellcolor{lime!100}\else\ifdim#1pt>96.3pt\cellcolor{lime!90}\else\ifdim#1pt>95.7pt\cellcolor{lime!80}\else\ifdim#1pt>95.1pt\cellcolor{lime!70}\else\ifdim#1pt>94.4pt\cellcolor{lime!60}\else\ifdim#1pt>93.8pt\cellcolor{lime!50}\else\ifdim#1pt>93.2pt\cellcolor{lime!40}\else\ifdim#1pt>92.6pt\cellcolor{lime!30}\else\ifdim#1pt>91.9pt\cellcolor{lime!20}\else\ifdim#1pt>91.3pt\cellcolor{lime!10}\else\cellcolor{lime!0}\fi\fi\fi\fi\fi\fi\fi\fi\fi\fi#1}

\def\officethirtyoneWA#1{\ifdim#1pt>75.8pt\cellcolor{lime!100}\else\ifdim#1pt>75.5pt\cellcolor{lime!90}\else\ifdim#1pt>75.2pt\cellcolor{lime!80}\else\ifdim#1pt>74.9pt\cellcolor{lime!70}\else\ifdim#1pt>74.7pt\cellcolor{lime!60}\else\ifdim#1pt>74.4pt\cellcolor{lime!50}\else\ifdim#1pt>74.1pt\cellcolor{lime!40}\else\ifdim#1pt>73.8pt\cellcolor{lime!30}\else\ifdim#1pt>73.5pt\cellcolor{lime!20}\else\ifdim#1pt>73.2pt\cellcolor{lime!10}\else\cellcolor{lime!0}\fi\fi\fi\fi\fi\fi\fi\fi\fi\fi#1}

\def\officethirtyoneWD#1{\ifdim#1pt>99.6pt\cellcolor{lime!100}\else\ifdim#1pt>99.5pt\cellcolor{lime!90}\else\ifdim#1pt>99.3pt\cellcolor{lime!80}\else\ifdim#1pt>99.1pt\cellcolor{lime!70}\else\ifdim#1pt>98.9pt\cellcolor{lime!60}\else\ifdim#1pt>98.8pt\cellcolor{lime!50}\else\ifdim#1pt>98.6pt\cellcolor{lime!40}\else\ifdim#1pt>98.4pt\cellcolor{lime!30}\else\ifdim#1pt>98.3pt\cellcolor{lime!20}\else\ifdim#1pt>98.1pt\cellcolor{lime!10}\else\cellcolor{lime!0}\fi\fi\fi\fi\fi\fi\fi\fi\fi\fi#1}

\def\officethirtyoneAvg#1{\ifdim#1pt>88.0pt\cellcolor{lime!100}\else\ifdim#1pt>87.2pt\cellcolor{lime!90}\else\ifdim#1pt>86.5pt\cellcolor{lime!80}\else\ifdim#1pt>85.7pt\cellcolor{lime!70}\else\ifdim#1pt>85.0pt\cellcolor{lime!60}\else\ifdim#1pt>84.3pt\cellcolor{lime!50}\else\ifdim#1pt>83.5pt\cellcolor{lime!40}\else\ifdim#1pt>82.8pt\cellcolor{lime!30}\else\ifdim#1pt>82.0pt\cellcolor{lime!20}\else\ifdim#1pt>81.3pt\cellcolor{lime!10}\else\cellcolor{lime!0}\fi\fi\fi\fi\fi\fi\fi\fi\fi\fi#1}

\begin{table}
\centering
\resizebox{\columnwidth}{!}{\begin{tabular}{lllllll|l}
\toprule
{} & {AD} & {AW} & {DA} & {DW} & {WA} & {WD} & {Avg} \\
\midrule
Source-only & \officethirtyoneAD{78.3} & \officethirtyoneAW{77.4} & \officethirtyoneDA{69.3} & \officethirtyoneDW{91.3} & \officethirtyoneWA{73.2} & \officethirtyoneWD{98.1} & \officethirtyoneAvg{81.3} \\
ADDA & \officethirtyoneAD{71.0} & \officethirtyoneAW{73.7} & \officethirtyoneDA{64.5} & \officethirtyoneDW{89.1} & \officethirtyoneWA{65.5} & \officethirtyoneWD{93.2} & \officethirtyoneAvg{76.2} \\
AFN & \officethirtyoneAD{88.6} & \officethirtyoneAW{85.8} & \officethirtyoneDA{69.6} & \officethirtyoneDW{96.8} & \officethirtyoneWA{69.6} & \officethirtyoneWD{99.4} & \officethirtyoneAvg{85.0} \\
AFN-DANN & \officethirtyoneAD{87.7} & \officethirtyoneAW{93.4} & \officethirtyoneDA{70.7} & \officethirtyoneDW{96.5} & \officethirtyoneWA{72.8} & \officethirtyoneWD{99.6} & \officethirtyoneAvg{86.8} \\
ATDOC & \officethirtyoneAD{85.8} & \officethirtyoneAW{84.0} & \officethirtyoneDA{73.3} & \officethirtyoneDW{95.0} & \officethirtyoneWA{72.0} & \officethirtyoneWD{99.1} & \officethirtyoneAvg{84.9} \\
ATDOC-DANN & \officethirtyoneAD{85.9} & \officethirtyoneAW{91.5} & \textbf{\officethirtyoneDA{74.5}} & \officethirtyoneDW{96.6} & \officethirtyoneWA{73.8} & \officethirtyoneWD{98.7} & \officethirtyoneAvg{86.8} \\
BNM & \officethirtyoneAD{86.7} & \officethirtyoneAW{91.2} & \officethirtyoneDA{73.3} & \officethirtyoneDW{97.1} & \officethirtyoneWA{75.6} & \officethirtyoneWD{98.9} & \officethirtyoneAvg{87.1} \\
BNM-DANN & \officethirtyoneAD{88.7} & \officethirtyoneAW{91.4} & \officethirtyoneDA{72.7} & \officethirtyoneDW{96.6} & \officethirtyoneWA{75.5} & \officethirtyoneWD{99.6} & \officethirtyoneAvg{87.4} \\
BSP & \officethirtyoneAD{81.3} & \officethirtyoneAW{78.2} & \officethirtyoneDA{70.0} & \officethirtyoneDW{96.2} & \officethirtyoneWA{69.7} & \textbf{\officethirtyoneWD{99.8}} & \officethirtyoneAvg{82.5} \\
BSP-DANN & \officethirtyoneAD{85.6} & \officethirtyoneAW{90.4} & \officethirtyoneDA{71.8} & \officethirtyoneDW{96.3} & \officethirtyoneWA{73.0} & \officethirtyoneWD{99.6} & \officethirtyoneAvg{86.1} \\
CDAN & \officethirtyoneAD{82.2} & \officethirtyoneAW{90.8} & \officethirtyoneDA{72.0} & \officethirtyoneDW{95.7} & \officethirtyoneWA{72.1} & \officethirtyoneWD{99.2} & \officethirtyoneAvg{85.3} \\
CORAL & \officethirtyoneAD{84.3} & \officethirtyoneAW{84.2} & \officethirtyoneDA{69.9} & \officethirtyoneDW{91.7} & \officethirtyoneWA{70.6} & \officethirtyoneWD{98.4} & \officethirtyoneAvg{83.2} \\
DANN & \officethirtyoneAD{87.5} & \officethirtyoneAW{91.7} & \officethirtyoneDA{71.8} & \officethirtyoneDW{96.3} & \officethirtyoneWA{73.5} & \officethirtyoneWD{99.4} & \officethirtyoneAvg{86.7} \\
DANN-FL8 & \officethirtyoneAD{85.1} & \officethirtyoneAW{91.1} & \officethirtyoneDA{72.5} & \officethirtyoneDW{96.7} & \officethirtyoneWA{74.0} & \officethirtyoneWD{99.6} & \officethirtyoneAvg{86.5} \\
DC & \officethirtyoneAD{82.7} & \officethirtyoneAW{87.3} & \officethirtyoneDA{71.4} & \officethirtyoneDW{95.6} & \officethirtyoneWA{71.0} & \officethirtyoneWD{99.4} & \officethirtyoneAvg{84.6} \\
GVB & \officethirtyoneAD{88.1} & \officethirtyoneAW{89.3} & \officethirtyoneDA{74.1} & \officethirtyoneDW{94.9} & \officethirtyoneWA{74.5} & \officethirtyoneWD{98.2} & \officethirtyoneAvg{86.5} \\
IM & \officethirtyoneAD{90.4} & \officethirtyoneAW{87.1} & \officethirtyoneDA{72.1} & \officethirtyoneDW{96.7} & \officethirtyoneWA{72.2} & \officethirtyoneWD{99.4} & \officethirtyoneAvg{86.3} \\
IM-DANN & \officethirtyoneAD{88.6} & \officethirtyoneAW{91.1} & \officethirtyoneDA{71.6} & \officethirtyoneDW{96.4} & \officethirtyoneWA{74.8} & \textbf{\officethirtyoneWD{99.8}} & \officethirtyoneAvg{87.1} \\
ITL & \officethirtyoneAD{89.4} & \officethirtyoneAW{88.8} & \officethirtyoneDA{72.7} & \officethirtyoneDW{96.5} & \officethirtyoneWA{72.7} & \officethirtyoneWD{99.1} & \officethirtyoneAvg{86.5} \\
JMMD & \officethirtyoneAD{86.2} & \officethirtyoneAW{87.8} & \officethirtyoneDA{70.8} & \officethirtyoneDW{96.9} & \officethirtyoneWA{71.7} & \textbf{\officethirtyoneWD{99.8}} & \officethirtyoneAvg{85.5} \\
MCC & \officethirtyoneAD{91.2} & \officethirtyoneAW{91.5} & \officethirtyoneDA{72.8} & \officethirtyoneDW{97.1} & \officethirtyoneWA{75.5} & \officethirtyoneWD{99.4} & \officethirtyoneAvg{87.9} \\
MCC-DANN & \textbf{\officethirtyoneAD{93.1}} & \textbf{\officethirtyoneAW{93.8}} & \officethirtyoneDA{73.2} & \officethirtyoneDW{96.7} & \textbf{\officethirtyoneWA{76.1}} & \officethirtyoneWD{99.4} & \textbf{\officethirtyoneAvg{88.7}} \\
MCD & \officethirtyoneAD{86.6} & \officethirtyoneAW{86.5} & \officethirtyoneDA{68.2} & \officethirtyoneDW{96.8} & \officethirtyoneWA{69.1} & \officethirtyoneWD{98.7} & \officethirtyoneAvg{84.3} \\
MMD & \officethirtyoneAD{85.8} & \officethirtyoneAW{86.0} & \officethirtyoneDA{71.1} & \officethirtyoneDW{96.1} & \officethirtyoneWA{71.7} & \officethirtyoneWD{99.6} & \officethirtyoneAvg{85.1} \\
MinEnt & \officethirtyoneAD{85.2} & \officethirtyoneAW{88.5} & \officethirtyoneDA{72.5} & \officethirtyoneDW{96.8} & \officethirtyoneWA{72.9} & \officethirtyoneWD{98.7} & \officethirtyoneAvg{85.8} \\
RTN & \officethirtyoneAD{85.7} & \officethirtyoneAW{87.0} & \officethirtyoneDA{72.0} & \textbf{\officethirtyoneDW{97.6}} & \officethirtyoneWA{72.1} & \officethirtyoneWD{98.8} & \officethirtyoneAvg{85.5} \\
STAR & \officethirtyoneAD{78.4} & \officethirtyoneAW{77.4} & \officethirtyoneDA{60.6} & \officethirtyoneDW{95.9} & \officethirtyoneWA{63.6} & \officethirtyoneWD{98.5} & \officethirtyoneAvg{79.1} \\
SWD & \officethirtyoneAD{80.9} & \officethirtyoneAW{79.0} & \officethirtyoneDA{68.9} & \officethirtyoneDW{96.4} & \officethirtyoneWA{68.3} & \officethirtyoneWD{97.9} & \officethirtyoneAvg{81.9} \\
SymNets & \officethirtyoneAD{83.4} & \officethirtyoneAW{84.8} & \officethirtyoneDA{64.5} & \officethirtyoneDW{95.8} & \officethirtyoneWA{70.4} & \officethirtyoneWD{99.6} & \officethirtyoneAvg{83.1} \\
VADA & \officethirtyoneAD{88.1} & \officethirtyoneAW{88.6} & \officethirtyoneDA{71.1} & \officethirtyoneDW{96.5} & \officethirtyoneWA{70.0} & \officethirtyoneWD{98.7} & \officethirtyoneAvg{85.5} \\
\bottomrule
\end{tabular}}
\caption{Accuracy on the Office31 transfer tasks.}
\label{office31_oracle_results}
\end{table}
\def\mnistofficehomeMM#1{\ifdim#1pt>92.1pt\cellcolor{lime!100}\else\ifdim#1pt>88.3pt\cellcolor{lime!90}\else\ifdim#1pt>84.4pt\cellcolor{lime!80}\else\ifdim#1pt>80.6pt\cellcolor{lime!70}\else\ifdim#1pt>76.8pt\cellcolor{lime!60}\else\ifdim#1pt>73.0pt\cellcolor{lime!50}\else\ifdim#1pt>69.2pt\cellcolor{lime!40}\else\ifdim#1pt>65.3pt\cellcolor{lime!30}\else\ifdim#1pt>61.5pt\cellcolor{lime!20}\else\ifdim#1pt>57.7pt\cellcolor{lime!10}\else\cellcolor{lime!0}\fi\fi\fi\fi\fi\fi\fi\fi\fi\fi#1}

\def\mnistofficehomeAC#1{\ifdim#1pt>54.7pt\cellcolor{lime!100}\else\ifdim#1pt>53.5pt\cellcolor{lime!90}\else\ifdim#1pt>52.2pt\cellcolor{lime!80}\else\ifdim#1pt>50.9pt\cellcolor{lime!70}\else\ifdim#1pt>49.6pt\cellcolor{lime!60}\else\ifdim#1pt>48.4pt\cellcolor{lime!50}\else\ifdim#1pt>47.1pt\cellcolor{lime!40}\else\ifdim#1pt>45.8pt\cellcolor{lime!30}\else\ifdim#1pt>44.6pt\cellcolor{lime!20}\else\ifdim#1pt>43.3pt\cellcolor{lime!10}\else\cellcolor{lime!0}\fi\fi\fi\fi\fi\fi\fi\fi\fi\fi#1}

\def\mnistofficehomeAP#1{\ifdim#1pt>74.9pt\cellcolor{lime!100}\else\ifdim#1pt>74.3pt\cellcolor{lime!90}\else\ifdim#1pt>73.6pt\cellcolor{lime!80}\else\ifdim#1pt>73.0pt\cellcolor{lime!70}\else\ifdim#1pt>72.3pt\cellcolor{lime!60}\else\ifdim#1pt>71.7pt\cellcolor{lime!50}\else\ifdim#1pt>71.0pt\cellcolor{lime!40}\else\ifdim#1pt>70.4pt\cellcolor{lime!30}\else\ifdim#1pt>69.8pt\cellcolor{lime!20}\else\ifdim#1pt>69.1pt\cellcolor{lime!10}\else\cellcolor{lime!0}\fi\fi\fi\fi\fi\fi\fi\fi\fi\fi#1}

\def\mnistofficehomeAR#1{\ifdim#1pt>79.4pt\cellcolor{lime!100}\else\ifdim#1pt>78.9pt\cellcolor{lime!90}\else\ifdim#1pt>78.5pt\cellcolor{lime!80}\else\ifdim#1pt>78.1pt\cellcolor{lime!70}\else\ifdim#1pt>77.7pt\cellcolor{lime!60}\else\ifdim#1pt>77.2pt\cellcolor{lime!50}\else\ifdim#1pt>76.8pt\cellcolor{lime!40}\else\ifdim#1pt>76.4pt\cellcolor{lime!30}\else\ifdim#1pt>75.9pt\cellcolor{lime!20}\else\ifdim#1pt>75.5pt\cellcolor{lime!10}\else\cellcolor{lime!0}\fi\fi\fi\fi\fi\fi\fi\fi\fi\fi#1}

\def\mnistofficehomeCA#1{\ifdim#1pt>65.6pt\cellcolor{lime!100}\else\ifdim#1pt>64.7pt\cellcolor{lime!90}\else\ifdim#1pt>63.8pt\cellcolor{lime!80}\else\ifdim#1pt>62.8pt\cellcolor{lime!70}\else\ifdim#1pt>61.8pt\cellcolor{lime!60}\else\ifdim#1pt>60.9pt\cellcolor{lime!50}\else\ifdim#1pt>60.0pt\cellcolor{lime!40}\else\ifdim#1pt>59.0pt\cellcolor{lime!30}\else\ifdim#1pt>58.0pt\cellcolor{lime!20}\else\ifdim#1pt>57.1pt\cellcolor{lime!10}\else\cellcolor{lime!0}\fi\fi\fi\fi\fi\fi\fi\fi\fi\fi#1}

\def\mnistofficehomeCP#1{\ifdim#1pt>74.1pt\cellcolor{lime!100}\else\ifdim#1pt>73.4pt\cellcolor{lime!90}\else\ifdim#1pt>72.7pt\cellcolor{lime!80}\else\ifdim#1pt>72.0pt\cellcolor{lime!70}\else\ifdim#1pt>71.3pt\cellcolor{lime!60}\else\ifdim#1pt>70.7pt\cellcolor{lime!50}\else\ifdim#1pt>70.0pt\cellcolor{lime!40}\else\ifdim#1pt>69.3pt\cellcolor{lime!30}\else\ifdim#1pt>68.6pt\cellcolor{lime!20}\else\ifdim#1pt>67.9pt\cellcolor{lime!10}\else\cellcolor{lime!0}\fi\fi\fi\fi\fi\fi\fi\fi\fi\fi#1}

\def\mnistofficehomeCR#1{\ifdim#1pt>74.2pt\cellcolor{lime!100}\else\ifdim#1pt>73.4pt\cellcolor{lime!90}\else\ifdim#1pt>72.7pt\cellcolor{lime!80}\else\ifdim#1pt>71.9pt\cellcolor{lime!70}\else\ifdim#1pt>71.2pt\cellcolor{lime!60}\else\ifdim#1pt>70.5pt\cellcolor{lime!50}\else\ifdim#1pt>69.7pt\cellcolor{lime!40}\else\ifdim#1pt>69.0pt\cellcolor{lime!30}\else\ifdim#1pt>68.2pt\cellcolor{lime!20}\else\ifdim#1pt>67.5pt\cellcolor{lime!10}\else\cellcolor{lime!0}\fi\fi\fi\fi\fi\fi\fi\fi\fi\fi#1}

\def\mnistofficehomePA#1{\ifdim#1pt>63.7pt\cellcolor{lime!100}\else\ifdim#1pt>63.3pt\cellcolor{lime!90}\else\ifdim#1pt>62.8pt\cellcolor{lime!80}\else\ifdim#1pt>62.3pt\cellcolor{lime!70}\else\ifdim#1pt>61.9pt\cellcolor{lime!60}\else\ifdim#1pt>61.4pt\cellcolor{lime!50}\else\ifdim#1pt>60.9pt\cellcolor{lime!40}\else\ifdim#1pt>60.4pt\cellcolor{lime!30}\else\ifdim#1pt>60.0pt\cellcolor{lime!20}\else\ifdim#1pt>59.5pt\cellcolor{lime!10}\else\cellcolor{lime!0}\fi\fi\fi\fi\fi\fi\fi\fi\fi\fi#1}

\def\mnistofficehomePC#1{\ifdim#1pt>52.4pt\cellcolor{lime!100}\else\ifdim#1pt>51.2pt\cellcolor{lime!90}\else\ifdim#1pt>50.0pt\cellcolor{lime!80}\else\ifdim#1pt>48.8pt\cellcolor{lime!70}\else\ifdim#1pt>47.7pt\cellcolor{lime!60}\else\ifdim#1pt>46.5pt\cellcolor{lime!50}\else\ifdim#1pt>45.3pt\cellcolor{lime!40}\else\ifdim#1pt>44.1pt\cellcolor{lime!30}\else\ifdim#1pt>42.9pt\cellcolor{lime!20}\else\ifdim#1pt>41.7pt\cellcolor{lime!10}\else\cellcolor{lime!0}\fi\fi\fi\fi\fi\fi\fi\fi\fi\fi#1}

\def\mnistofficehomePR#1{\ifdim#1pt>81.4pt\cellcolor{lime!100}\else\ifdim#1pt>80.9pt\cellcolor{lime!90}\else\ifdim#1pt>80.5pt\cellcolor{lime!80}\else\ifdim#1pt>80.0pt\cellcolor{lime!70}\else\ifdim#1pt>79.6pt\cellcolor{lime!60}\else\ifdim#1pt>79.2pt\cellcolor{lime!50}\else\ifdim#1pt>78.7pt\cellcolor{lime!40}\else\ifdim#1pt>78.3pt\cellcolor{lime!30}\else\ifdim#1pt>77.8pt\cellcolor{lime!20}\else\ifdim#1pt>77.4pt\cellcolor{lime!10}\else\cellcolor{lime!0}\fi\fi\fi\fi\fi\fi\fi\fi\fi\fi#1}

\def\mnistofficehomeRA#1{\ifdim#1pt>72.7pt\cellcolor{lime!100}\else\ifdim#1pt>72.3pt\cellcolor{lime!90}\else\ifdim#1pt>72.0pt\cellcolor{lime!80}\else\ifdim#1pt>71.6pt\cellcolor{lime!70}\else\ifdim#1pt>71.2pt\cellcolor{lime!60}\else\ifdim#1pt>70.9pt\cellcolor{lime!50}\else\ifdim#1pt>70.5pt\cellcolor{lime!40}\else\ifdim#1pt>70.2pt\cellcolor{lime!30}\else\ifdim#1pt>69.8pt\cellcolor{lime!20}\else\ifdim#1pt>69.5pt\cellcolor{lime!10}\else\cellcolor{lime!0}\fi\fi\fi\fi\fi\fi\fi\fi\fi\fi#1}

\def\mnistofficehomeRC#1{\ifdim#1pt>56.0pt\cellcolor{lime!100}\else\ifdim#1pt>54.8pt\cellcolor{lime!90}\else\ifdim#1pt>53.5pt\cellcolor{lime!80}\else\ifdim#1pt>52.3pt\cellcolor{lime!70}\else\ifdim#1pt>51.1pt\cellcolor{lime!60}\else\ifdim#1pt>49.9pt\cellcolor{lime!50}\else\ifdim#1pt>48.7pt\cellcolor{lime!40}\else\ifdim#1pt>47.4pt\cellcolor{lime!30}\else\ifdim#1pt>46.2pt\cellcolor{lime!20}\else\ifdim#1pt>45.0pt\cellcolor{lime!10}\else\cellcolor{lime!0}\fi\fi\fi\fi\fi\fi\fi\fi\fi\fi#1}

\def\mnistofficehomeRP#1{\ifdim#1pt>82.5pt\cellcolor{lime!100}\else\ifdim#1pt>82.0pt\cellcolor{lime!90}\else\ifdim#1pt>81.4pt\cellcolor{lime!80}\else\ifdim#1pt>80.9pt\cellcolor{lime!70}\else\ifdim#1pt>80.3pt\cellcolor{lime!60}\else\ifdim#1pt>79.7pt\cellcolor{lime!50}\else\ifdim#1pt>79.2pt\cellcolor{lime!40}\else\ifdim#1pt>78.6pt\cellcolor{lime!30}\else\ifdim#1pt>78.1pt\cellcolor{lime!20}\else\ifdim#1pt>77.5pt\cellcolor{lime!10}\else\cellcolor{lime!0}\fi\fi\fi\fi\fi\fi\fi\fi\fi\fi#1}

\def\mnistofficehomeAvg#1{\ifdim#1pt>69.1pt\cellcolor{lime!100}\else\ifdim#1pt>68.4pt\cellcolor{lime!90}\else\ifdim#1pt>67.6pt\cellcolor{lime!80}\else\ifdim#1pt>66.9pt\cellcolor{lime!70}\else\ifdim#1pt>66.2pt\cellcolor{lime!60}\else\ifdim#1pt>65.5pt\cellcolor{lime!50}\else\ifdim#1pt>64.8pt\cellcolor{lime!40}\else\ifdim#1pt>64.0pt\cellcolor{lime!30}\else\ifdim#1pt>63.3pt\cellcolor{lime!20}\else\ifdim#1pt>62.6pt\cellcolor{lime!10}\else\cellcolor{lime!0}\fi\fi\fi\fi\fi\fi\fi\fi\fi\fi#1}

\begin{table*}
\centering
\begin{tabular}{ll||llllllllllll|l}
\toprule
{} & {MM} & {AC} & {AP} & {AR} & {CA} & {CP} & {CR} & {PA} & {PC} & {PR} & {RA} & {RC} & {RP} & {Avg} \\
\midrule
Source-only & \mnistofficehomeMM{57.7} & \mnistofficehomeAC{43.3} & \mnistofficehomeAP{69.1} & \mnistofficehomeAR{75.5} & \mnistofficehomeCA{57.1} & \mnistofficehomeCP{67.9} & \mnistofficehomeCR{67.5} & \mnistofficehomePA{59.5} & \mnistofficehomePC{41.7} & \mnistofficehomePR{77.4} & \mnistofficehomeRA{69.5} & \mnistofficehomeRC{45.0} & \mnistofficehomeRP{77.5} & \mnistofficehomeAvg{62.6} \\
ADDA & \mnistofficehomeMM{84.9} & \mnistofficehomeAC{42.5} & \mnistofficehomeAP{64.9} & \mnistofficehomeAR{70.4} & \mnistofficehomeCA{56.8} & \mnistofficehomeCP{60.9} & \mnistofficehomeCR{65.0} & \mnistofficehomePA{56.7} & \mnistofficehomePC{38.5} & \mnistofficehomePR{74.1} & \mnistofficehomeRA{66.9} & \mnistofficehomeRC{45.6} & \mnistofficehomeRP{74.2} & \mnistofficehomeAvg{59.7} \\
AFN & \mnistofficehomeMM{60.9} & \mnistofficehomeAC{47.7} & \mnistofficehomeAP{69.5} & \mnistofficehomeAR{75.0} & \mnistofficehomeCA{60.4} & \mnistofficehomeCP{64.5} & \mnistofficehomeCR{69.3} & \mnistofficehomePA{58.6} & \mnistofficehomePC{42.5} & \mnistofficehomePR{78.0} & \mnistofficehomeRA{69.6} & \mnistofficehomeRC{49.7} & \mnistofficehomeRP{79.1} & \mnistofficehomeAvg{63.7} \\
AFN-DANN & \mnistofficehomeMM{93.6} & \mnistofficehomeAC{51.6} & \mnistofficehomeAP{70.5} & \mnistofficehomeAR{74.8} & \mnistofficehomeCA{62.3} & \mnistofficehomeCP{67.7} & \mnistofficehomeCR{71.4} & \mnistofficehomePA{60.2} & \mnistofficehomePC{47.5} & \mnistofficehomePR{78.6} & \mnistofficehomeRA{69.0} & \mnistofficehomeRC{55.1} & \mnistofficehomeRP{80.1} & \mnistofficehomeAvg{65.7} \\
ATDOC & \mnistofficehomeMM{66.8} & \mnistofficehomeAC{48.0} & \mnistofficehomeAP{73.0} & \mnistofficehomeAR{75.9} & \mnistofficehomeCA{62.5} & \mnistofficehomeCP{70.7} & \mnistofficehomeCR{74.0} & \mnistofficehomePA{61.7} & \mnistofficehomePC{44.9} & \mnistofficehomePR{79.2} & \mnistofficehomeRA{69.4} & \mnistofficehomeRC{50.3} & \mnistofficehomeRP{80.5} & \mnistofficehomeAvg{65.8} \\
ATDOC-DANN & \mnistofficehomeMM{86.8} & \mnistofficehomeAC{51.6} & \mnistofficehomeAP{73.7} & \mnistofficehomeAR{76.8} & \mnistofficehomeCA{63.4} & \mnistofficehomeCP{71.0} & \mnistofficehomeCR{73.2} & \mnistofficehomePA{60.8} & \mnistofficehomePC{46.4} & \mnistofficehomePR{77.5} & \mnistofficehomeRA{69.4} & \mnistofficehomeRC{54.2} & \mnistofficehomeRP{81.9} & \mnistofficehomeAvg{66.7} \\
BNM & \mnistofficehomeMM{63.0} & \mnistofficehomeAC{53.3} & \mnistofficehomeAP{74.9} & \mnistofficehomeAR{79.0} & \mnistofficehomeCA{65.7} & \mnistofficehomeCP{72.6} & \mnistofficehomeCR{74.8} & \mnistofficehomePA{62.5} & \mnistofficehomePC{50.2} & \mnistofficehomePR{80.5} & \mnistofficehomeRA{71.5} & \mnistofficehomeRC{55.7} & \mnistofficehomeRP{82.3} & \mnistofficehomeAvg{68.6} \\
BNM-DANN & \mnistofficehomeMM{94.5} & \mnistofficehomeAC{53.9} & \mnistofficehomeAP{74.9} & \mnistofficehomeAR{78.9} & \mnistofficehomeCA{64.8} & \mnistofficehomeCP{71.9} & \mnistofficehomeCR{74.3} & \mnistofficehomePA{61.7} & \mnistofficehomePC{51.1} & \mnistofficehomePR{79.7} & \mnistofficehomeRA{71.1} & \mnistofficehomeRC{56.4} & \mnistofficehomeRP{81.5} & \mnistofficehomeAvg{68.3} \\
BSP & \mnistofficehomeMM{58.4} & \mnistofficehomeAC{44.6} & \mnistofficehomeAP{68.1} & \mnistofficehomeAR{74.6} & \mnistofficehomeCA{59.1} & \mnistofficehomeCP{63.4} & \mnistofficehomeCR{68.0} & \mnistofficehomePA{57.8} & \mnistofficehomePC{40.6} & \mnistofficehomePR{76.7} & \mnistofficehomeRA{68.4} & \mnistofficehomeRC{46.6} & \mnistofficehomeRP{77.4} & \mnistofficehomeAvg{62.1} \\
BSP-DANN & \textbf{\mnistofficehomeMM{95.9}} & \mnistofficehomeAC{51.6} & \mnistofficehomeAP{70.8} & \mnistofficehomeAR{75.0} & \mnistofficehomeCA{60.5} & \mnistofficehomeCP{66.4} & \mnistofficehomeCR{70.0} & \mnistofficehomePA{59.3} & \mnistofficehomePC{47.8} & \mnistofficehomePR{77.9} & \mnistofficehomeRA{69.9} & \mnistofficehomeRC{55.1} & \mnistofficehomeRP{79.5} & \mnistofficehomeAvg{65.3} \\
CDAN & \mnistofficehomeMM{88.1} & \mnistofficehomeAC{51.4} & \mnistofficehomeAP{71.0} & \mnistofficehomeAR{74.5} & \mnistofficehomeCA{60.2} & \mnistofficehomeCP{67.3} & \mnistofficehomeCR{71.0} & \mnistofficehomePA{59.2} & \mnistofficehomePC{49.9} & \mnistofficehomePR{80.1} & \mnistofficehomeRA{70.9} & \mnistofficehomeRC{55.8} & \mnistofficehomeRP{80.1} & \mnistofficehomeAvg{66.0} \\
CORAL & \mnistofficehomeMM{69.6} & \mnistofficehomeAC{47.1} & \mnistofficehomeAP{69.2} & \mnistofficehomeAR{74.9} & \mnistofficehomeCA{60.4} & \mnistofficehomeCP{64.1} & \mnistofficehomeCR{67.9} & \mnistofficehomePA{57.9} & \mnistofficehomePC{41.5} & \mnistofficehomePR{78.5} & \mnistofficehomeRA{69.2} & \mnistofficehomeRC{49.3} & \mnistofficehomeRP{79.0} & \mnistofficehomeAvg{63.2} \\
DANN & \mnistofficehomeMM{93.8} & \mnistofficehomeAC{51.6} & \mnistofficehomeAP{70.5} & \mnistofficehomeAR{75.3} & \mnistofficehomeCA{60.3} & \mnistofficehomeCP{66.9} & \mnistofficehomeCR{70.9} & \mnistofficehomePA{60.7} & \mnistofficehomePC{48.3} & \mnistofficehomePR{78.1} & \mnistofficehomeRA{70.0} & \mnistofficehomeRC{55.5} & \mnistofficehomeRP{79.9} & \mnistofficehomeAvg{65.7} \\
DANN-FL8 & \mnistofficehomeMM{69.1} & \mnistofficehomeAC{52.7} & \mnistofficehomeAP{71.2} & \mnistofficehomeAR{76.4} & \mnistofficehomeCA{62.9} & \mnistofficehomeCP{69.5} & \mnistofficehomeCR{71.2} & \mnistofficehomePA{61.8} & \mnistofficehomePC{50.4} & \mnistofficehomePR{80.4} & \mnistofficehomeRA{72.1} & \mnistofficehomeRC{55.7} & \mnistofficehomeRP{82.5} & \mnistofficehomeAvg{67.2} \\
DC & \mnistofficehomeMM{84.6} & \mnistofficehomeAC{48.8} & \mnistofficehomeAP{69.0} & \mnistofficehomeAR{74.3} & \mnistofficehomeCA{59.7} & \mnistofficehomeCP{64.5} & \mnistofficehomeCR{68.7} & \mnistofficehomePA{61.1} & \mnistofficehomePC{44.5} & \mnistofficehomePR{77.8} & \mnistofficehomeRA{68.2} & \mnistofficehomeRC{52.4} & \mnistofficehomeRP{78.5} & \mnistofficehomeAvg{63.9} \\
GVB & \mnistofficehomeMM{74.4} & \mnistofficehomeAC{52.6} & \mnistofficehomeAP{72.0} & \mnistofficehomeAR{75.3} & \mnistofficehomeCA{62.5} & \mnistofficehomeCP{69.6} & \mnistofficehomeCR{73.6} & \textbf{\mnistofficehomePA{64.2}} & \mnistofficehomePC{51.7} & \mnistofficehomePR{80.3} & \mnistofficehomeRA{71.9} & \mnistofficehomeRC{56.0} & \mnistofficehomeRP{82.4} & \mnistofficehomeAvg{67.7} \\
IM & \mnistofficehomeMM{60.7} & \mnistofficehomeAC{51.4} & \mnistofficehomeAP{73.9} & \mnistofficehomeAR{76.8} & \mnistofficehomeCA{63.3} & \mnistofficehomeCP{70.1} & \mnistofficehomeCR{71.7} & \mnistofficehomePA{62.5} & \mnistofficehomePC{49.0} & \mnistofficehomePR{79.9} & \mnistofficehomeRA{72.7} & \mnistofficehomeRC{52.5} & \mnistofficehomeRP{81.2} & \mnistofficehomeAvg{67.1} \\
IM-DANN & \mnistofficehomeMM{95.4} & \mnistofficehomeAC{53.2} & \mnistofficehomeAP{73.9} & \mnistofficehomeAR{76.6} & \mnistofficehomeCA{64.6} & \mnistofficehomeCP{71.0} & \mnistofficehomeCR{73.6} & \mnistofficehomePA{63.0} & \mnistofficehomePC{51.1} & \mnistofficehomePR{80.1} & \textbf{\mnistofficehomeRA{73.0}} & \mnistofficehomeRC{55.3} & \mnistofficehomeRP{82.4} & \mnistofficehomeAvg{68.1} \\
ITL & \mnistofficehomeMM{61.0} & \mnistofficehomeAC{52.5} & \mnistofficehomeAP{73.6} & \mnistofficehomeAR{75.8} & \mnistofficehomeCA{62.4} & \mnistofficehomeCP{69.7} & \mnistofficehomeCR{72.1} & \mnistofficehomePA{62.4} & \mnistofficehomePC{48.0} & \mnistofficehomePR{80.2} & \mnistofficehomeRA{72.3} & \mnistofficehomeRC{52.2} & \mnistofficehomeRP{81.6} & \mnistofficehomeAvg{66.9} \\
JMMD & \mnistofficehomeMM{64.8} & \mnistofficehomeAC{49.2} & \mnistofficehomeAP{71.1} & \mnistofficehomeAR{74.7} & \mnistofficehomeCA{60.4} & \mnistofficehomeCP{66.9} & \mnistofficehomeCR{69.6} & \mnistofficehomePA{59.8} & \mnistofficehomePC{44.0} & \mnistofficehomePR{78.5} & \mnistofficehomeRA{70.7} & \mnistofficehomeRC{51.3} & \mnistofficehomeRP{78.7} & \mnistofficehomeAvg{64.6} \\
MCC & \mnistofficehomeMM{63.1} & \textbf{\mnistofficehomeAC{56.0}} & \textbf{\mnistofficehomeAP{75.6}} & \textbf{\mnistofficehomeAR{79.8}} & \textbf{\mnistofficehomeCA{66.6}} & \textbf{\mnistofficehomeCP{74.8}} & \mnistofficehomeCR{74.8} & \mnistofficehomePA{63.4} & \textbf{\mnistofficehomePC{53.6}} & \textbf{\mnistofficehomePR{81.8}} & \mnistofficehomeRA{71.3} & \mnistofficehomeRC{56.6} & \textbf{\mnistofficehomeRP{83.1}} & \textbf{\mnistofficehomeAvg{69.8}} \\
MCC-DANN & \mnistofficehomeMM{94.3} & \mnistofficehomeAC{54.6} & \mnistofficehomeAP{75.3} & \mnistofficehomeAR{79.6} & \mnistofficehomeCA{66.5} & \mnistofficehomeCP{74.4} & \textbf{\mnistofficehomeCR{74.9}} & \mnistofficehomePA{62.8} & \mnistofficehomePC{53.2} & \textbf{\mnistofficehomePR{81.8}} & \mnistofficehomeRA{72.0} & \textbf{\mnistofficehomeRC{57.2}} & \mnistofficehomeRP{82.7} & \mnistofficehomeAvg{69.6} \\
MCD & \mnistofficehomeMM{94.3} & \mnistofficehomeAC{45.1} & \mnistofficehomeAP{67.5} & \mnistofficehomeAR{73.9} & \mnistofficehomeCA{58.8} & \mnistofficehomeCP{64.1} & \mnistofficehomeCR{67.1} & \mnistofficehomePA{58.2} & \mnistofficehomePC{39.4} & \mnistofficehomePR{77.7} & \mnistofficehomeRA{67.6} & \mnistofficehomeRC{45.2} & \mnistofficehomeRP{78.5} & \mnistofficehomeAvg{61.9} \\
MMD & \mnistofficehomeMM{72.4} & \mnistofficehomeAC{50.7} & \mnistofficehomeAP{70.6} & \mnistofficehomeAR{74.4} & \mnistofficehomeCA{61.1} & \mnistofficehomeCP{66.9} & \mnistofficehomeCR{70.3} & \mnistofficehomePA{60.5} & \mnistofficehomePC{45.2} & \mnistofficehomePR{78.6} & \mnistofficehomeRA{70.2} & \mnistofficehomeRC{52.0} & \mnistofficehomeRP{79.9} & \mnistofficehomeAvg{65.0} \\
MinEnt & \mnistofficehomeMM{56.4} & \mnistofficehomeAC{49.9} & \mnistofficehomeAP{72.9} & \mnistofficehomeAR{76.5} & \mnistofficehomeCA{61.3} & \mnistofficehomeCP{71.2} & \mnistofficehomeCR{73.0} & \mnistofficehomePA{62.0} & \mnistofficehomePC{47.9} & \mnistofficehomePR{80.2} & \mnistofficehomeRA{72.6} & \mnistofficehomeRC{51.7} & \mnistofficehomeRP{81.8} & \mnistofficehomeAvg{66.7} \\
RTN & \mnistofficehomeMM{58.6} & \mnistofficehomeAC{50.9} & \mnistofficehomeAP{72.5} & \mnistofficehomeAR{75.9} & \mnistofficehomeCA{62.0} & \mnistofficehomeCP{70.7} & \mnistofficehomeCR{72.3} & \mnistofficehomePA{62.2} & \mnistofficehomePC{46.7} & \mnistofficehomePR{80.2} & \mnistofficehomeRA{69.6} & \mnistofficehomeRC{53.3} & \mnistofficehomeRP{82.0} & \mnistofficehomeAvg{66.5} \\
STAR & \mnistofficehomeMM{95.0} & \mnistofficehomeAC{41.2} & \mnistofficehomeAP{65.9} & \mnistofficehomeAR{71.3} & \mnistofficehomeCA{53.0} & \mnistofficehomeCP{54.5} & \mnistofficehomeCR{61.6} & \mnistofficehomePA{52.0} & \mnistofficehomePC{32.1} & \mnistofficehomePR{68.0} & \mnistofficehomeRA{62.8} & \mnistofficehomeRC{40.9} & \mnistofficehomeRP{71.3} & \mnistofficehomeAvg{56.2} \\
SWD & \mnistofficehomeMM{80.2} & \mnistofficehomeAC{44.9} & \mnistofficehomeAP{66.9} & \mnistofficehomeAR{73.1} & \mnistofficehomeCA{58.7} & \mnistofficehomeCP{64.5} & \mnistofficehomeCR{68.0} & \mnistofficehomePA{58.5} & \mnistofficehomePC{41.9} & \mnistofficehomePR{77.0} & \mnistofficehomeRA{68.5} & \mnistofficehomeRC{47.0} & \mnistofficehomeRP{78.1} & \mnistofficehomeAvg{62.3} \\
SymNets & \mnistofficehomeMM{82.3} & \mnistofficehomeAC{35.2} & \mnistofficehomeAP{56.1} & \mnistofficehomeAR{64.4} & \mnistofficehomeCA{52.5} & \mnistofficehomeCP{46.7} & \mnistofficehomeCR{57.4} & \mnistofficehomePA{60.7} & \mnistofficehomePC{38.6} & \mnistofficehomePR{75.9} & \mnistofficehomeRA{66.9} & \mnistofficehomeRC{44.5} & \mnistofficehomeRP{78.6} & \mnistofficehomeAvg{56.5} \\
VADA & \mnistofficehomeMM{93.0} & \mnistofficehomeAC{45.1} & \mnistofficehomeAP{66.8} & \mnistofficehomeAR{73.8} & \mnistofficehomeCA{57.6} & \mnistofficehomeCP{63.8} & \mnistofficehomeCR{67.2} & \mnistofficehomePA{57.2} & \mnistofficehomePC{46.3} & \mnistofficehomePR{76.1} & \mnistofficehomeRA{65.0} & \mnistofficehomeRC{51.7} & \mnistofficehomeRP{75.6} & \mnistofficehomeAvg{62.2} \\
\bottomrule
\end{tabular}
\caption{Accuracy on the MNIST $\rightarrow$ MNISTM (MM) and OfficeHome transfer tasks. The Avg column is the OfficeHome average.}
\label{mnist_officehome_oracle_results}
\end{table*}

\begin{table}
\centering
\begin{tabular}{llrr}
\toprule
{} & Model & Office31 & OfficeHome \\
\midrule
\multirow{2}{*}{Reported} & Source-only & 26.5 & 32.5  \\
& DANN & 15.7 & 20.3 \\
\arrayrulecolor{gray}\hline
\multirow{3}{*}{Ours} & Source-only & 16.4 & 12.7 \\
& DANN & 5.6 & 7.9 \\
& DANN-FL8 & 8.0 & 5.3 \\
\arrayrulecolor{black}\bottomrule
\end{tabular}
\caption{Average reported performance gap in 2021 papers vs ours. Each number corresponds with the transfer task with the largest performance gap.}
\label{reported_vs_our_diffs_table}
\end{table}

\subsection{Impact of validation methods on accuracy}\label{section_impact_of_validation_methods}
\begin{figure}
     \centering
     \begin{subfigure}[b]{\columnwidth}
         \centering
         \includegraphics[width=\columnwidth]{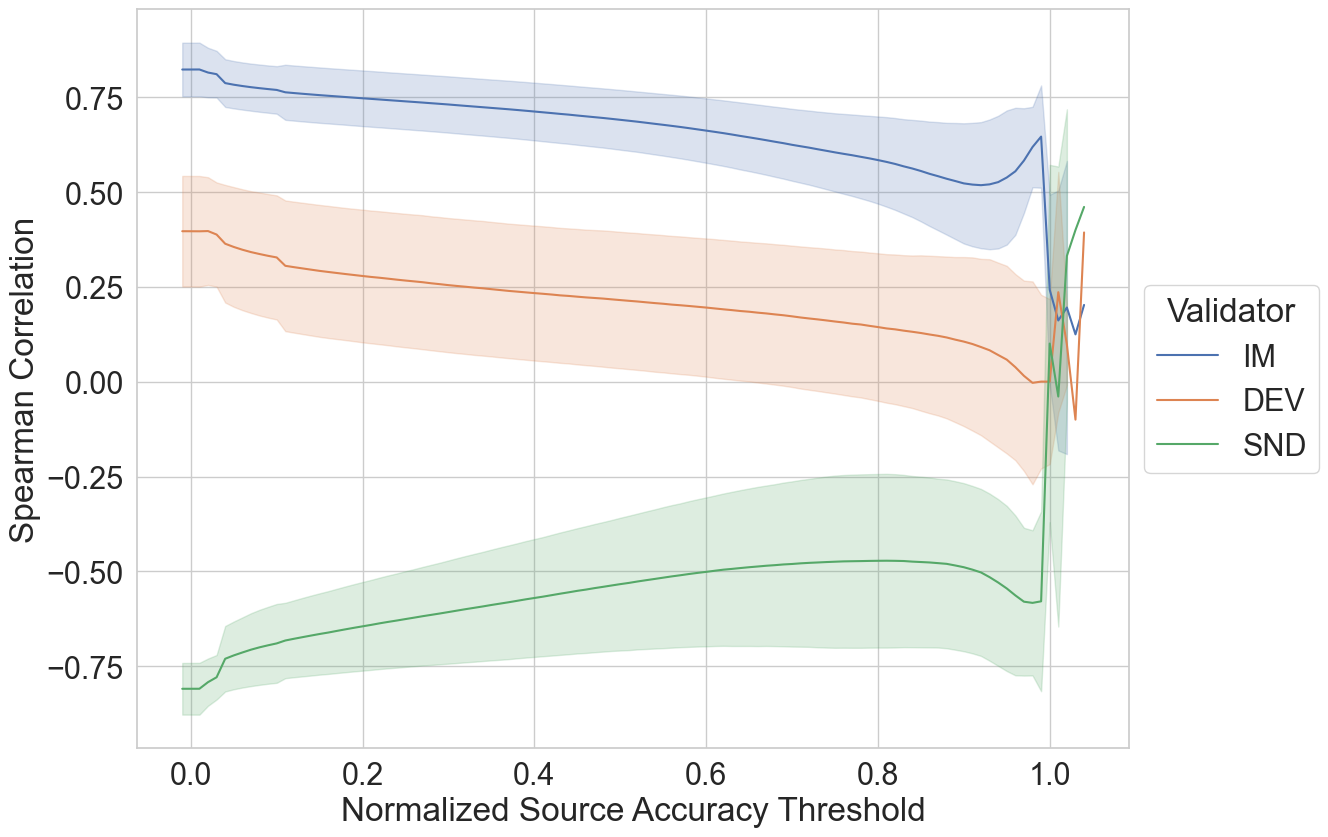}
         \caption{Correlation vs source accuracy threshold}
         \label{corrs_per_validator:a}
     \end{subfigure}
     \vskip\baselineskip
     \begin{subfigure}[b]{\columnwidth}
         \centering
         \includegraphics[width=\columnwidth]{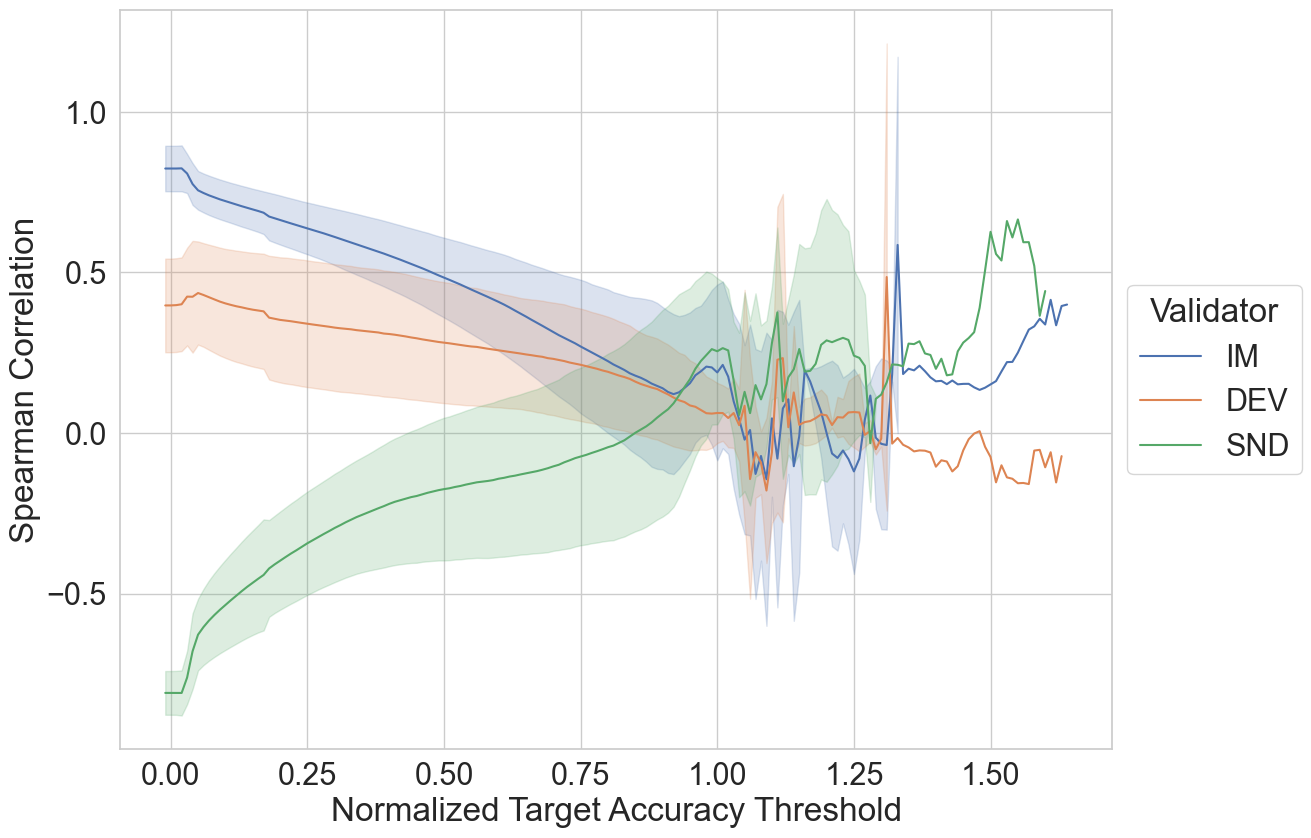}
         \caption{Correlation vs target accuracy threshold}
         \label{corrs_per_validator:b}
     \end{subfigure}
     \caption{The Spearman correlation between validation scores and target accuracy, as a function of accuracy threshold. At a threshold of $x$, only models with source/target accuracy greater than $x$ are kept. Accuracies are normalized so that the source-only model has a score of 1. Correlations are computed per transfer task. The lines and bands represent mean and standard deviation. As alluded to in Section \ref{section_impact_of_validation_methods}, the correlation within some tasks might be higher than Figure \ref{src_val_vs_target_train_scatter} suggests. For example, with no thresholding, DEV's mean, min, and max correlations are 0.40, 0.14, 0.59.}
    \label{corrs_per_validator}
\end{figure}

\begin{figure*}
     \centering
      \begin{subfigure}[b]{0.31\textwidth}
         \centering
         \includegraphics[width=\textwidth]{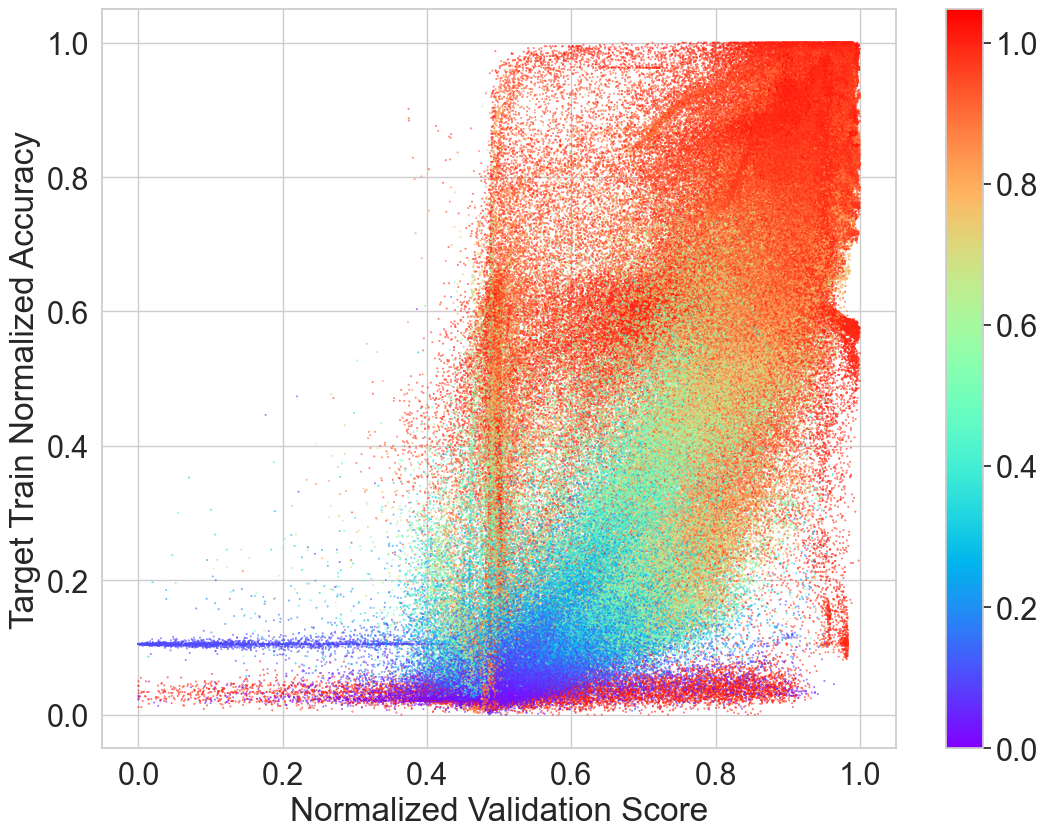}
         \caption{IM}
         \label{src_val_vs_target_train_scatter:a}
     \end{subfigure}
     \hfill
     \begin{subfigure}[b]{0.31\textwidth}
         \centering
         \includegraphics[width=\textwidth]{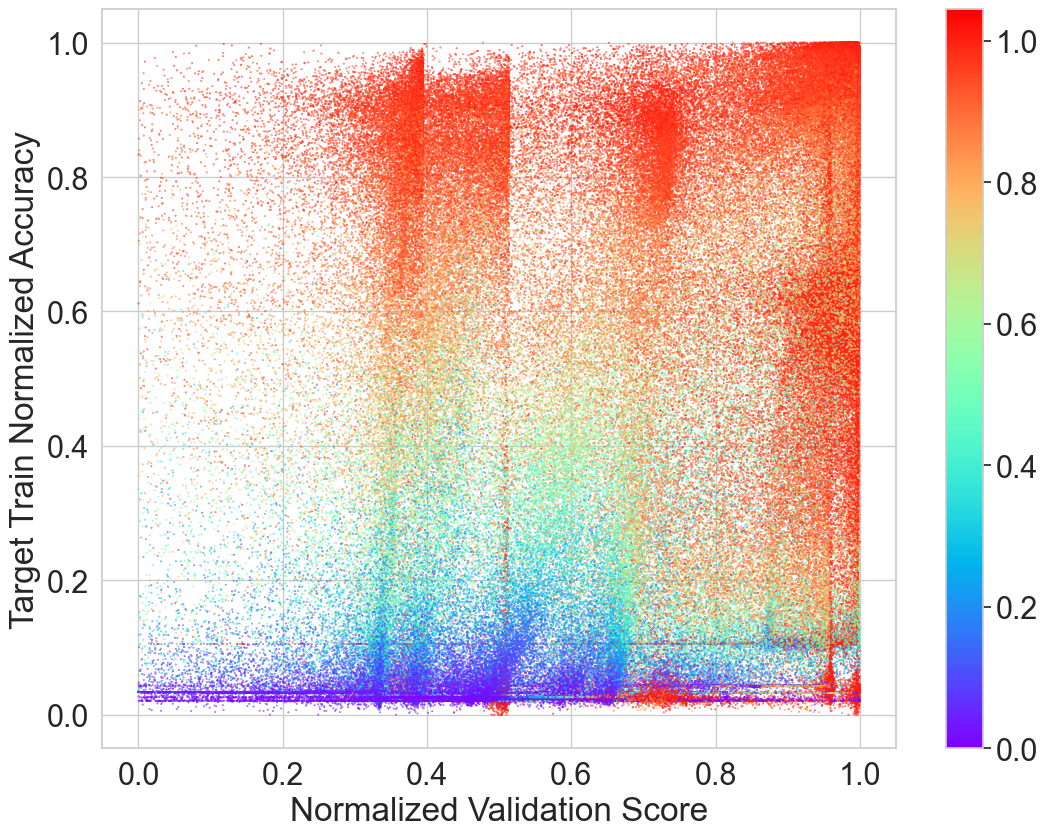}
         \caption{DEV}
         \label{src_val_vs_target_train_scatter:b}
     \end{subfigure}
     \hfill
     \begin{subfigure}[b]{0.31\textwidth}
         \centering
         \includegraphics[width=\textwidth]{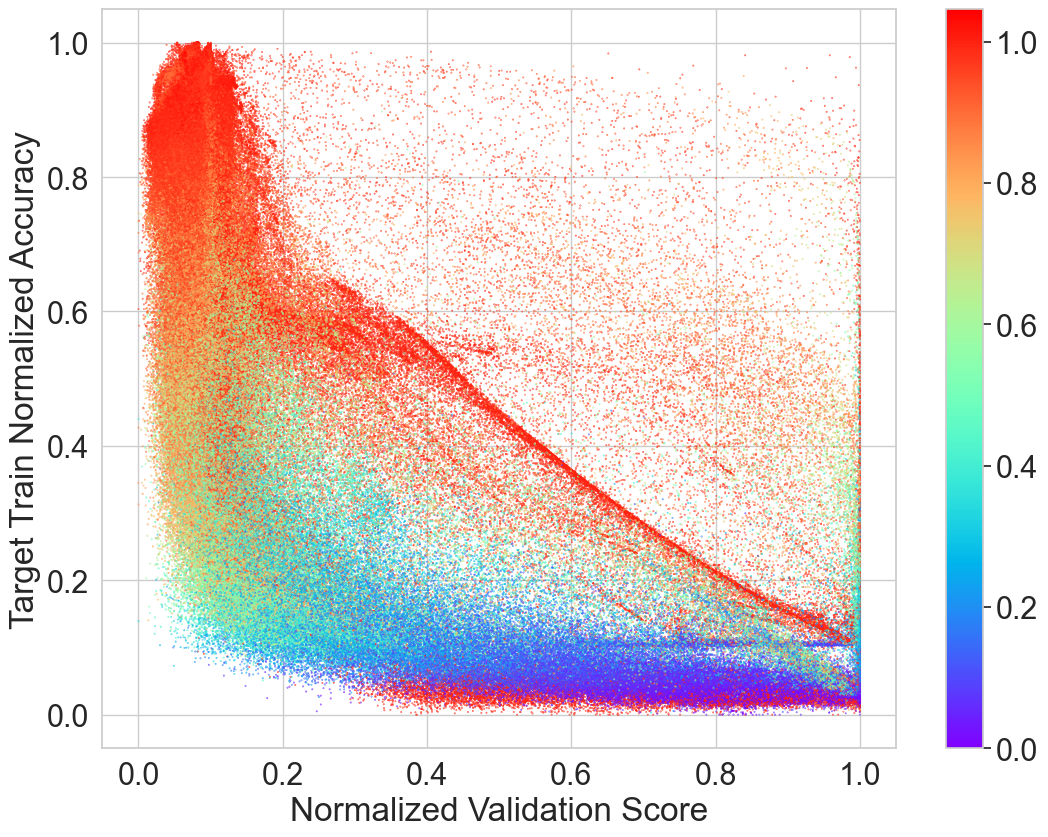}
         \caption{SND}
         \label{src_val_vs_target_train_scatter:c}
     \end{subfigure}
     \vskip\baselineskip
     \begin{subfigure}[b]{0.31\textwidth}
         \centering
         \includegraphics[width=\textwidth]{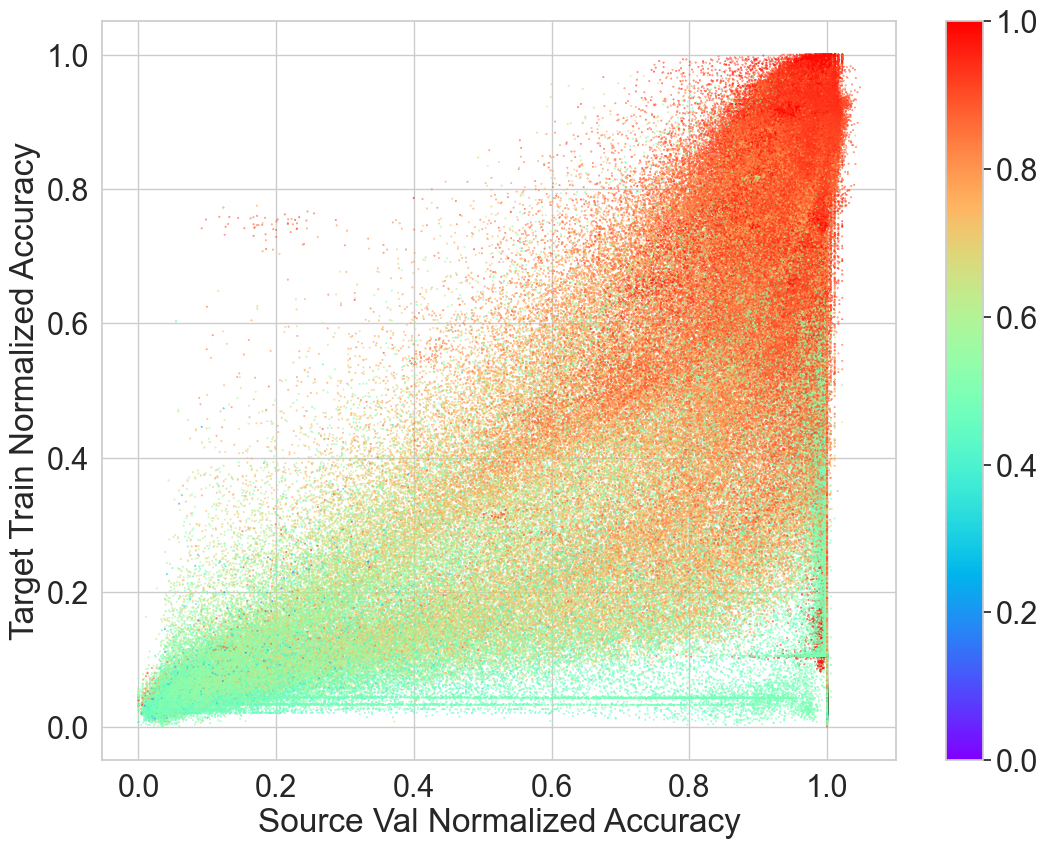}
         \caption{IM}
         \label{src_val_vs_target_train_scatter:d}
     \end{subfigure}
     \hfill
     \begin{subfigure}[b]{0.31\textwidth}
         \centering
         \includegraphics[width=\textwidth]{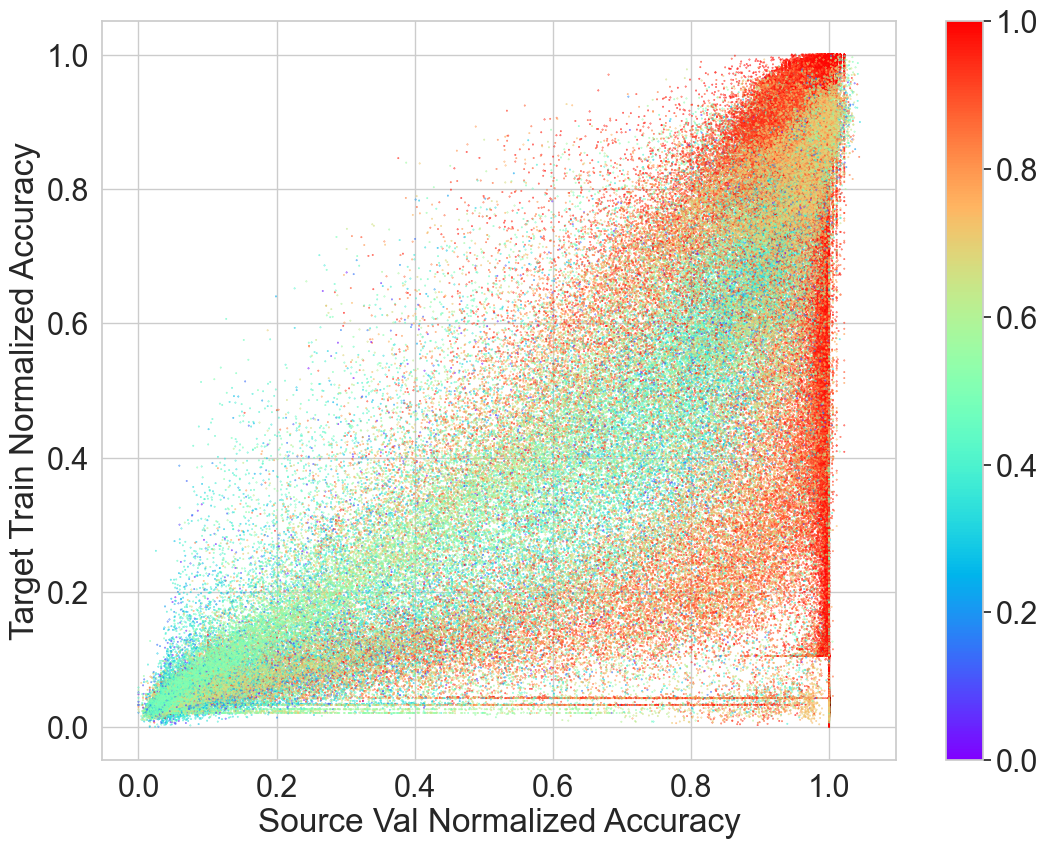}
         \caption{DEV}
         \label{src_val_vs_target_train_scatter:e}
     \end{subfigure}
     \hfill
     \begin{subfigure}[b]{0.31\textwidth}
         \centering
         \includegraphics[width=\textwidth]{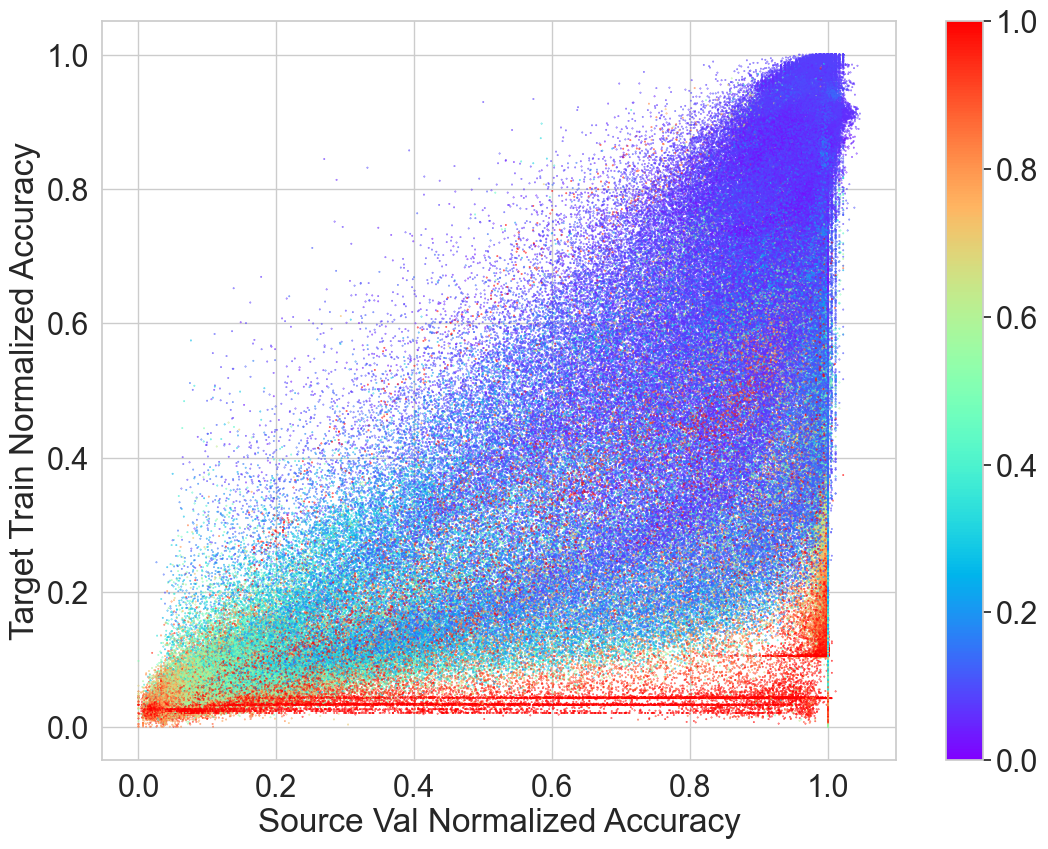}
         \caption{SND}
         \label{src_val_vs_target_train_scatter:f}
     \end{subfigure}
        \caption{The relationship between source accuracy, target accuracy, and validation scores. For each validation method and task, the validation score is min-max normalized, the target accuracy is max normalized, and the source accuracy is normalized by the source-only model's accuracy. \textbf{Top row colorbars} represent normalized source accuracy. \textbf{Bottom row colorbars} represent normalized validation score. As discussed in Section \ref{section_validation_methods_in_uda}, DEV can produce extremely large values, and our experiments confirm this. To make plots (b) and (e) legible, we exclude the lowest and highest 5\% of DEV validation scores.}
        \label{src_val_vs_target_train_scatter}
        \vspace{15pt}
\end{figure*}

We first consider the ``global" scenario in which validators are used to select model checkpoints, hyperparameters, \textit{and} algorithms. Figures \ref{src_val_vs_target_train_scatter:a}-\ref{src_val_vs_target_train_scatter:c} show the relationship between validation scores and target accuracy, using data from all transfer tasks. It appears that none of the methods are well-correlated with accuracy. (In fact, SND seems inversely correlated, which prompts us to add the negative SND score, NegSND, to our evaluation.) However, it is possible that the validators are well-correlated \textit{within} tasks, and are just producing inconsistent scores \textit{across} tasks (see Figure \ref{corrs_per_validator:a}). In addition, it may be possible to increase correlation by filtering out degenerate models.

\begin{figure}
    \centering
    \includegraphics[width=\columnwidth]{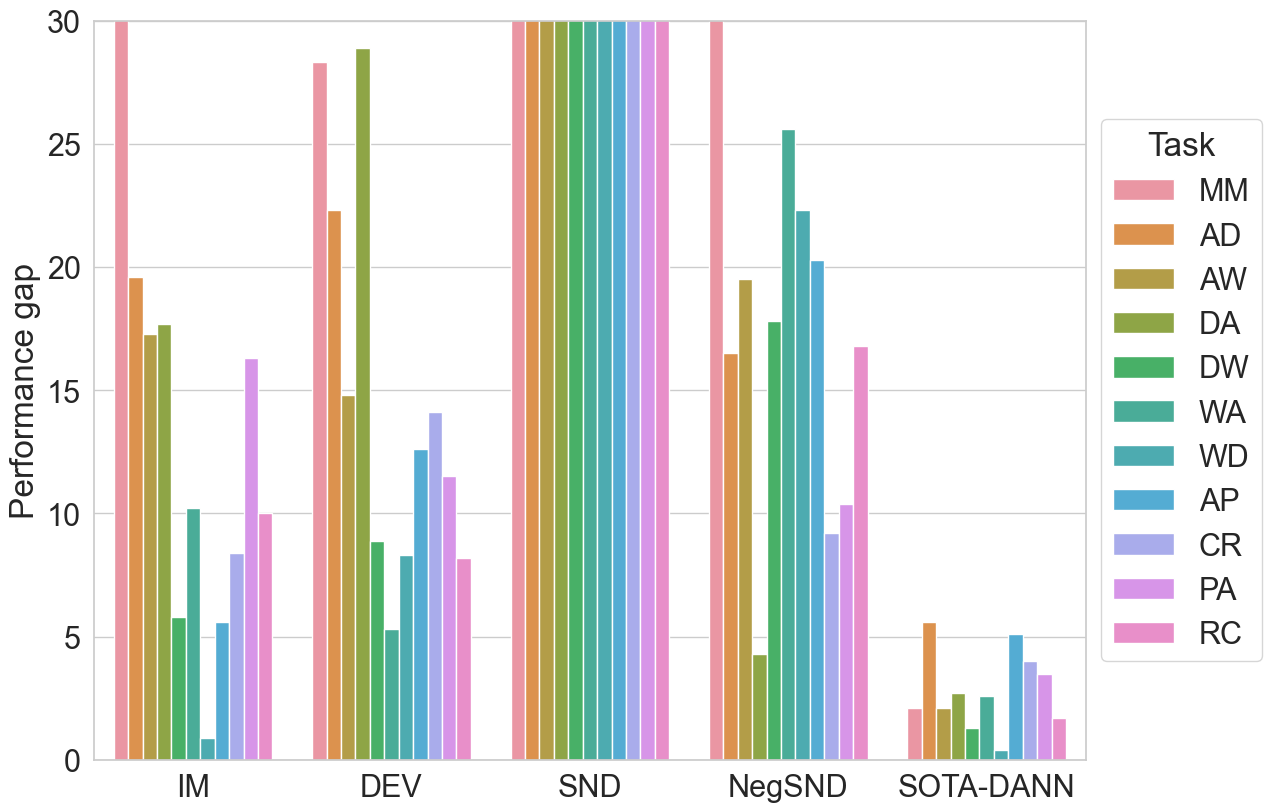}
    \caption{Performance gaps between the oracle and non-oracle validators (IM, DEV, SND, NegSND) using 0.98 source thresholding. SOTA-DANN is the difference between SOTA and DANN accuracies in the oracle setting. The y-axis is truncated at 30 for legibility.}
    \label{true_pred_global_diff}
\end{figure}
\begin{table}
\centering
\resizebox{\columnwidth}{!}{\begin{tabular}{lrrrr}
\toprule
Algorithm &         IM &        DEV &        SND &     NegSND \\
\midrule
AFN       &    3.4$\pm$3.1 &   9.6$\pm$13.8 &  17.0$\pm$21.0 &   7.5$\pm$12.5 \\
ATDOC     &  10.5$\pm$17.1 &  12.4$\pm$15.8 &  22.3$\pm$19.1 &    4.6$\pm$3.3 \\
BNM       &    4.7$\pm$3.9 &   7.3$\pm$12.1 &  15.4$\pm$21.7 &    6.1$\pm$2.9 \\
BSP       &    1.2$\pm$1.4 &   9.1$\pm$11.5 &  18.8$\pm$15.3 &   6.8$\pm$12.4 \\
CORAL     &    7.8$\pm$5.9 &   12.0$\pm$8.7 &  13.8$\pm$15.0 &    4.6$\pm$5.0 \\
DANN      &    9.3$\pm$6.9 &    5.5$\pm$6.5 &  11.5$\pm$14.4 &    7.5$\pm$7.6 \\
DC        &    5.0$\pm$2.9 &    3.0$\pm$2.6 &  12.1$\pm$14.0 &    7.1$\pm$7.6 \\
GVB       &    9.9$\pm$5.4 &  16.0$\pm$14.8 &  19.8$\pm$17.5 &    8.2$\pm$4.6 \\
JMMD      &    9.2$\pm$9.9 &   8.7$\pm$11.7 &  18.1$\pm$18.5 &    7.5$\pm$7.5 \\
MCC       &    6.4$\pm$3.2 &    3.3$\pm$1.9 &  11.6$\pm$18.0 &    7.2$\pm$3.0 \\
MCD       &    5.5$\pm$5.0 &   7.6$\pm$10.1 &  21.1$\pm$26.0 &    5.1$\pm$8.5 \\
MMD       &    8.9$\pm$7.1 &   9.6$\pm$12.1 &  20.6$\pm$15.1 &   8.7$\pm$11.0 \\
RTN       &    2.6$\pm$2.1 &  11.0$\pm$19.5 &  35.5$\pm$32.5 &    5.1$\pm$4.9 \\
SWD       &  10.0$\pm$10.3 &  10.7$\pm$15.7 &  17.7$\pm$13.7 &    5.9$\pm$7.3 \\
SymNets   &   10.1$\pm$9.0 &    8.8$\pm$8.9 &  57.8$\pm$30.8 &  14.7$\pm$14.2 \\
\bottomrule
\end{tabular}}
\caption{Performance gaps between oracle and non-oracle validators, per algorithm, using a 0.98 source threshold. The mean and standard deviations are computed across transfer tasks. Unlike Figure \ref{true_pred_global_diff} and Table \ref{src_threshold_table}, the oracle and non-oracle accuracies are collected \textit{per algorithm} instead of across algorithms.}
\label{true_pred_local_diff_table}
\end{table}

Saito et al \cite{saito2021tune} suggest discarding models with low source accuracy, since they are unlikely to score well on target data. This brings us to Figures \ref{src_val_vs_target_train_scatter:d}-\ref{src_val_vs_target_train_scatter:f}, which show that low source accuracy does indeed correspond with low target accuracy, though not vice versa. To determine a suitable threshold, we select the models with the best target accuracy for each transfer task, and take the average of their normalized source accuracies (i.e. normalized by the accuracy of the source-only model). The result is a normalized threshold of 0.98. Table \ref{src_threshold_table} shows how validators perform with and without a 0.98 threshold. In most cases, the threshold significantly boosts accuracy. But even so, the validators are still a long way from matching the oracle. On most tasks, the drops in accuracy caused by the validators are still much larger than the SOTA-baseline performance gaps (see Figure \ref{true_pred_global_diff}). In other words, the poor performance of the validators is of much greater concern than the relatively small differences between UDA algorithms.

\begin{table*}
\centering
\begin{tabular}{ll|rrrrrrrrrrr}
\toprule
Validator & Setting &    MM &    AD &    AW &    DA &    DW &    WA &    WD &    AP &    CR &    PA &    RC \\
\midrule
\multirow{3}{*}{IM} & None &  54.1 &  80.4 &  77.8 &  56.4 &  84.5 &  67.8 &  99.1 &  57.5 &  66.9 &  50.2 &  55.0 \\
       &  0.98 &  54.1 &  75.1 &  77.4 &  56.4 &  93.0 &  66.0 &  99.1 &  68.2 &  66.9 &  50.2 &  49.5 \\
       & Oracle & 95.2 &  94.7 &  94.7 &  74.1 &  98.9 &  76.2 &  100.0 &  73.8 &  75.3 &  66.5 &  59.5 \\
       \arrayrulecolor{gray}\hline
\multirow{3}{*}{DEV} & None &  10.0 &   3.2 &   3.2 &   3.2 &   3.2 &   3.2 &   3.2 &   1.5 &   0.6 &   1.5 &   1.5 \\
       &  0.98 &  67.0 &  73.3 &  79.1 &  45.0 &  89.8 &  70.0 &  91.7 &  61.0 &  61.8 &  53.5 &  50.5 \\
       & Oracle & 95.3 &  95.6 &  94.0 &  73.9 &  98.7 &  75.3 &  100.0 &  73.6 &  75.9 &  65.0 &  58.7 \\
    \hline
\multirow{3}{*}{SND} & None &  10.0 &   3.2 &   3.2 &   3.2 &   3.2 &   3.2 &   3.2 &   1.5 &   1.5 &   1.5 &   1.5 \\
       &  0.98 &  10.0 &   3.2 &   3.2 &   3.2 &   3.2 &   3.2 &   3.2 &   1.5 &   1.5 &   1.5 &   1.5 \\
       & Oracle & 93.4 &  95.4 &  94.8 &  73.4 &  99.0 &  75.4 &  100.0 &  73.6 &  74.9 &  66.1 &  57.9 \\
       \hline
\multirow{3}{*}{NegSND} & None &  40.7 &  65.8 &  29.6 &  25.6 &  43.5 &  10.6 &  74.6 &  58.1 &  54.3 &  40.2 &  43.7 \\
       &  0.98 &  53.6 &  78.9 &  75.3 &  69.2 &  81.2 &  49.8 &  77.7 &  53.4 &  65.8 &  55.7 &  41.1 \\
       & Oracle & 93.4 &  95.4 &  94.8 &  73.4 &  99.0 &  75.4 &  100.0 &  73.6 &  74.9 &  66.1 &  57.9 \\
\arrayrulecolor{black}\bottomrule
\end{tabular}
\caption{The best target train accuracy for each validation method under two settings: no source thresholding (``None"), and 0.98 source thresholding. For example, say the source-only model has 50\% source accuracy. The 0.98 setting will keep only the models that score higher than 49\% on the source data, while the None setting will keep all models. The third setting, Oracle, is the true best target accuracy. Note that these oracle values differ from Tables \ref{office31_oracle_results} and \ref{mnist_officehome_oracle_results} because these are computed on the target \textit{train} set, and are also from entirely different training runs.}
\label{src_threshold_table}
\end{table*}

Now we consider the ``local" scenario in which the validator selects checkpoints and hyperparameters, but not algorithms. In this case, some algorithm-validator pairs can work quite well, as shown in Table \ref{true_pred_local_diff_table}. However, many pairs have high variance, so it is difficult to know how reliable they will be when given a new transfer task.

Finally, we consider an unrealistic scenario in which we are able to discard models with low target accuracy. Figure \ref{corrs_per_validator:b} shows that even if we remove models with a target accuracy less than that of the source-only model, the validators' correlations with accuracy are still below 0.3 on average.

\section{Discussion}
We have shown that the gap between SOTA and baseline UDA algorithms is smaller than previously thought. Furthermore, existing validators cause large drops in accuracy that make the differences between algorithms seem insignificant. In the scenario where the algorithm is already chosen, some algorithm-validator pairs can be effective, though most suffer from inconsistent performance across tasks. Consistency matters, because if a validator returns a high score, we need to be confident that the accuracy will also be high. Otherwise we will waste time and money that could be better spent on labeling the target data, eliminating the need for UDA altogether. Thus, one direction of research could be to create validators that work \textit{consistently} well, even if they work with just a single UDA algorithm.

\textbf{Limitations}: To compare algorithms fairly, and to limit the scope of the hyperparameter search, we used the same optimizer, weight decay, learning rate (LR) scheduler, and batch size across all experiments. In addition, for any chosen LR, we applied the same LR to all models, which may not always be optimal. We believe we chose reasonable defaults, and we also allowed for plenty of flexibility in the weighting of loss terms for each algorithm  (see the supplementary material). That said, it is possible that some algorithms require a different setting to reach their full potential.

\textbf{Societal impact}: Large scale machine learning experiments consume a great deal of energy. In our case, the end result is a better understanding of UDA, which is an area of central importance in the data efficiency agenda. As unlabeled data becomes available in new domains, UDA will allow for efficient reuse of existing models.

{\small
\bibliographystyle{ieee_fullname}
\bibliography{egbib}
}

\clearpage

\appendix
\section{Paper Meta Analysis}
\begin{figure*}
     \centering
      \begin{subfigure}[b]{0.48\textwidth}
         \centering
         \includegraphics[width=\textwidth]{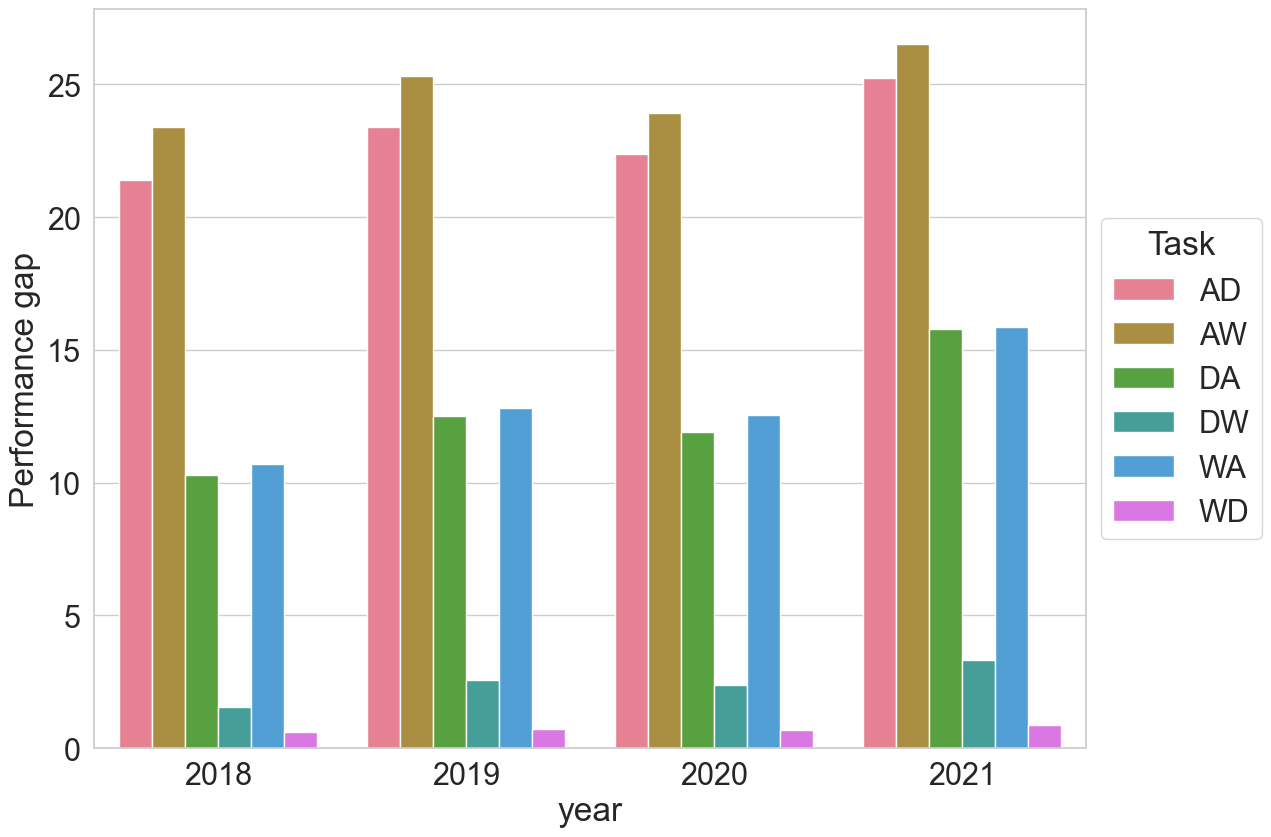}
         \caption{Reported performance gap over ResNet50 (source-only) for Office31.}
         
     \end{subfigure}
     \hfill
     \begin{subfigure}[b]{0.48\textwidth}
         \centering
         \includegraphics[width=\textwidth]{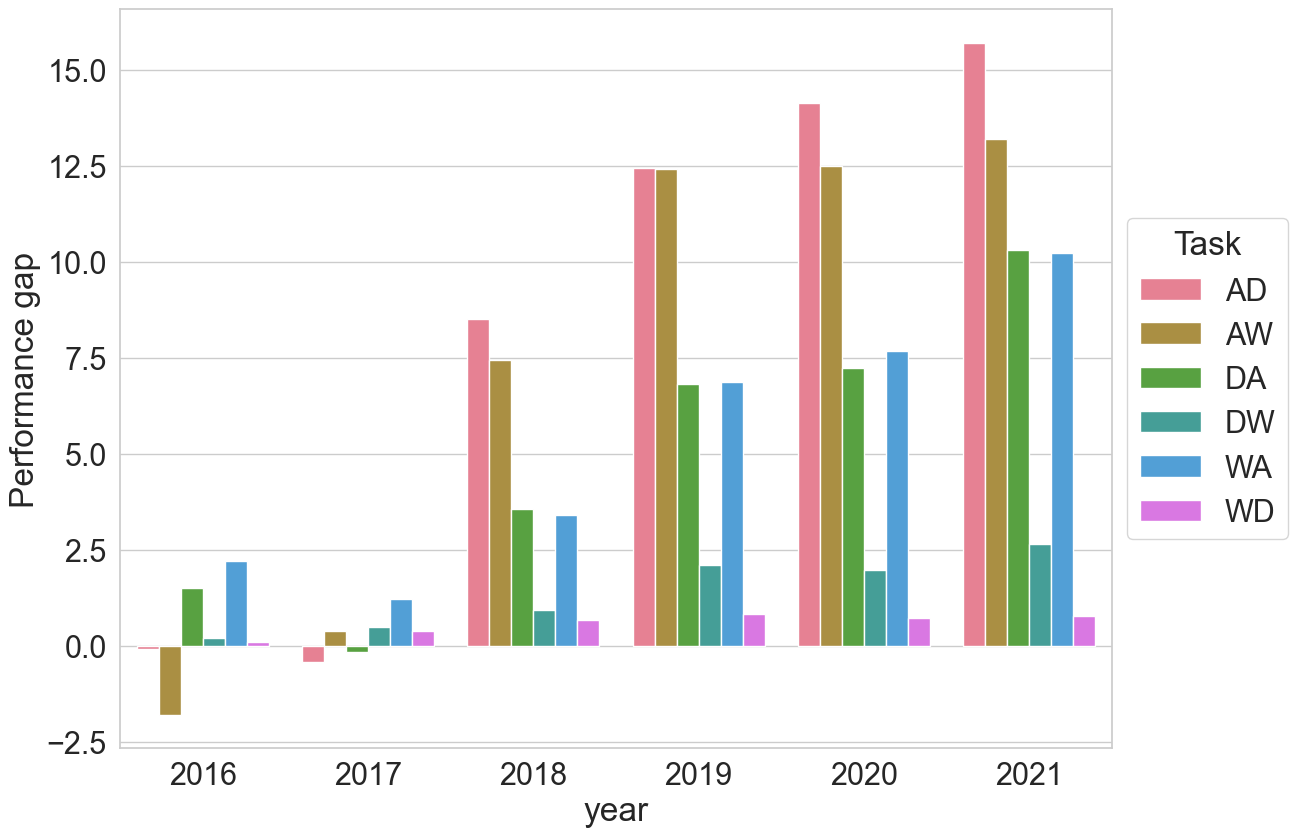}
         \caption{Reported performance gap over DANN for Office31.}
         
     \end{subfigure}
     \vskip\baselineskip
     \begin{subfigure}[b]{0.48\textwidth}
         \centering
         \includegraphics[width=\textwidth]{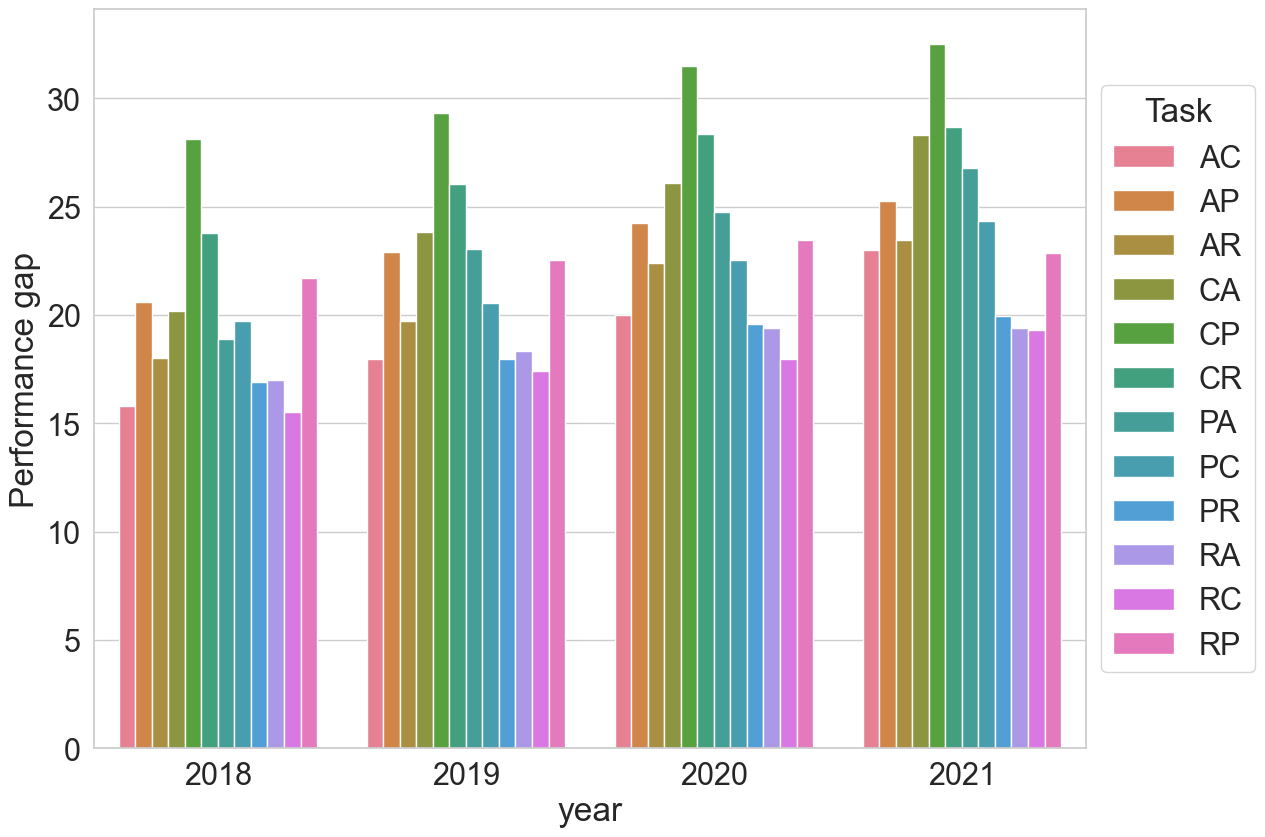}
         \caption{Reported performance gap over ResNet50 (source-only) for OfficeHome.}
         
     \end{subfigure}
     \hfill
     \begin{subfigure}[b]{0.48\textwidth}
         \centering
         \includegraphics[width=\textwidth]{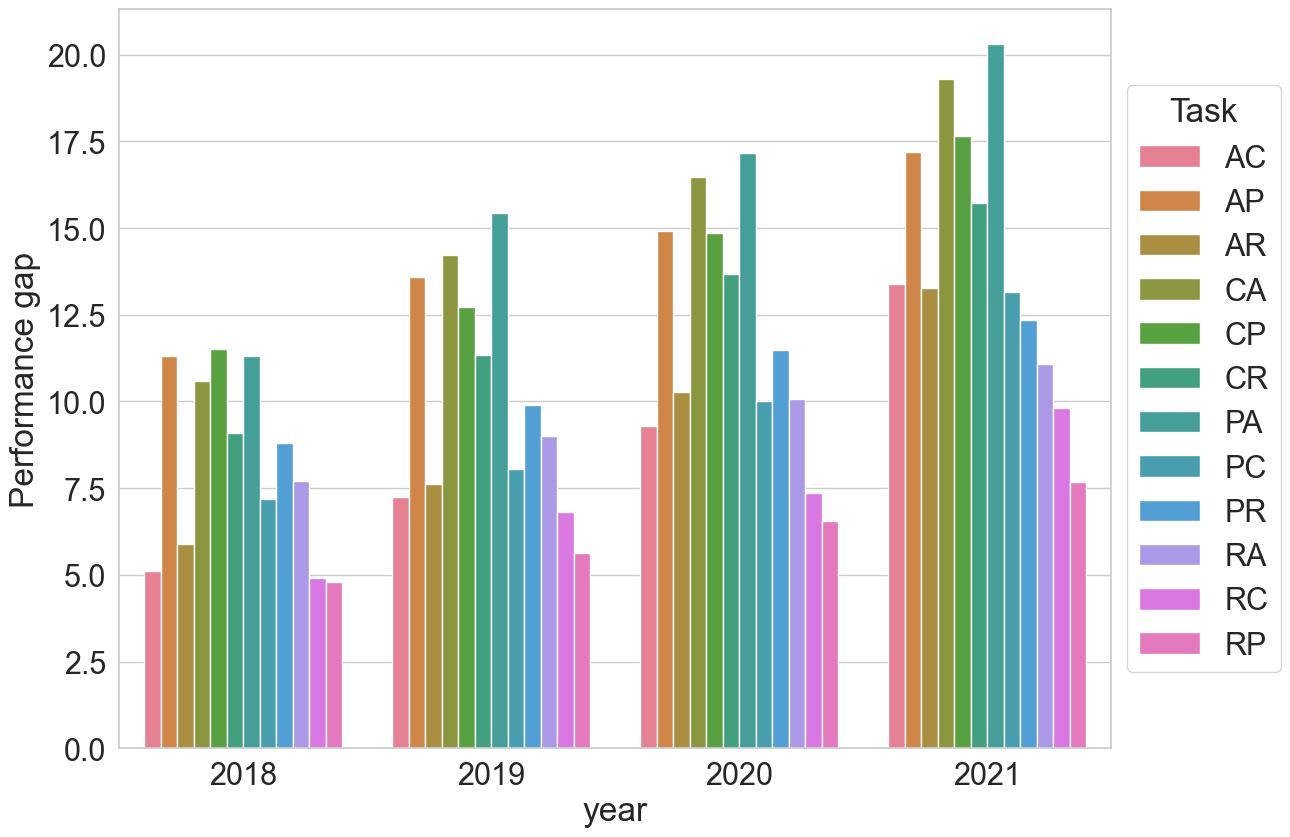}
         \caption{Reported performance gap over DANN for OfficeHome.}
         
     \end{subfigure}
    \caption{The average reported SOTA-baseline performance gap per year. For example, in Figure (d), the OfficeHome Product$\rightarrow$Art (PA) value for DANN in 2021 is 20.3. This means that, on average, 2021 papers report that the best performing algorithm in the PA task has a 20.3 point advantage over the reported DANN accuracy.}
    \label{supp_reported_gap}
\end{figure*}

Figure \ref{supp_reported_gap} shows the reported performance gaps for Office31 and OfficeHome. Table \ref{supp_meta_analysis_types_of_validation_method} contains definitions of the validation methods we found in papers and repos.

\section{Experiment Methodology}
Tables \ref{supp_datasets_table}-\ref{supp_num_datapoints} provide details about dataset splits, models, and other experiment settings.

The $x$-DANN combinations (like MCC-DANN) are missing from the hyperparameter search table (Table \ref{supp_hyperparameter_search}). For these combinations, we searched only the $x$ hyperparameters, and kept the DANN hyperparameters frozen to the best values found in the DANN experiments.

\section{Results}
Tables \ref{supp_office31_oracle_results_std} and \ref{supp_mnist_officehome_oracle_results_std} show the standard deviations of the 5 runs for each algorithm in the oracle setting. Bold indicates the lowest value per column, and lower values have a stronger green color. Dashes indicate that reproductions had not yet run when the tables were constructed, so a standard deviation could not be calculated.

Tables \ref{supp_per_algorithm_diff_IM}-\ref{supp_per_algorithm_diff_NegSND} show the performance gap between oracle and non-oracle validators, per algorithm, at a 0.98 source threshold. Dashes indicate that either all models were discarded with the 0.98 threshold, or that those algorithm/validator/task combinations had not yet run.

\begin{figure*}[p]
     \centering
      \begin{subfigure}[b]{0.31\textwidth}
         \centering
         \includegraphics[width=\textwidth]{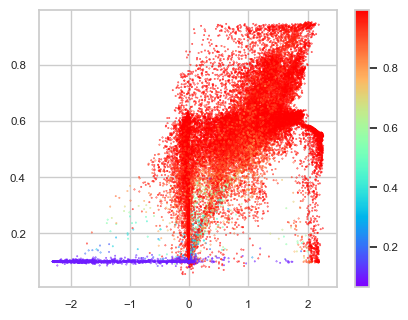}
         \caption{IM FL0}
         
     \end{subfigure}
     \hfill
       \begin{subfigure}[b]{0.31\textwidth}
         \centering
         \includegraphics[width=\textwidth]{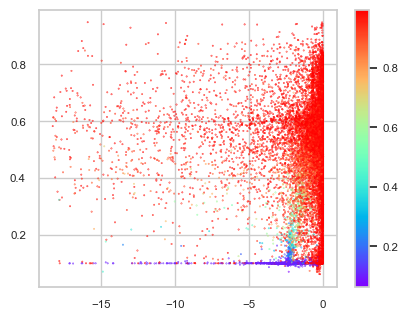}
         \caption{DEV FL0}
         
     \end{subfigure}
          \hfill
       \begin{subfigure}[b]{0.31\textwidth}
         \centering
         \includegraphics[width=\textwidth]{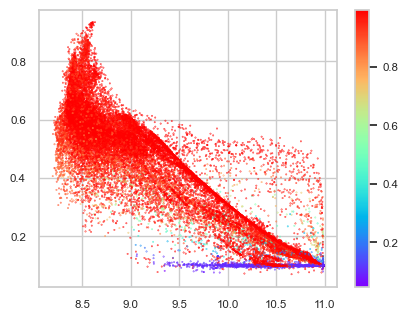}
         \caption{SND FL0}
         
     \end{subfigure}
     \vskip\baselineskip
       \begin{subfigure}[b]{0.31\textwidth}
         \centering
         \includegraphics[width=\textwidth]{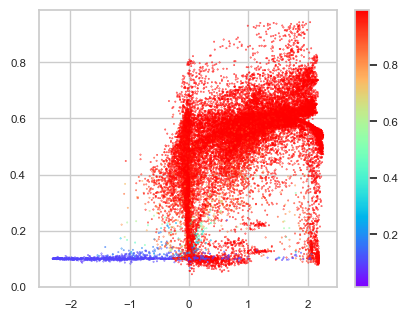}
         \caption{IM FL6}
         
     \end{subfigure}
     \hfill
       \begin{subfigure}[b]{0.31\textwidth}
         \centering
         \includegraphics[width=\textwidth]{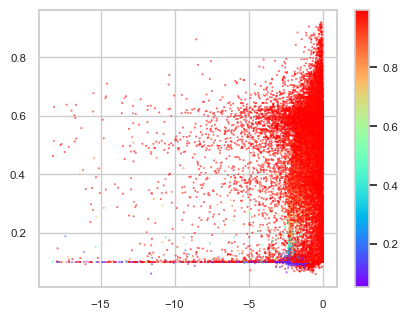}
         \caption{DEV FL6}
         
     \end{subfigure}
          \hfill
       \begin{subfigure}[b]{0.31\textwidth}
         \centering
         \includegraphics[width=\textwidth]{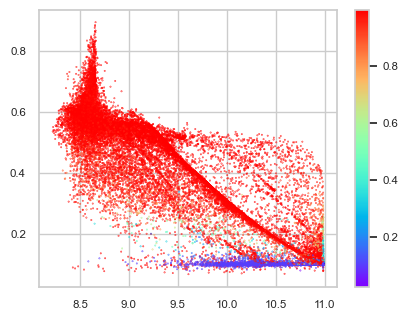}
         \caption{SND FL6}
         
     \end{subfigure}
        \caption{MNIST$\rightarrow$MNISTM task. \textbf{x-axis}: validation score, \textbf{y-axis}: target train accuracy, \textbf{colorbar}: source accuracy.}
        \label{supp_mm_scatter_plots}
\end{figure*}

\begin{figure*}[p]
     \centering
      \begin{subfigure}[b]{0.31\textwidth}
         \centering
         \includegraphics[width=\textwidth]{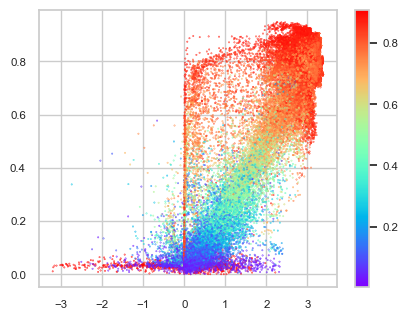}
         \caption{IM FL0}
         
     \end{subfigure}
     \hfill
       \begin{subfigure}[b]{0.31\textwidth}
         \centering
         \includegraphics[width=\textwidth]{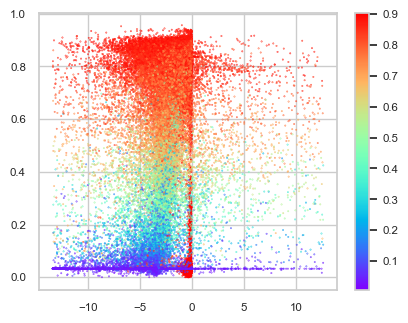}
         \caption{DEV FL0}
         
     \end{subfigure}
          \hfill
       \begin{subfigure}[b]{0.31\textwidth}
         \centering
         \includegraphics[width=\textwidth]{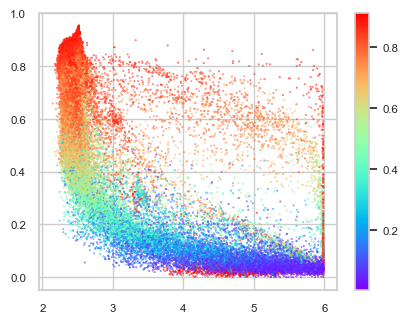}
         \caption{SND FL0}
         
     \end{subfigure}
        \caption{Office31 AD task. \textbf{x-axis}: validation score, \textbf{y-axis}: target train accuracy, \textbf{colorbar}: source accuracy.}
        \label{supp_ad_scatter_plots}
\end{figure*}

\begin{figure*}[p]
     \centering
      \begin{subfigure}[b]{0.31\textwidth}
         \centering
         \includegraphics[width=\textwidth]{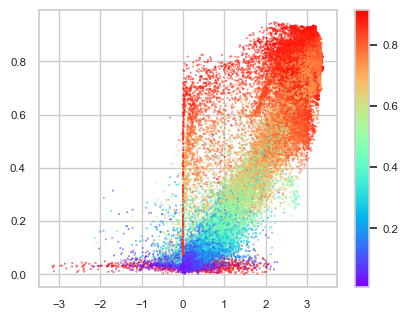}
         \caption{IM FL0}
         
     \end{subfigure}
     \hfill
       \begin{subfigure}[b]{0.31\textwidth}
         \centering
         \includegraphics[width=\textwidth]{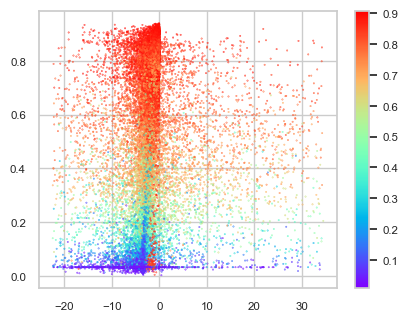}
         \caption{DEV FL0}
         
     \end{subfigure}
          \hfill
       \begin{subfigure}[b]{0.31\textwidth}
         \centering
         \includegraphics[width=\textwidth]{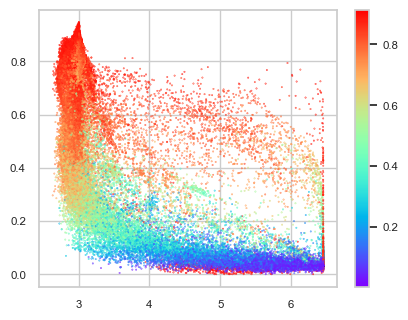}
         \caption{SND FL0}
         
     \end{subfigure}
        \caption{Office31 AW task. \textbf{x-axis}: validation score, \textbf{y-axis}: target train accuracy, \textbf{colorbar}: source accuracy.}
        \label{supp_aw_scatter_plots}
\end{figure*}

\begin{figure*}[p]
     \centering
      \begin{subfigure}[b]{0.31\textwidth}
         \centering
         \includegraphics[width=\textwidth]{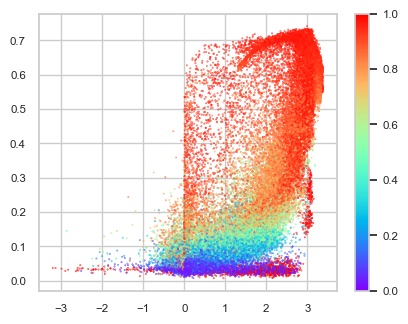}
         \caption{IM FL0}
         
     \end{subfigure}
     \hfill
       \begin{subfigure}[b]{0.31\textwidth}
         \centering
         \includegraphics[width=\textwidth]{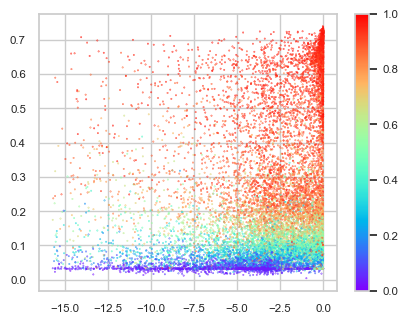}
         \caption{DEV FL0}
         
     \end{subfigure}
          \hfill
       \begin{subfigure}[b]{0.31\textwidth}
         \centering
         \includegraphics[width=\textwidth]{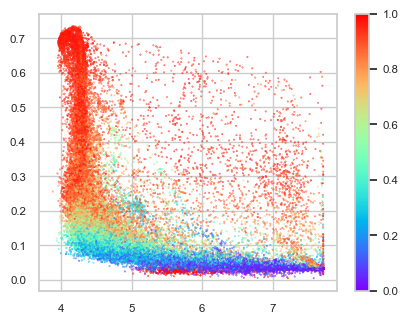}
         \caption{SND FL0}
         
     \end{subfigure}
        \caption{Office31 DA task. \textbf{x-axis}: validation score, \textbf{y-axis}: target train accuracy, \textbf{colorbar}: source accuracy.}
        \label{supp_da_scatter_plots}
\end{figure*}

\begin{figure*}[p]
     \centering
      \begin{subfigure}[b]{0.31\textwidth}
         \centering
         \includegraphics[width=\textwidth]{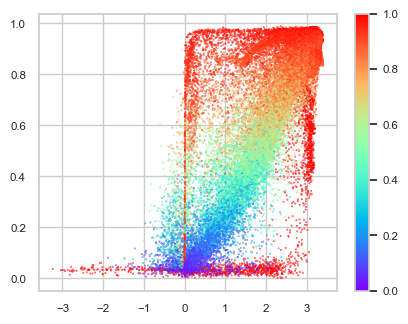}
         \caption{IM FL0}
         
     \end{subfigure}
     \hfill
       \begin{subfigure}[b]{0.31\textwidth}
         \centering
         \includegraphics[width=\textwidth]{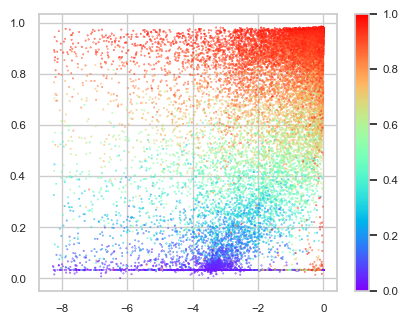}
         \caption{DEV FL0}
         
     \end{subfigure}
          \hfill
       \begin{subfigure}[b]{0.31\textwidth}
         \centering
         \includegraphics[width=\textwidth]{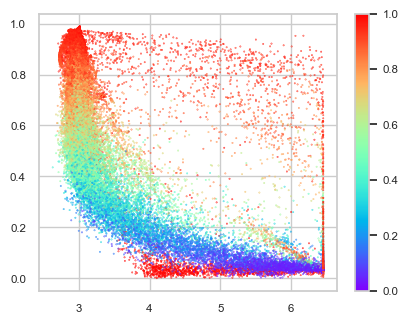}
         \caption{SND FL0}
         
     \end{subfigure}
        \caption{Office31 DW task. \textbf{x-axis}: validation score, \textbf{y-axis}: target train accuracy, \textbf{colorbar}: source accuracy.}
        \label{supp_dw_scatter_plots}
\end{figure*}

\begin{figure*}[p]
     \centering
      \begin{subfigure}[b]{0.31\textwidth}
         \centering
         \includegraphics[width=\textwidth]{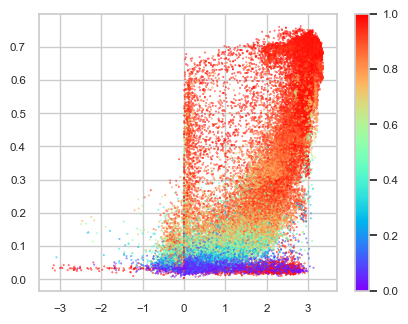}
         \caption{IM FL0}
         
     \end{subfigure}
     \hfill
       \begin{subfigure}[b]{0.31\textwidth}
         \centering
         \includegraphics[width=\textwidth]{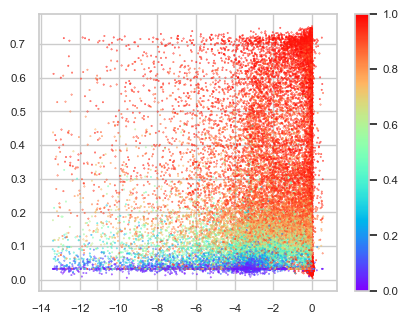}
         \caption{DEV FL0}
         
     \end{subfigure}
          \hfill
       \begin{subfigure}[b]{0.31\textwidth}
         \centering
         \includegraphics[width=\textwidth]{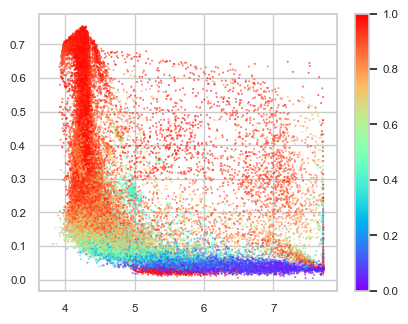}
         \caption{SND FL0}
         
     \end{subfigure}
        \caption{Office31 WA task. \textbf{x-axis}: validation score, \textbf{y-axis}: target train accuracy, \textbf{colorbar}: source accuracy.}
        \label{supp_wa_scatter_plots}
\end{figure*}

\begin{figure*}[p]
     \centering
      \begin{subfigure}[b]{0.31\textwidth}
         \centering
         \includegraphics[width=\textwidth]{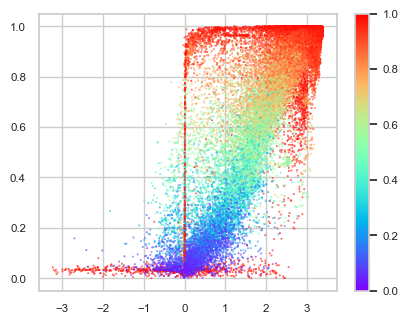}
         \caption{IM FL0}
     \end{subfigure}
     \hfill
       \begin{subfigure}[b]{0.31\textwidth}
         \centering
         \includegraphics[width=\textwidth]{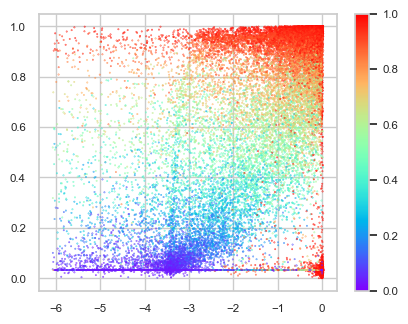}
         \caption{DEV FL0}
     \end{subfigure}
          \hfill
       \begin{subfigure}[b]{0.31\textwidth}
         \centering
         \includegraphics[width=\textwidth]{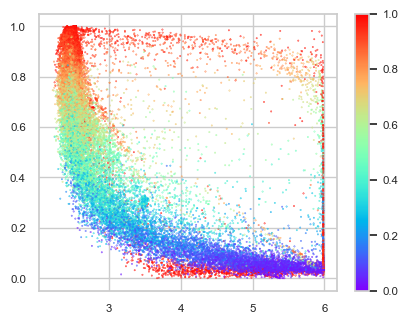}
         \caption{SND FL0}
     \end{subfigure}
        \caption{Office31 WD task. \textbf{x-axis}: validation score, \textbf{y-axis}: target train accuracy, \textbf{colorbar}: source accuracy.}
        \label{supp_wd_scatter_plots}
\end{figure*}

\begin{figure*}[p]
     \centering
      \begin{subfigure}[b]{0.31\textwidth}
         \centering
         \includegraphics[width=\textwidth]{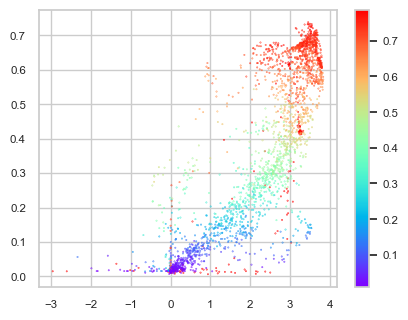}
         \caption{IM FL0}
         
     \end{subfigure}
     \hfill
       \begin{subfigure}[b]{0.31\textwidth}
         \centering
         \includegraphics[width=\textwidth]{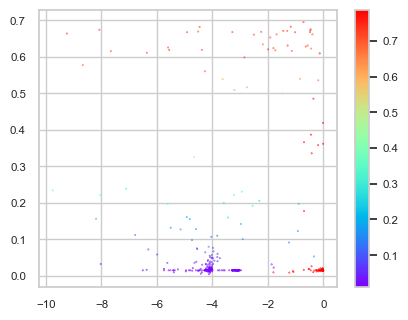}
         \caption{DEV FL0}
         
     \end{subfigure}
          \hfill
       \begin{subfigure}[b]{0.31\textwidth}
         \centering
         \includegraphics[width=\textwidth]{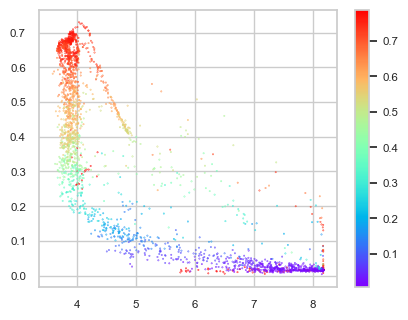}
         \caption{SND FL0}
         
     \end{subfigure}
          \vskip\baselineskip
       \begin{subfigure}[b]{0.31\textwidth}
         \centering
         \includegraphics[width=\textwidth]{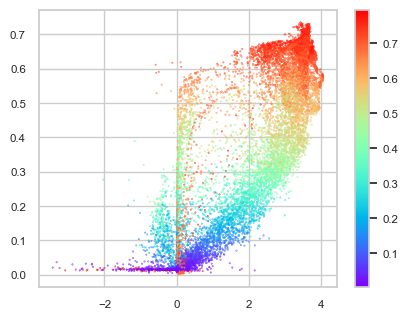}
         \caption{IM FL6}
         
     \end{subfigure}
     \hfill
       \begin{subfigure}[b]{0.31\textwidth}
         \centering
         \includegraphics[width=\textwidth]{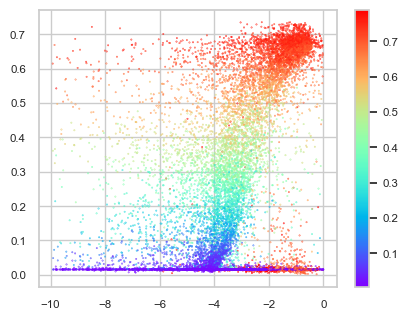}
         \caption{DEV FL6}
         
     \end{subfigure}
          \hfill
       \begin{subfigure}[b]{0.31\textwidth}
         \centering
         \includegraphics[width=\textwidth]{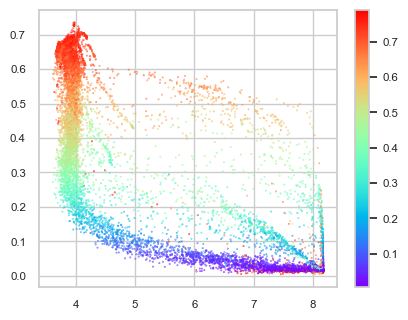}
         \caption{SND FL6}
         
     \end{subfigure}
        \caption{OfficeHome AP task. \textbf{x-axis}: validation score, \textbf{y-axis}: target train accuracy, \textbf{colorbar}: source accuracy.}
        \label{supp_ap_scatter_plots}
\end{figure*}

\begin{figure*}[p]
     \centering
      \begin{subfigure}[b]{0.31\textwidth}
         \centering
         \includegraphics[width=\textwidth]{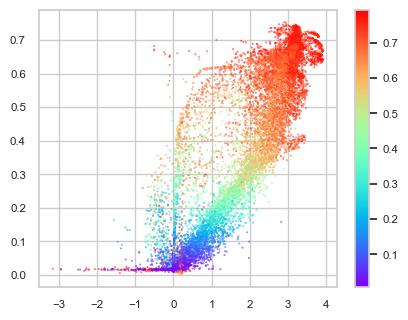}
         \caption{IM FL6}
         
     \end{subfigure}
     \hfill
       \begin{subfigure}[b]{0.31\textwidth}
         \centering
         \includegraphics[width=\textwidth]{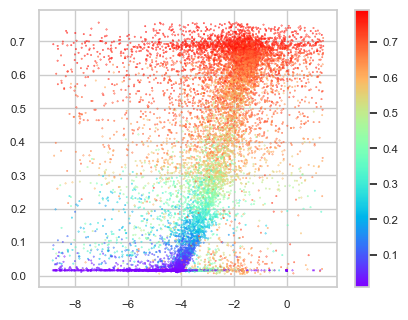}
         \caption{DEV FL6}
         
     \end{subfigure}
          \hfill
       \begin{subfigure}[b]{0.31\textwidth}
         \centering
         \includegraphics[width=\textwidth]{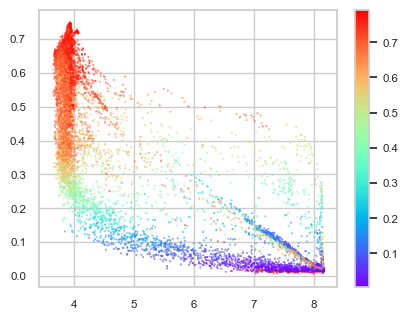}
         \caption{SND FL6}
         
     \end{subfigure}
        \caption{OfficeHome CR task. \textbf{x-axis}: validation score, \textbf{y-axis}: target train accuracy, \textbf{colorbar}: source accuracy.}
        \label{supp_cr_scatter_plots}
\end{figure*}

\begin{figure*}[p]
     \centering
      \begin{subfigure}[b]{0.31\textwidth}
         \centering
         \includegraphics[width=\textwidth]{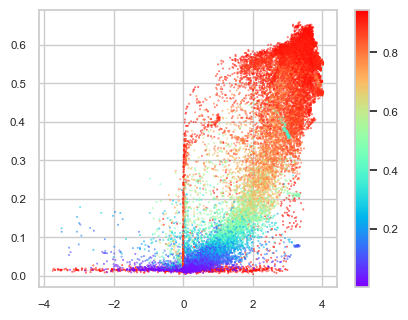}
         \caption{IM FL0}
         
     \end{subfigure}
     \hfill
       \begin{subfigure}[b]{0.31\textwidth}
         \centering
         \includegraphics[width=\textwidth]{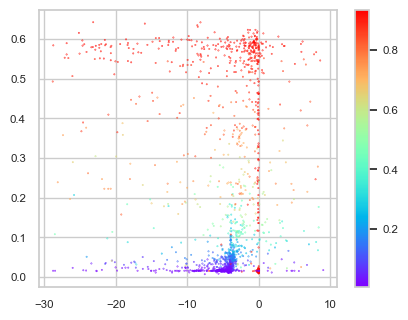}
         \caption{DEV FL0}
         
     \end{subfigure}
          \hfill
       \begin{subfigure}[b]{0.31\textwidth}
         \centering
         \includegraphics[width=\textwidth]{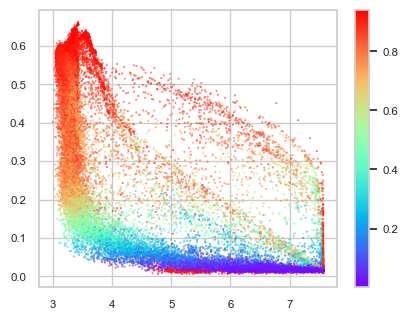}
         \caption{SND FL0}
         
     \end{subfigure}
          \vskip\baselineskip
       \begin{subfigure}[b]{0.31\textwidth}
         \centering
         \includegraphics[width=\textwidth]{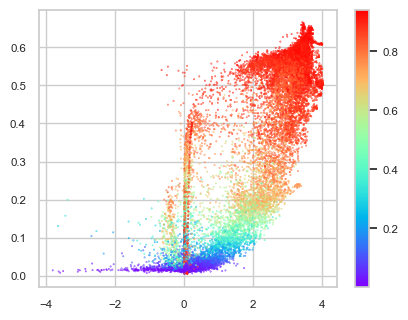}
         \caption{IM FL6}
         
     \end{subfigure}
     \hfill
       \begin{subfigure}[b]{0.31\textwidth}
         \centering
         \includegraphics[width=\textwidth]{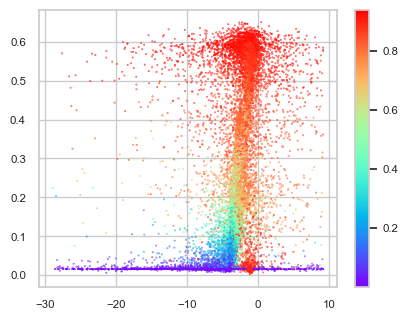}
         \caption{DEV FL6}
         
     \end{subfigure}
          \hfill
       \begin{subfigure}[b]{0.31\textwidth}
         \centering
         \includegraphics[width=\textwidth]{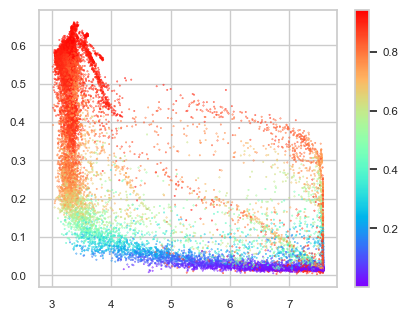}
         \caption{SND FL6}
         
     \end{subfigure}
        \caption{OfficeHome PA task. \textbf{x-axis}: validation score, \textbf{y-axis}: target train accuracy, \textbf{colorbar}: source accuracy.}
        \label{supp_pa_scatter_plots}
\end{figure*}

\begin{figure*}[p]
     \centering
      \begin{subfigure}[b]{0.31\textwidth}
         \centering
         \includegraphics[width=\textwidth]{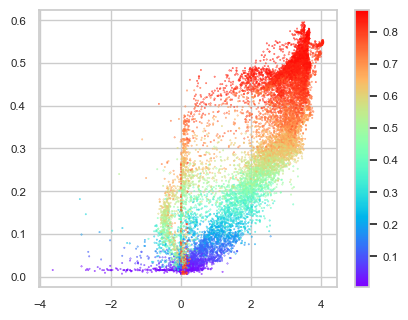}
         \caption{IM FL6}
         
     \end{subfigure}
     \hfill
       \begin{subfigure}[b]{0.31\textwidth}
         \centering
         \includegraphics[width=\textwidth]{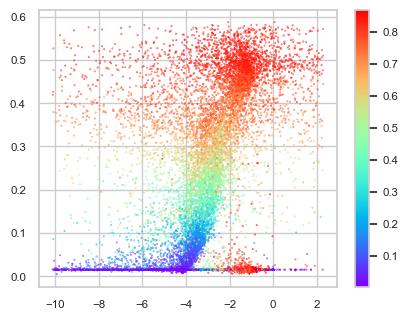}
         \caption{DEV FL6}
         
     \end{subfigure}
          \hfill
       \begin{subfigure}[b]{0.31\textwidth}
         \centering
         \includegraphics[width=\textwidth]{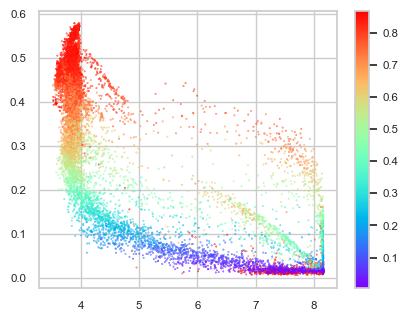}
         \caption{SND FL6}
         
     \end{subfigure}
        \caption{OfficeHome RC task. \textbf{x-axis}: validation score, \textbf{y-axis}: target train accuracy, \textbf{colorbar}: source accuracy.}
        \label{supp_rc_scatter_plots}
\end{figure*}

Figures \ref{supp_mm_scatter_plots}-\ref{supp_rc_scatter_plots} are scatter plots of validation scores vs target accuracy, per transfer task and feature layer. All values are unnormalized. For DEV, the lowest and highest 5\% are excluded to make the plots legible.

\textbf{Note about DEV}: The original DEV risk score is supposed to be minimized. Our code is designed to maximize validation scores, so we maximize the negative DEV risk. For the loss function $\ell$ (described in the DEV paper), we use cross entropy.

\begin{table}[h]
\centering
\begin{tabular}{c c c c} 
 \toprule
 Dataset & Domain & Train & Val \\
 \midrule
 \begin{tabular}{@{}c@{}} MNIST \end{tabular} &
\begin{tabular}{@{}c@{}} MNIST \\ MNISTM \end{tabular} &
\begin{tabular}{@{}c@{}} 60000 \\ 59001 \end{tabular} &
\begin{tabular}{@{}c@{}} 10000 \\ 9001 \end{tabular} \\
 \arrayrulecolor{gray}\hline
  \begin{tabular}{@{}c@{}} Office31 \end{tabular} &
\begin{tabular}{@{}c@{}} Amazon (A) \\ DSLR (D) \\ Webcam (W) \end{tabular} &
\begin{tabular}{@{}c@{}} 2253 \\ 398 \\ 636  \end{tabular} &
\begin{tabular}{@{}c@{}} 564 \\ 100 \\ 159  \end{tabular} \\
\hline
  \begin{tabular}{@{}c@{}} OfficeHome \end{tabular} &
\begin{tabular}{@{}c@{}} Art (A) \\ Clipart (C) \\ Product (P) \\ Real (R) \end{tabular} &
\begin{tabular}{@{}c@{}} 1941 \\ 3492 \\ 3551 \\ 3485 \end{tabular} &
\begin{tabular}{@{}c@{}} 486 \\ 873 \\ 888 \\ 872 \end{tabular} \\
\arrayrulecolor{black}\bottomrule
\end{tabular}
\caption{The size of the train/val split for each domain.}
\label{supp_datasets_table}
\end{table}
\begin{table}[h]
\centering
\begin{tabularx}{\columnwidth}{>{\raggedright\arraybackslash\hsize=.5\hsize}X|>{\raggedright\arraybackslash\hsize=1.5\hsize}X}
\toprule
\multicolumn{1}{c}{Method} & \multicolumn{1}{c}{Description} \\
\midrule
full oracle & accuracy on all target data \\
\arrayrulecolor{gray}\hline
subset oracle & accuracy on a subset of target data \\
\hline
consistency + oracle & cluster / pseudo-label consistency for early stopping, but oracle for hyperparameter tuning \\
\hline
src accuracy &  accuracy on the source data \\
\hline
\nohyphens{src accuracy + loss} & accuracy on the source data plus a loss measuring distance between source and target features \\
\hline
\nohyphens{target entropy} & entropy of predictions in the target domain \\
\hline
\nohyphens{reverse validation} & see main paper for explanation \\
\hline
IWCV & importance weighted cross validation \\
\hline
DEV & see main paper for explanation \\
\bottomrule
\end{tabularx}
\caption{A description of the validation methods we found in papers and code repos.}
\label{supp_meta_analysis_types_of_validation_method}
\end{table}
\begin{table}
\centering
\begin{tabular}{l  l  l}
\toprule
 & Layers & Feature name \\
 \midrule
  \begin{tabular}{@{}l@{}} Trunk \end{tabular} &
 \begin{tabular}{@{}l@{}} \texttt{LeNet} or \texttt{ResNet50} \end{tabular} & 
  \begin{tabular}{@{}l@{}} \texttt{FL0} \end{tabular} \\
 \arrayrulecolor{gray}\hline
 \begin{tabular}{@{}l@{}} Classifier \end{tabular} &
 \begin{tabular}{@{}l@{}} \texttt{Linear(256)} \\ \texttt{ReLU()} \\ \texttt{Dropout(0.5)} \\ \texttt{Linear(128)} \\ \texttt{ReLU()} \\ \texttt{Dropout(0.5)} \\ \texttt{Linear(num\_cls)} \\ \texttt{Softmax()} \end{tabular} & 
  \begin{tabular}{@{}l@{}} \\ \\ \\ \\ \\ \texttt{FL6} \\ \\ \texttt{FL8} \end{tabular} \\
 \hline
 \begin{tabular}{@{}l@{}} Discriminator \end{tabular} &
 \begin{tabular}{@{}l@{}} \texttt{Linear(2048)} \\ \texttt{ReLU()} \\ \texttt{Linear(2048)} \\ \texttt{ReLU()} \\ \texttt{Linear(1)} \end{tabular} &
 \\
 \arrayrulecolor{black}\bottomrule
\end{tabular}
\caption{The models used in our experiments. Two classifiers are used for MCD, STAR, SWD, and SymNets; one is pretrained and the other is randomly initialized. The depth of the classifier depends on the choice of feature layer. Using feature layer 6 results in the first 6 layers of the classifier moving to the trunk, i.e. the classifier becomes \texttt{Linear(num\_cls)}$\rightarrow$ \texttt{Softmax()}. Using feature layer 8 eliminates the classifier model, so this setting can be used only by certain algorithms. The discriminator is used only for adversarial methods. It receives the feature layer as input, but keeps the same depth regardless of feature layer.}
\label{supp_models_table}
\end{table}

\begin{table}
\centering
\resizebox{\columnwidth}{!}{\begin{tabular}{l l} 
 \toprule
 Category & Settings \\
 \midrule
 \begin{tabular}{@{}l@{}} Optimizer \end{tabular} &
 \begin{tabular}{@{}l@{}} Adam \cite{kingma2014adam} \\ Weight decay of \texttt{1e-4} \\ lr $\in$ \texttt{log([1e-5,0.1])} \end{tabular} \\
 \arrayrulecolor{gray}\hline
  \begin{tabular}{@{}l@{}} LR scheduler \end{tabular} &
 \begin{tabular}{@{}l@{}} One Cycle \cite{onelr2019} \\ 5\% warmup period \\ $\text{lr}_{init} = \text{lr}_{max} / 100$ \\$\text{lr}_{final} = 0$ \\ Cosine annealing \end{tabular} \\
 \hline
 Batch size & 64 source + 64 target \\
 \hline
 \begin{tabular}{@{}l@{}} Epochs / patience / val interval \end{tabular} &
 \begin{tabular}{@{}l@{}} Digits: 100 / 10 / 1 \\ Office31: 2000 / 200 / 10 \\ OfficeHome: 200 / 20 / 2 \end{tabular} \\
 \hline
  \begin{tabular}{@{}l@{}} Training image transforms \end{tabular} &
 \begin{tabular}{@{}l@{}} \texttt{Resize(256)} \\ \texttt{RandomCrop(224)} \\ \texttt{RandomHorizontalFlip()} \\ \texttt{Normalize()} \end{tabular} \\
 \hline
  \begin{tabular}{@{}l@{}} Val/testing image transforms \end{tabular} &
 \begin{tabular}{@{}l@{}} \texttt{Resize(256)} \\ \texttt{CenterCrop(224)} \\ \texttt{Normalize()} \end{tabular} \\
 \hline
   \begin{tabular}{@{}l@{}} MNIST image transforms \end{tabular} &
 \begin{tabular}{@{}l@{}} \texttt{Resize(32)} \\ \texttt{GrayscaleToRGB()} \\ \texttt{Normalize()} \end{tabular} \\
 \arrayrulecolor{black}\bottomrule
\end{tabular}}
\caption{Various experiment settings. The learning rate (lr) is one of the hyperparameters, and the same lr is used by trunk, classifier, and discriminator.}
\label{supp_various_experiment_settings}
\end{table}

\begin{table}
\centering
\resizebox{\columnwidth}{!}{\begin{tabular}{c c c} 
 \toprule
 Algorithm & Hyperparameter & Search space \\ [0.5ex] 
 \midrule
\begin{tabular}{@{}c@{}} ADDA \end{tabular} &
\begin{tabular}{@{}c@{}} $\lambda_D$ \\ $\lambda_G$ \\ $T_{adda}$ \end{tabular} &
\begin{tabular}{@{}c@{}} \texttt{[0,1]} \\ \texttt{[0,1]} \\ \texttt{[0,1]} \end{tabular} \\
 \arrayrulecolor{gray}\hline
  \begin{tabular}{@{}c@{}} AFN \end{tabular} &
\begin{tabular}{@{}c@{}} $\lambda_{afn}$ \\ $S_{afn}$ \\ $\lambda_L$ \end{tabular} &
\begin{tabular}{@{}c@{}} \texttt{log([1e-6,1])} \\ \texttt{[0,2]} \\ \texttt{[0,1]} \end{tabular} \\
 \hline
 \begin{tabular}{@{}c@{}} ATDOC \end{tabular} &
\begin{tabular}{@{}c@{}} $\lambda_{atdoc}$ \\ $k_{atdoc}$ \\ $\lambda_L$ \end{tabular} &
\begin{tabular}{@{}c@{}} \texttt{[0,1]} \\ \texttt{int([5, 25], step=5)} \\ \texttt{[0,1]} \end{tabular} \\
 \hline
 \begin{tabular}{@{}c@{}} BNM \end{tabular} &
\begin{tabular}{@{}c@{}} $\lambda_{bnm}$ \\ $\lambda_L$ \end{tabular} &
\begin{tabular}{@{}c@{}} \texttt{[0,1]} \\ \texttt{[0,1]} \end{tabular} \\
\hline
\begin{tabular}{@{}c@{}} BSP \end{tabular} &
\begin{tabular}{@{}c@{}} $\lambda_{bsp}$ \\ $\lambda_L$ \end{tabular} &
\begin{tabular}{@{}c@{}} \texttt{log([1e-6,1])} \\ \texttt{[0,1]} \end{tabular} \\
 \hline
 \begin{tabular}{@{}c@{}} CDAN \\ DC \\ \end{tabular} &
\begin{tabular}{@{}c@{}} $\lambda_D$ \\ $\lambda_G$ \\ $\lambda_L$ \end{tabular} &
\begin{tabular}{@{}c@{}} \texttt{[0,1]} \\ \texttt{[0,1]} \\ \texttt{[0,1]} \end{tabular} \\
 \hline
\begin{tabular}{@{}c@{}} CORAL \end{tabular} &
\begin{tabular}{@{}c@{}} $\lambda_{F}$ \\ $\lambda_L$ \end{tabular} &
\begin{tabular}{@{}c@{}} \texttt{[0,1]} \\ \texttt{[0,1]} \end{tabular} \\
 \hline
 \begin{tabular}{@{}c@{}} DANN \end{tabular} &
\begin{tabular}{@{}c@{}} $\lambda_{D}$ \\ $\lambda_{grl}$ \\ $\lambda_{L}$  \end{tabular} &
\begin{tabular}{@{}c@{}} \texttt{[0,1]} \\ \texttt{log([0.1,10])} \\ \texttt{[0,1]} \end{tabular} \\
 \hline
 \begin{tabular}{@{}c@{}} GVB \end{tabular} &
\begin{tabular}{@{}c@{}} $\lambda_{D}$ \\ $\lambda_{B_G}$ \\ $\lambda_{B_D}$ \\ $\lambda_{grl}$ \end{tabular} &
\begin{tabular}{@{}c@{}} \texttt{[0,1]} \\ \texttt{[0,1]} \\ \texttt{[0,1]} \\ \texttt{log([0.1,10])} \end{tabular} \\
 \hline
\begin{tabular}{@{}c@{}} IM \end{tabular} &
\begin{tabular}{@{}c@{}} $\lambda_{imax}$ \\$\lambda_{L}$  \end{tabular} &
\begin{tabular}{@{}c@{}} \texttt{[0,1]} \\ \texttt{[0,1]} \end{tabular} \\
 \hline
\begin{tabular}{@{}c@{}} ITL \end{tabular} &
\begin{tabular}{@{}c@{}} $\lambda_{imax}$ \\ $\lambda_{imin}$ \\ $\lambda_{L}$ \end{tabular} &
\begin{tabular}{@{}c@{}} \texttt{[0,1]} \\ \texttt{[0,1]} \\ \texttt{[0,1]} \end{tabular} \\
 \hline
\begin{tabular}{@{}c@{}} JMMD \\ MMD \end{tabular} &
\begin{tabular}{@{}c@{}} $\lambda_F$ \\ $\lambda_L$ \\ $\gamma_{exp}$ \end{tabular} &
\begin{tabular}{@{}c@{}} \texttt{[0,1]} \\ \texttt{[0,1]} \\ \texttt{int([1,8])} \end{tabular} \\
 \hline
 \begin{tabular}{@{}c@{}} MCC \end{tabular} &
\begin{tabular}{@{}c@{}} $\lambda_{mcc}$ \\ $T_{mcc}$ \\ $\lambda_L$ \end{tabular} &
\begin{tabular}{@{}c@{}} \texttt{[0,1]} \\ \texttt{[0.2,5]} \\ \texttt{[0,1]} \end{tabular} \\
\hline
\begin{tabular}{@{}c@{}} MCD \\ STAR \\ SWD \end{tabular} &
\begin{tabular}{@{}c@{}} $N_{mcd}$ \\ $\lambda_{L}$ \\ $\lambda_{disc}$ \\ \end{tabular} &
\begin{tabular}{@{}c@{}} \texttt{int([1,10])} \\ \texttt{[0,1]} \\ \texttt{[0,1]} \\ \end{tabular} \\
 \hline
\begin{tabular}{@{}c@{}} MinEnt \end{tabular} &
\begin{tabular}{@{}c@{}} $\lambda_{ent}$ \\$\lambda_{L}$  \end{tabular} &
\begin{tabular}{@{}c@{}} \texttt{[0,1]} \\ \texttt{[0,1]} \end{tabular} \\
 \hline
\begin{tabular}{@{}c@{}} RTN \end{tabular} &
\begin{tabular}{@{}c@{}} $\lambda_F$ \\ $\lambda_L$ \\ $\lambda_{ent}$ \end{tabular} &
\begin{tabular}{@{}c@{}} \texttt{[0,1]} \\ \texttt{[0,1]} \\ \texttt{[0,1]} \end{tabular} \\
 \hline
\begin{tabular}{@{}c@{}} SymNets \end{tabular} &
\begin{tabular}{@{}c@{}} $\lambda_{Sym_{D}}$ \\ $\lambda_{Sym_{C}}$ \\ $\lambda_{Sym_{conf}}$ \\ $\lambda_{Sym_{ent}}$ \end{tabular} &
\begin{tabular}{@{}c@{}} \texttt{[0,1]} \\ \texttt{[0,1]} \\ \texttt{[0,1]} \\ \texttt{[0,1]} \end{tabular} \\
 \hline
\begin{tabular}{@{}c@{}} VADA \end{tabular} &
\begin{tabular}{@{}c@{}} $\lambda_D$ \\ $\lambda_G$ \\ $\lambda_{V_s}$ \\ $\lambda_{V_t}$ \end{tabular} &
\begin{tabular}{@{}c@{}} \texttt{[0,1]} \\ \texttt{[0,1]} \\ \texttt{[0,1]} \\ \texttt{[0,1]} \end{tabular} \\

 \arrayrulecolor{black}\bottomrule
\end{tabular}}
\caption{Hyperparameter search settings.}
\label{supp_hyperparameter_search}
\end{table}

\begin{table}
\centering
\begin{tabularx}{\columnwidth}{>{\raggedright\arraybackslash\hsize=.6\hsize}X|>{\raggedright\arraybackslash\hsize=1.4\hsize}X}
\toprule
Hyperparameter & Description \\
\midrule
$\lambda_{afn}$ & AFN loss weight \\
\hline
$\lambda_{atdoc}$ & ATDOC loss weight \\
\hline
$\lambda_{bnm}$ & BNM loss weight \\
\hline
$\lambda_{bsp}$ & BSP loss weight \\
\hline
$\lambda_{disc}$ & Classifier discrepancy loss weight for MCD \\
\hline
$\lambda_{ent}$ & Target entropy loss weight \\
\hline
$\lambda_{grl}$ & Gradient reversal weight, i.e. gradients are multiplied by $-\lambda_{grl}$ \\
\hline
$\lambda_{imax}$ & Information maximization loss weight \\
\hline
$\lambda_{imin}$ & Information minimization loss weight \\
\hline
$\lambda_{mcc}$ & MCC loss weight \\
\hline
$\lambda_{B_G}$ & Generator bridge loss weight for GVB \\
\hline
$\lambda_{B_D}$ & Discriminator bridge loss weight for GVB \\
\hline
$\lambda_D$ & Discriminator loss weight\\
\arrayrulecolor{gray}\hline
$\lambda_{F}$ & Feature distance loss weight \\
\hline
$\lambda_G$ & Generator loss weight\\
\hline
$\lambda_{L}$ & Source classification loss weight \\
\hline
$\lambda_{Sym_{D}}$ & SymNets classifier domain loss weight \\
\hline 
$\lambda_{Sym_{C}}$ & SymNets generator category loss weight \\
\hline 
$\lambda_{Sym_{conf}}$ & SymNets generator domain loss weight  \\
\hline 
$\lambda_{Sym_{ent}}$ & SymNets entropy loss weight \\
\hline
$\lambda_{V_s}$ & VAT loss weight for the source domain \\
\hline
$\lambda_{V_t}$ & VAT loss weight and entropy weight for the target domain \\
\hline
$\gamma_{exp}$ & Exponent of the bandwidth multiplier for MMD. For example, if $\gamma_{exp}=2$, then the bandwidths used will be $\{2^{-2}x, 2^{-1}x, 2^{0}x, 2^{1}x, 2^{2}x\}$, where $x$ is the base bandwidth. \\
\hline
$k_{atdoc}$ & Number of nearest neighbors to retrieve for computing pseudolabels in ATDOC \\
\hline
$N_{mcd}$ & Number of times the MCD generator is updated per batch \\
\hline
$S_{afn}$ & Step size used by the AFN loss function \\
\hline
$T_{adda}$ & Minimum discriminator accuracy required to trigger a generator update in ADDA \\
\hline
$T_{mcc}$ & Softmax temperature used by MCC \\
\arrayrulecolor{black}\bottomrule
\end{tabularx}
\caption{Description of every hyperparameter in Table \ref{supp_hyperparameter_search}.}
\label{supp_hyperparameter_description}
\end{table}
\begin{table}
\centering
\begin{tabular}{lrrr|r}
\toprule
Task &   IM &  DEV &  SND &  Total \\
\midrule
MM    &   82 &   65 &   78 &    225 \\
AD    &   61 &   48 &   62 &    171 \\
AW    &   64 &   32 &   65 &    161 \\
DA    &   48 &   17 &   36 &    101 \\
DW    &   54 &   25 &   57 &    136 \\
WA    &   49 &   33 &   49 &    131 \\
WD    &   50 &   47 &   51 &    148 \\
AP    &   23 &   20 &   17 &     60 \\
CR    &   20 &   15 &   14 &     49 \\
PA    &   49 &   22 &   56 &    127 \\
RC    &   17 &   19 &   15 &     51 \\
\hline
Total &  517 &  343 &  500 &   1360 \\
\bottomrule
\end{tabular}
\caption{Number of datapoints (thousands) collected per validator/task pair.}
\label{supp_num_datapoints}
\end{table}

\def\officethirtyonestdAD#1{\ifdim#1pt<0.5pt\cellcolor{lime!100}\else\ifdim#1pt<1.0pt\cellcolor{lime!90}\else\ifdim#1pt<1.5pt\cellcolor{lime!80}\else\ifdim#1pt<2.0pt\cellcolor{lime!70}\else\ifdim#1pt<2.5pt\cellcolor{lime!60}\else\ifdim#1pt<2.9pt\cellcolor{lime!50}\else\ifdim#1pt<3.4pt\cellcolor{lime!40}\else\ifdim#1pt<3.9pt\cellcolor{lime!30}\else\ifdim#1pt<4.4pt\cellcolor{lime!20}\else\ifdim#1pt<4.9pt\cellcolor{lime!10}\else\cellcolor{lime!0}\fi\fi\fi\fi\fi\fi\fi\fi\fi\fi#1}

\def\officethirtyonestdAW#1{\ifdim#1pt<0.3pt\cellcolor{lime!100}\else\ifdim#1pt<0.7pt\cellcolor{lime!90}\else\ifdim#1pt<1.0pt\cellcolor{lime!80}\else\ifdim#1pt<1.4pt\cellcolor{lime!70}\else\ifdim#1pt<1.7pt\cellcolor{lime!60}\else\ifdim#1pt<2.0pt\cellcolor{lime!50}\else\ifdim#1pt<2.4pt\cellcolor{lime!40}\else\ifdim#1pt<2.7pt\cellcolor{lime!30}\else\ifdim#1pt<3.1pt\cellcolor{lime!20}\else\ifdim#1pt<3.4pt\cellcolor{lime!10}\else\cellcolor{lime!0}\fi\fi\fi\fi\fi\fi\fi\fi\fi\fi#1}

\def\officethirtyonestdDA#1{\ifdim#1pt<0.3pt\cellcolor{lime!100}\else\ifdim#1pt<0.5pt\cellcolor{lime!90}\else\ifdim#1pt<0.8pt\cellcolor{lime!80}\else\ifdim#1pt<1.1pt\cellcolor{lime!70}\else\ifdim#1pt<1.4pt\cellcolor{lime!60}\else\ifdim#1pt<1.6pt\cellcolor{lime!50}\else\ifdim#1pt<1.9pt\cellcolor{lime!40}\else\ifdim#1pt<2.2pt\cellcolor{lime!30}\else\ifdim#1pt<2.4pt\cellcolor{lime!20}\else\ifdim#1pt<2.7pt\cellcolor{lime!10}\else\cellcolor{lime!0}\fi\fi\fi\fi\fi\fi\fi\fi\fi\fi#1}

\def\officethirtyonestdDW#1{\ifdim#1pt<0.2pt\cellcolor{lime!100}\else\ifdim#1pt<0.4pt\cellcolor{lime!90}\else\ifdim#1pt<0.6pt\cellcolor{lime!80}\else\ifdim#1pt<0.8pt\cellcolor{lime!70}\else\ifdim#1pt<1.1pt\cellcolor{lime!60}\else\ifdim#1pt<1.3pt\cellcolor{lime!50}\else\ifdim#1pt<1.5pt\cellcolor{lime!40}\else\ifdim#1pt<1.7pt\cellcolor{lime!30}\else\ifdim#1pt<1.9pt\cellcolor{lime!20}\else\ifdim#1pt<2.1pt\cellcolor{lime!10}\else\cellcolor{lime!0}\fi\fi\fi\fi\fi\fi\fi\fi\fi\fi#1}

\def\officethirtyonestdWA#1{\ifdim#1pt<0.1pt\cellcolor{lime!100}\else\ifdim#1pt<0.3pt\cellcolor{lime!90}\else\ifdim#1pt<0.4pt\cellcolor{lime!80}\else\ifdim#1pt<0.6pt\cellcolor{lime!70}\else\ifdim#1pt<0.8pt\cellcolor{lime!60}\else\ifdim#1pt<0.9pt\cellcolor{lime!50}\else\ifdim#1pt<1.1pt\cellcolor{lime!40}\else\ifdim#1pt<1.2pt\cellcolor{lime!30}\else\ifdim#1pt<1.3pt\cellcolor{lime!20}\else\ifdim#1pt<1.5pt\cellcolor{lime!10}\else\cellcolor{lime!0}\fi\fi\fi\fi\fi\fi\fi\fi\fi\fi#1}

\def\officethirtyonestdWD#1{\ifdim#1pt<0.3pt\cellcolor{lime!100}\else\ifdim#1pt<0.6pt\cellcolor{lime!90}\else\ifdim#1pt<0.9pt\cellcolor{lime!80}\else\ifdim#1pt<1.2pt\cellcolor{lime!70}\else\ifdim#1pt<1.5pt\cellcolor{lime!60}\else\ifdim#1pt<1.8pt\cellcolor{lime!50}\else\ifdim#1pt<2.1pt\cellcolor{lime!40}\else\ifdim#1pt<2.4pt\cellcolor{lime!30}\else\ifdim#1pt<2.7pt\cellcolor{lime!20}\else\ifdim#1pt<3.0pt\cellcolor{lime!10}\else\cellcolor{lime!0}\fi\fi\fi\fi\fi\fi\fi\fi\fi\fi#1}

\def\officethirtyonestdAvg#1{\ifdim#1pt<0.2pt\cellcolor{lime!100}\else\ifdim#1pt<0.4pt\cellcolor{lime!90}\else\ifdim#1pt<0.5pt\cellcolor{lime!80}\else\ifdim#1pt<0.7pt\cellcolor{lime!70}\else\ifdim#1pt<0.9pt\cellcolor{lime!60}\else\ifdim#1pt<1.1pt\cellcolor{lime!50}\else\ifdim#1pt<1.3pt\cellcolor{lime!40}\else\ifdim#1pt<1.4pt\cellcolor{lime!30}\else\ifdim#1pt<1.6pt\cellcolor{lime!20}\else\ifdim#1pt<1.8pt\cellcolor{lime!10}\else\cellcolor{lime!0}\fi\fi\fi\fi\fi\fi\fi\fi\fi\fi#1}

\begin{table}
\centering
\resizebox{\columnwidth}{!}{\begin{tabular}{lllllll|l}
\toprule
{} & {AD} & {AW} & {DA} & {DW} & {WA} & {WD} & {Avg} \\
\midrule
ADDA & \officethirtyonestdAD{2.1} & \officethirtyonestdAW{1.8} & \officethirtyonestdDA{0.6} & \officethirtyonestdDW{0.7} & \officethirtyonestdWA{1.1} & \officethirtyonestdWD{3.0} & \officethirtyonestdAvg{1.6} \\
AFN & \officethirtyonestdAD{2.9} & \officethirtyonestdAW{2.5} & \officethirtyonestdDA{0.7} & \officethirtyonestdDW{0.9} & \officethirtyonestdWA{0.4} & \officethirtyonestdWD{0.6} & \officethirtyonestdAvg{1.3} \\
AFN-DANN & \officethirtyonestdAD{2.0} & \officethirtyonestdAW{1.9} & \officethirtyonestdDA{0.7} & \officethirtyonestdDW{0.7} & \officethirtyonestdWA{0.9} & \officethirtyonestdWD{0.6} & \officethirtyonestdAvg{1.1} \\
ATDOC & \officethirtyonestdAD{3.3} & \officethirtyonestdAW{1.3} & \officethirtyonestdDA{1.0} & \officethirtyonestdDW{1.5} & \officethirtyonestdWA{0.5} & \officethirtyonestdWD{0.5} & \officethirtyonestdAvg{1.3} \\
ATDOC-DANN & \officethirtyonestdAD{1.0} & \officethirtyonestdAW{2.0} & \officethirtyonestdDA{0.5} & \officethirtyonestdDW{0.7} & \officethirtyonestdWA{0.4} & \officethirtyonestdWD{1.3} & \officethirtyonestdAvg{1.0} \\
BNM & \officethirtyonestdAD{1.2} & \officethirtyonestdAW{2.0} & \officethirtyonestdDA{1.0} & \officethirtyonestdDW{0.5} & \textbf{\officethirtyonestdWA{0.3}} & \textbf{\officethirtyonestdWD{0.0}} & \officethirtyonestdAvg{0.8} \\
BNM-DANN & \officethirtyonestdAD{2.2} & \officethirtyonestdAW{2.4} & \officethirtyonestdDA{0.7} & \officethirtyonestdDW{0.4} & \officethirtyonestdWA{0.8} & \officethirtyonestdWD{0.6} & \officethirtyonestdAvg{1.2} \\
BSP & \officethirtyonestdAD{2.9} & \officethirtyonestdAW{1.0} & \officethirtyonestdDA{0.7} & \officethirtyonestdDW{0.4} & \officethirtyonestdWA{0.8} & \officethirtyonestdWD{0.5} & \officethirtyonestdAvg{1.1} \\
BSP-DANN & \officethirtyonestdAD{2.0} & \officethirtyonestdAW{1.3} & \officethirtyonestdDA{0.5} & \textbf{\officethirtyonestdDW{0.0}} & \textbf{\officethirtyonestdWA{0.3}} & \officethirtyonestdWD{0.6} & \officethirtyonestdAvg{0.8} \\
CDAN & \textbf{\officethirtyonestdAD{0.7}} & \officethirtyonestdAW{2.2} & \officethirtyonestdDA{0.6} & \officethirtyonestdDW{0.9} & \textbf{\officethirtyonestdWA{0.3}} & \officethirtyonestdWD{0.8} & \officethirtyonestdAvg{0.9} \\
CORAL & \officethirtyonestdAD{1.2} & \officethirtyonestdAW{1.3} & \officethirtyonestdDA{0.9} & \officethirtyonestdDW{1.7} & \officethirtyonestdWA{0.5} & \officethirtyonestdWD{2.3} & \officethirtyonestdAvg{1.3} \\
DANN & \officethirtyonestdAD{0.9} & \officethirtyonestdAW{1.0} & \officethirtyonestdDA{1.0} & \officethirtyonestdDW{0.5} & \officethirtyonestdWA{0.6} & \officethirtyonestdWD{0.6} & \textbf{\officethirtyonestdAvg{0.7}} \\
DANN-FL8 & \officethirtyonestdAD{1.6} & \officethirtyonestdAW{1.1} & \textbf{\officethirtyonestdDA{0.4}} & \officethirtyonestdDW{0.3} & \officethirtyonestdWA{0.6} & \officethirtyonestdWD{0.6} & \officethirtyonestdAvg{0.8} \\
DC & \officethirtyonestdAD{4.2} & \officethirtyonestdAW{1.3} & \officethirtyonestdDA{0.8} & \officethirtyonestdDW{1.6} & \officethirtyonestdWA{0.7} & \officethirtyonestdWD{0.6} & \officethirtyonestdAvg{1.5} \\
GVB & \textbf{\officethirtyonestdAD{0.7}} & \officethirtyonestdAW{2.0} & \officethirtyonestdDA{1.2} & \officethirtyonestdDW{1.6} & \officethirtyonestdWA{0.7} & \officethirtyonestdWD{0.5} & \officethirtyonestdAvg{1.1} \\
IM & \officethirtyonestdAD{1.5} & \officethirtyonestdAW{2.0} & \officethirtyonestdDA{0.5} & \officethirtyonestdDW{1.8} & \officethirtyonestdWA{0.8} & \officethirtyonestdWD{0.8} & \officethirtyonestdAvg{1.2} \\
IM-DANN & \officethirtyonestdAD{2.7} & \officethirtyonestdAW{1.9} & \officethirtyonestdDA{0.7} & \officethirtyonestdDW{0.6} & \officethirtyonestdWA{0.8} & \officethirtyonestdWD{0.5} & \officethirtyonestdAvg{1.2} \\
ITL & \officethirtyonestdAD{2.2} & \officethirtyonestdAW{1.4} & \officethirtyonestdDA{0.8} & \officethirtyonestdDW{0.9} & \officethirtyonestdWA{0.5} & \officethirtyonestdWD{0.6} & \officethirtyonestdAvg{1.1} \\
JMMD & \officethirtyonestdAD{1.9} & \officethirtyonestdAW{1.6} & \officethirtyonestdDA{1.0} & \officethirtyonestdDW{0.9} & \officethirtyonestdWA{0.9} & \officethirtyonestdWD{0.5} & \officethirtyonestdAvg{1.1} \\
MCC & \officethirtyonestdAD{0.8} & \officethirtyonestdAW{2.3} & \officethirtyonestdDA{0.7} & \officethirtyonestdDW{0.9} & \textbf{\officethirtyonestdWA{0.3}} & \officethirtyonestdWD{0.6} & \officethirtyonestdAvg{0.9} \\
MCC-DANN & \textbf{\officethirtyonestdAD{0.7}} & \officethirtyonestdAW{2.3} & \officethirtyonestdDA{0.9} & \officethirtyonestdDW{0.6} & \officethirtyonestdWA{0.7} & \officethirtyonestdWD{0.6} & \officethirtyonestdAvg{1.0} \\
MCD & \officethirtyonestdAD{2.9} & \officethirtyonestdAW{2.0} & \officethirtyonestdDA{0.6} & \officethirtyonestdDW{1.2} & \textbf{\officethirtyonestdWA{0.3}} & \officethirtyonestdWD{1.0} & \officethirtyonestdAvg{1.3} \\
MMD & \officethirtyonestdAD{1.6} & \officethirtyonestdAW{2.6} & \officethirtyonestdDA{0.6} & \officethirtyonestdDW{0.3} & \officethirtyonestdWA{1.1} & \officethirtyonestdWD{0.6} & \officethirtyonestdAvg{1.1} \\
MinEnt & \officethirtyonestdAD{4.2} & \officethirtyonestdAW{2.9} & \officethirtyonestdDA{1.0} & \officethirtyonestdDW{0.5} & \officethirtyonestdWA{0.7} & \officethirtyonestdWD{0.4} & \officethirtyonestdAvg{1.6} \\
RTN & \officethirtyonestdAD{3.1} & \textbf{\officethirtyonestdAW{0.6}} & \officethirtyonestdDA{0.9} & \officethirtyonestdDW{0.6} & \officethirtyonestdWA{1.0} & \officethirtyonestdWD{1.2} & \officethirtyonestdAvg{1.3} \\
STAR & \officethirtyonestdAD{2.9} & \officethirtyonestdAW{0.8} & \officethirtyonestdDA{1.3} & \officethirtyonestdDW{2.1} & \officethirtyonestdWA{0.5} & \officethirtyonestdWD{0.8} & \officethirtyonestdAvg{1.4} \\
SWD & \officethirtyonestdAD{4.9} & \officethirtyonestdAW{0.9} & \officethirtyonestdDA{0.6} & \officethirtyonestdDW{0.9} & \officethirtyonestdWA{0.7} & \officethirtyonestdWD{2.9} & \officethirtyonestdAvg{1.8} \\
SymNets & \officethirtyonestdAD{2.8} & \officethirtyonestdAW{1.3} & \officethirtyonestdDA{2.7} & \officethirtyonestdDW{1.2} & \officethirtyonestdWA{1.5} & \officethirtyonestdWD{0.6} & \officethirtyonestdAvg{1.7} \\
VADA & - & \officethirtyonestdAW{3.4} & \officethirtyonestdDA{0.5} & \officethirtyonestdDW{0.5} & \officethirtyonestdWA{0.5} & \officethirtyonestdWD{0.3} & \officethirtyonestdAvg{1.1} \\
\bottomrule
\end{tabular}}
\caption{Standard deviation on Office31.}
\label{supp_office31_oracle_results_std}
\end{table}

\def\mnistofficehomestdMM#1{\ifdim#1pt<2.5pt\cellcolor{lime!100}\else\ifdim#1pt<4.9pt\cellcolor{lime!90}\else\ifdim#1pt<7.4pt\cellcolor{lime!80}\else\ifdim#1pt<9.9pt\cellcolor{lime!70}\else\ifdim#1pt<12.3pt\cellcolor{lime!60}\else\ifdim#1pt<14.8pt\cellcolor{lime!50}\else\ifdim#1pt<17.3pt\cellcolor{lime!40}\else\ifdim#1pt<19.8pt\cellcolor{lime!30}\else\ifdim#1pt<22.2pt\cellcolor{lime!20}\else\ifdim#1pt<24.7pt\cellcolor{lime!10}\else\cellcolor{lime!0}\fi\fi\fi\fi\fi\fi\fi\fi\fi\fi#1}

\def\mnistofficehomestdAC#1{\ifdim#1pt<0.2pt\cellcolor{lime!100}\else\ifdim#1pt<0.4pt\cellcolor{lime!90}\else\ifdim#1pt<0.6pt\cellcolor{lime!80}\else\ifdim#1pt<0.8pt\cellcolor{lime!70}\else\ifdim#1pt<0.9pt\cellcolor{lime!60}\else\ifdim#1pt<1.1pt\cellcolor{lime!50}\else\ifdim#1pt<1.3pt\cellcolor{lime!40}\else\ifdim#1pt<1.5pt\cellcolor{lime!30}\else\ifdim#1pt<1.7pt\cellcolor{lime!20}\else\ifdim#1pt<1.9pt\cellcolor{lime!10}\else\cellcolor{lime!0}\fi\fi\fi\fi\fi\fi\fi\fi\fi\fi#1}

\def\mnistofficehomestdAP#1{\ifdim#1pt<0.1pt\cellcolor{lime!100}\else\ifdim#1pt<0.2pt\cellcolor{lime!90}\else\ifdim#1pt<0.4pt\cellcolor{lime!80}\else\ifdim#1pt<0.5pt\cellcolor{lime!70}\else\ifdim#1pt<0.6pt\cellcolor{lime!60}\else\ifdim#1pt<0.7pt\cellcolor{lime!50}\else\ifdim#1pt<0.8pt\cellcolor{lime!40}\else\ifdim#1pt<1.0pt\cellcolor{lime!30}\else\ifdim#1pt<1.1pt\cellcolor{lime!20}\else\ifdim#1pt<1.2pt\cellcolor{lime!10}\else\cellcolor{lime!0}\fi\fi\fi\fi\fi\fi\fi\fi\fi\fi#1}

\def\mnistofficehomestdAR#1{\ifdim#1pt<0.1pt\cellcolor{lime!100}\else\ifdim#1pt<0.2pt\cellcolor{lime!90}\else\ifdim#1pt<0.3pt\cellcolor{lime!80}\else\ifdim#1pt<0.4pt\cellcolor{lime!70}\else\ifdim#1pt<0.6pt\cellcolor{lime!60}\else\ifdim#1pt<0.7pt\cellcolor{lime!50}\else\ifdim#1pt<0.8pt\cellcolor{lime!40}\else\ifdim#1pt<0.9pt\cellcolor{lime!30}\else\ifdim#1pt<1.0pt\cellcolor{lime!20}\else\ifdim#1pt<1.1pt\cellcolor{lime!10}\else\cellcolor{lime!0}\fi\fi\fi\fi\fi\fi\fi\fi\fi\fi#1}

\def\mnistofficehomestdCA#1{\ifdim#1pt<0.2pt\cellcolor{lime!100}\else\ifdim#1pt<0.3pt\cellcolor{lime!90}\else\ifdim#1pt<0.5pt\cellcolor{lime!80}\else\ifdim#1pt<0.6pt\cellcolor{lime!70}\else\ifdim#1pt<0.8pt\cellcolor{lime!60}\else\ifdim#1pt<1.0pt\cellcolor{lime!50}\else\ifdim#1pt<1.1pt\cellcolor{lime!40}\else\ifdim#1pt<1.3pt\cellcolor{lime!30}\else\ifdim#1pt<1.4pt\cellcolor{lime!20}\else\ifdim#1pt<1.6pt\cellcolor{lime!10}\else\cellcolor{lime!0}\fi\fi\fi\fi\fi\fi\fi\fi\fi\fi#1}

\def\mnistofficehomestdCP#1{\ifdim#1pt<0.4pt\cellcolor{lime!100}\else\ifdim#1pt<0.7pt\cellcolor{lime!90}\else\ifdim#1pt<1.1pt\cellcolor{lime!80}\else\ifdim#1pt<1.5pt\cellcolor{lime!70}\else\ifdim#1pt<1.9pt\cellcolor{lime!60}\else\ifdim#1pt<2.2pt\cellcolor{lime!50}\else\ifdim#1pt<2.6pt\cellcolor{lime!40}\else\ifdim#1pt<3.0pt\cellcolor{lime!30}\else\ifdim#1pt<3.3pt\cellcolor{lime!20}\else\ifdim#1pt<3.7pt\cellcolor{lime!10}\else\cellcolor{lime!0}\fi\fi\fi\fi\fi\fi\fi\fi\fi\fi#1}

\def\mnistofficehomestdCR#1{\ifdim#1pt<0.1pt\cellcolor{lime!100}\else\ifdim#1pt<0.3pt\cellcolor{lime!90}\else\ifdim#1pt<0.4pt\cellcolor{lime!80}\else\ifdim#1pt<0.6pt\cellcolor{lime!70}\else\ifdim#1pt<0.8pt\cellcolor{lime!60}\else\ifdim#1pt<0.9pt\cellcolor{lime!50}\else\ifdim#1pt<1.1pt\cellcolor{lime!40}\else\ifdim#1pt<1.2pt\cellcolor{lime!30}\else\ifdim#1pt<1.3pt\cellcolor{lime!20}\else\ifdim#1pt<1.5pt\cellcolor{lime!10}\else\cellcolor{lime!0}\fi\fi\fi\fi\fi\fi\fi\fi\fi\fi#1}

\def\mnistofficehomestdPA#1{\ifdim#1pt<0.2pt\cellcolor{lime!100}\else\ifdim#1pt<0.4pt\cellcolor{lime!90}\else\ifdim#1pt<0.7pt\cellcolor{lime!80}\else\ifdim#1pt<0.9pt\cellcolor{lime!70}\else\ifdim#1pt<1.1pt\cellcolor{lime!60}\else\ifdim#1pt<1.3pt\cellcolor{lime!50}\else\ifdim#1pt<1.5pt\cellcolor{lime!40}\else\ifdim#1pt<1.8pt\cellcolor{lime!30}\else\ifdim#1pt<2.0pt\cellcolor{lime!20}\else\ifdim#1pt<2.2pt\cellcolor{lime!10}\else\cellcolor{lime!0}\fi\fi\fi\fi\fi\fi\fi\fi\fi\fi#1}

\def\mnistofficehomestdPC#1{\ifdim#1pt<0.2pt\cellcolor{lime!100}\else\ifdim#1pt<0.5pt\cellcolor{lime!90}\else\ifdim#1pt<0.7pt\cellcolor{lime!80}\else\ifdim#1pt<0.9pt\cellcolor{lime!70}\else\ifdim#1pt<1.1pt\cellcolor{lime!60}\else\ifdim#1pt<1.4pt\cellcolor{lime!50}\else\ifdim#1pt<1.6pt\cellcolor{lime!40}\else\ifdim#1pt<1.8pt\cellcolor{lime!30}\else\ifdim#1pt<2.1pt\cellcolor{lime!20}\else\ifdim#1pt<2.3pt\cellcolor{lime!10}\else\cellcolor{lime!0}\fi\fi\fi\fi\fi\fi\fi\fi\fi\fi#1}

\def\mnistofficehomestdPR#1{\ifdim#1pt<0.1pt\cellcolor{lime!100}\else\ifdim#1pt<0.2pt\cellcolor{lime!90}\else\ifdim#1pt<0.3pt\cellcolor{lime!80}\else\ifdim#1pt<0.4pt\cellcolor{lime!70}\else\ifdim#1pt<0.6pt\cellcolor{lime!60}\else\ifdim#1pt<0.7pt\cellcolor{lime!50}\else\ifdim#1pt<0.8pt\cellcolor{lime!40}\else\ifdim#1pt<0.9pt\cellcolor{lime!30}\else\ifdim#1pt<1.0pt\cellcolor{lime!20}\else\ifdim#1pt<1.1pt\cellcolor{lime!10}\else\cellcolor{lime!0}\fi\fi\fi\fi\fi\fi\fi\fi\fi\fi#1}

\def\mnistofficehomestdRA#1{\ifdim#1pt<0.2pt\cellcolor{lime!100}\else\ifdim#1pt<0.5pt\cellcolor{lime!90}\else\ifdim#1pt<0.7pt\cellcolor{lime!80}\else\ifdim#1pt<0.9pt\cellcolor{lime!70}\else\ifdim#1pt<1.1pt\cellcolor{lime!60}\else\ifdim#1pt<1.4pt\cellcolor{lime!50}\else\ifdim#1pt<1.6pt\cellcolor{lime!40}\else\ifdim#1pt<1.8pt\cellcolor{lime!30}\else\ifdim#1pt<2.1pt\cellcolor{lime!20}\else\ifdim#1pt<2.3pt\cellcolor{lime!10}\else\cellcolor{lime!0}\fi\fi\fi\fi\fi\fi\fi\fi\fi\fi#1}

\def\mnistofficehomestdRC#1{\ifdim#1pt<0.2pt\cellcolor{lime!100}\else\ifdim#1pt<0.5pt\cellcolor{lime!90}\else\ifdim#1pt<0.7pt\cellcolor{lime!80}\else\ifdim#1pt<1.0pt\cellcolor{lime!70}\else\ifdim#1pt<1.2pt\cellcolor{lime!60}\else\ifdim#1pt<1.4pt\cellcolor{lime!50}\else\ifdim#1pt<1.7pt\cellcolor{lime!40}\else\ifdim#1pt<1.9pt\cellcolor{lime!30}\else\ifdim#1pt<2.2pt\cellcolor{lime!20}\else\ifdim#1pt<2.4pt\cellcolor{lime!10}\else\cellcolor{lime!0}\fi\fi\fi\fi\fi\fi\fi\fi\fi\fi#1}

\def\mnistofficehomestdRP#1{\ifdim#1pt<0.2pt\cellcolor{lime!100}\else\ifdim#1pt<0.3pt\cellcolor{lime!90}\else\ifdim#1pt<0.5pt\cellcolor{lime!80}\else\ifdim#1pt<0.6pt\cellcolor{lime!70}\else\ifdim#1pt<0.8pt\cellcolor{lime!60}\else\ifdim#1pt<1.0pt\cellcolor{lime!50}\else\ifdim#1pt<1.1pt\cellcolor{lime!40}\else\ifdim#1pt<1.3pt\cellcolor{lime!30}\else\ifdim#1pt<1.4pt\cellcolor{lime!20}\else\ifdim#1pt<1.6pt\cellcolor{lime!10}\else\cellcolor{lime!0}\fi\fi\fi\fi\fi\fi\fi\fi\fi\fi#1}

\def\mnistofficehomestdAvg#1{\ifdim#1pt<0.1pt\cellcolor{lime!100}\else\ifdim#1pt<0.3pt\cellcolor{lime!90}\else\ifdim#1pt<0.4pt\cellcolor{lime!80}\else\ifdim#1pt<0.6pt\cellcolor{lime!70}\else\ifdim#1pt<0.7pt\cellcolor{lime!60}\else\ifdim#1pt<0.8pt\cellcolor{lime!50}\else\ifdim#1pt<1.0pt\cellcolor{lime!40}\else\ifdim#1pt<1.1pt\cellcolor{lime!30}\else\ifdim#1pt<1.3pt\cellcolor{lime!20}\else\ifdim#1pt<1.4pt\cellcolor{lime!10}\else\cellcolor{lime!0}\fi\fi\fi\fi\fi\fi\fi\fi\fi\fi#1}

\begin{table*}
\centering
\begin{tabular}{ll||llllllllllll|l}
\toprule
{} & {MM} & {AC} & {AP} & {AR} & {CA} & {CP} & {CR} & {PA} & {PC} & {PR} & {RA} & {RC} & {RP} & {Avg} \\
\midrule
ADDA & \mnistofficehomestdMM{0.4} & \mnistofficehomestdAC{1.0} & \mnistofficehomestdAP{0.8} & \mnistofficehomestdAR{0.4} & \mnistofficehomestdCA{1.1} & \mnistofficehomestdCP{0.8} & \mnistofficehomestdCR{0.6} & \mnistofficehomestdPA{1.1} & \mnistofficehomestdPC{0.9} & \mnistofficehomestdPR{0.8} & \mnistofficehomestdRA{1.3} & \mnistofficehomestdRC{1.5} & \mnistofficehomestdRP{0.6} & \mnistofficehomestdAvg{0.9} \\
AFN & \mnistofficehomestdMM{1.0} & \mnistofficehomestdAC{0.5} & \mnistofficehomestdAP{0.6} & \mnistofficehomestdAR{0.5} & \mnistofficehomestdCA{1.1} & \mnistofficehomestdCP{0.6} & \mnistofficehomestdCR{0.8} & \textbf{\mnistofficehomestdPA{0.2}} & \mnistofficehomestdPC{0.7} & \mnistofficehomestdPR{0.7} & \mnistofficehomestdRA{1.0} & \mnistofficehomestdRC{1.0} & \mnistofficehomestdRP{0.9} & \textbf{\mnistofficehomestdAvg{0.7}} \\
AFN-DANN & \textbf{\mnistofficehomestdMM{0.1}} & \mnistofficehomestdAC{0.7} & \mnistofficehomestdAP{0.7} & \mnistofficehomestdAR{0.4} & \mnistofficehomestdCA{0.7} & \mnistofficehomestdCP{0.9} & \mnistofficehomestdCR{1.2} & \mnistofficehomestdPA{1.7} & \mnistofficehomestdPC{1.0} & \mnistofficehomestdPR{0.7} & \mnistofficehomestdRA{0.5} & \mnistofficehomestdRC{0.9} & \mnistofficehomestdRP{1.4} & \mnistofficehomestdAvg{0.9} \\
ATDOC & \mnistofficehomestdMM{9.3} & \mnistofficehomestdAC{0.8} & \mnistofficehomestdAP{0.9} & \mnistofficehomestdAR{0.7} & \mnistofficehomestdCA{1.3} & \mnistofficehomestdCP{1.3} & \mnistofficehomestdCR{0.5} & \mnistofficehomestdPA{1.0} & \mnistofficehomestdPC{0.5} & \mnistofficehomestdPR{0.4} & \mnistofficehomestdRA{0.7} & \mnistofficehomestdRC{0.8} & \mnistofficehomestdRP{0.9} & \mnistofficehomestdAvg{0.8} \\
ATDOC-DANN & \mnistofficehomestdMM{6.1} & \mnistofficehomestdAC{0.7} & \mnistofficehomestdAP{0.4} & \mnistofficehomestdAR{0.4} & \mnistofficehomestdCA{0.8} & \mnistofficehomestdCP{1.6} & \mnistofficehomestdCR{0.7} & \mnistofficehomestdPA{0.8} & \mnistofficehomestdPC{0.8} & \mnistofficehomestdPR{0.9} & \mnistofficehomestdRA{1.8} & \mnistofficehomestdRC{1.5} & \textbf{\mnistofficehomestdRP{0.3}} & \mnistofficehomestdAvg{0.9} \\
BNM & \mnistofficehomestdMM{0.4} & \mnistofficehomestdAC{0.9} & \mnistofficehomestdAP{0.6} & \mnistofficehomestdAR{0.6} & \mnistofficehomestdCA{1.4} & \mnistofficehomestdCP{0.9} & \mnistofficehomestdCR{1.0} & \mnistofficehomestdPA{0.8} & \mnistofficehomestdPC{0.8} & \mnistofficehomestdPR{0.8} & \mnistofficehomestdRA{1.0} & \mnistofficehomestdRC{1.2} & \mnistofficehomestdRP{0.6} & \mnistofficehomestdAvg{0.9} \\
BNM-DANN & - & \mnistofficehomestdAC{1.1} & \mnistofficehomestdAP{0.8} & \mnistofficehomestdAR{0.4} & \mnistofficehomestdCA{1.3} & \mnistofficehomestdCP{0.9} & \mnistofficehomestdCR{1.0} & \mnistofficehomestdPA{1.2} & \mnistofficehomestdPC{1.3} & \mnistofficehomestdPR{0.8} & \mnistofficehomestdRA{1.3} & \mnistofficehomestdRC{1.5} & \mnistofficehomestdRP{1.0} & \mnistofficehomestdAvg{1.0} \\
BSP & \textbf{\mnistofficehomestdMM{0.1}} & \mnistofficehomestdAC{0.8} & \textbf{\mnistofficehomestdAP{0.2}} & \mnistofficehomestdAR{0.6} & \mnistofficehomestdCA{1.3} & \mnistofficehomestdCP{0.6} & \mnistofficehomestdCR{0.9} & \mnistofficehomestdPA{0.8} & \textbf{\mnistofficehomestdPC{0.3}} & \textbf{\mnistofficehomestdPR{0.3}} & \mnistofficehomestdRA{1.1} & \mnistofficehomestdRC{0.9} & \mnistofficehomestdRP{0.7} & \textbf{\mnistofficehomestdAvg{0.7}} \\
BSP-DANN & - & \mnistofficehomestdAC{0.6} & \mnistofficehomestdAP{0.8} & \mnistofficehomestdAR{0.6} & \mnistofficehomestdCA{1.5} & \mnistofficehomestdCP{1.4} & \mnistofficehomestdCR{0.8} & \mnistofficehomestdPA{0.4} & \mnistofficehomestdPC{1.3} & \mnistofficehomestdPR{0.6} & \mnistofficehomestdRA{1.0} & \mnistofficehomestdRC{2.4} & \mnistofficehomestdRP{0.6} & \mnistofficehomestdAvg{1.0} \\
CDAN & \mnistofficehomestdMM{6.3} & \mnistofficehomestdAC{1.5} & \mnistofficehomestdAP{0.6} & \mnistofficehomestdAR{0.6} & \mnistofficehomestdCA{1.0} & \textbf{\mnistofficehomestdCP{0.2}} & \mnistofficehomestdCR{0.5} & \mnistofficehomestdPA{0.6} & \mnistofficehomestdPC{2.3} & \mnistofficehomestdPR{0.5} & \mnistofficehomestdRA{0.9} & \mnistofficehomestdRC{1.3} & \mnistofficehomestdRP{0.4} & \mnistofficehomestdAvg{0.9} \\
CORAL & \mnistofficehomestdMM{1.5} & \mnistofficehomestdAC{0.4} & \mnistofficehomestdAP{0.6} & \mnistofficehomestdAR{0.4} & \mnistofficehomestdCA{1.2} & \mnistofficehomestdCP{0.5} & \mnistofficehomestdCR{0.6} & \mnistofficehomestdPA{0.7} & \mnistofficehomestdPC{0.6} & \mnistofficehomestdPR{0.5} & \mnistofficehomestdRA{1.1} & \mnistofficehomestdRC{1.3} & \mnistofficehomestdRP{0.6} & \textbf{\mnistofficehomestdAvg{0.7}} \\
DANN & \mnistofficehomestdMM{0.5} & \mnistofficehomestdAC{0.5} & \mnistofficehomestdAP{0.4} & \mnistofficehomestdAR{0.8} & \mnistofficehomestdCA{1.0} & \mnistofficehomestdCP{1.6} & \textbf{\mnistofficehomestdCR{0.4}} & \mnistofficehomestdPA{0.8} & \mnistofficehomestdPC{1.7} & \mnistofficehomestdPR{1.1} & \mnistofficehomestdRA{2.3} & \mnistofficehomestdRC{0.9} & \mnistofficehomestdRP{1.6} & \mnistofficehomestdAvg{1.1} \\
DANN-FL8 & \mnistofficehomestdMM{8.7} & \mnistofficehomestdAC{1.0} & \mnistofficehomestdAP{0.9} & \mnistofficehomestdAR{0.6} & \mnistofficehomestdCA{0.6} & \mnistofficehomestdCP{0.6} & \mnistofficehomestdCR{1.0} & \mnistofficehomestdPA{0.7} & \mnistofficehomestdPC{1.6} & \mnistofficehomestdPR{1.1} & \mnistofficehomestdRA{0.8} & \mnistofficehomestdRC{0.7} & \mnistofficehomestdRP{0.6} & \mnistofficehomestdAvg{0.9} \\
DC & \mnistofficehomestdMM{2.6} & \mnistofficehomestdAC{1.4} & \mnistofficehomestdAP{0.8} & \mnistofficehomestdAR{0.4} & \textbf{\mnistofficehomestdCA{0.4}} & \mnistofficehomestdCP{1.2} & \mnistofficehomestdCR{0.6} & \mnistofficehomestdPA{1.7} & \mnistofficehomestdPC{1.3} & \mnistofficehomestdPR{0.7} & \mnistofficehomestdRA{1.1} & \mnistofficehomestdRC{1.2} & \mnistofficehomestdRP{0.7} & \mnistofficehomestdAvg{1.0} \\
GVB & \mnistofficehomestdMM{5.4} & \mnistofficehomestdAC{0.5} & \mnistofficehomestdAP{0.6} & \mnistofficehomestdAR{1.0} & \mnistofficehomestdCA{1.3} & \mnistofficehomestdCP{1.1} & \mnistofficehomestdCR{1.1} & \mnistofficehomestdPA{2.2} & \mnistofficehomestdPC{1.1} & \mnistofficehomestdPR{0.5} & \mnistofficehomestdRA{1.0} & \mnistofficehomestdRC{1.0} & \mnistofficehomestdRP{0.6} & \mnistofficehomestdAvg{1.0} \\
IM & \mnistofficehomestdMM{0.2} & \mnistofficehomestdAC{0.9} & \mnistofficehomestdAP{0.6} & \mnistofficehomestdAR{0.8} & \mnistofficehomestdCA{0.8} & \mnistofficehomestdCP{0.8} & \mnistofficehomestdCR{0.7} & \mnistofficehomestdPA{0.5} & \mnistofficehomestdPC{0.7} & \mnistofficehomestdPR{0.4} & \mnistofficehomestdRA{0.7} & \mnistofficehomestdRC{0.5} & \mnistofficehomestdRP{0.7} & \textbf{\mnistofficehomestdAvg{0.7}} \\
IM-DANN & - & \mnistofficehomestdAC{0.7} & \mnistofficehomestdAP{1.1} & \mnistofficehomestdAR{0.8} & \mnistofficehomestdCA{1.2} & \mnistofficehomestdCP{1.5} & \textbf{\mnistofficehomestdCR{0.4}} & \mnistofficehomestdPA{2.2} & \mnistofficehomestdPC{0.8} & \mnistofficehomestdPR{1.1} & \mnistofficehomestdRA{0.9} & \mnistofficehomestdRC{0.9} & \mnistofficehomestdRP{1.2} & \mnistofficehomestdAvg{1.0} \\
ITL & \mnistofficehomestdMM{0.2} & \mnistofficehomestdAC{0.4} & \mnistofficehomestdAP{0.7} & \mnistofficehomestdAR{0.9} & \mnistofficehomestdCA{1.1} & \mnistofficehomestdCP{0.4} & \textbf{\mnistofficehomestdCR{0.4}} & \mnistofficehomestdPA{1.3} & \mnistofficehomestdPC{1.2} & \mnistofficehomestdPR{0.6} & \mnistofficehomestdRA{0.9} & \mnistofficehomestdRC{1.2} & \mnistofficehomestdRP{1.1} & \mnistofficehomestdAvg{0.8} \\
JMMD & \mnistofficehomestdMM{8.1} & \textbf{\mnistofficehomestdAC{0.3}} & \mnistofficehomestdAP{0.5} & \mnistofficehomestdAR{0.5} & \mnistofficehomestdCA{0.9} & \textbf{\mnistofficehomestdCP{0.2}} & \mnistofficehomestdCR{1.3} & \mnistofficehomestdPA{1.4} & \mnistofficehomestdPC{1.0} & \mnistofficehomestdPR{0.4} & \textbf{\mnistofficehomestdRA{0.3}} & \mnistofficehomestdRC{0.8} & \mnistofficehomestdRP{0.9} & \textbf{\mnistofficehomestdAvg{0.7}} \\
MCC & \mnistofficehomestdMM{4.0} & \mnistofficehomestdAC{1.9} & \mnistofficehomestdAP{0.4} & \mnistofficehomestdAR{0.4} & \mnistofficehomestdCA{1.4} & \mnistofficehomestdCP{0.9} & \mnistofficehomestdCR{0.8} & \mnistofficehomestdPA{0.8} & \mnistofficehomestdPC{1.0} & \mnistofficehomestdPR{0.4} & \mnistofficehomestdRA{1.1} & \mnistofficehomestdRC{1.4} & \mnistofficehomestdRP{1.1} & \mnistofficehomestdAvg{1.0} \\
MCC-DANN & - & \mnistofficehomestdAC{1.1} & \mnistofficehomestdAP{1.2} & \mnistofficehomestdAR{0.9} & \mnistofficehomestdCA{0.9} & \mnistofficehomestdCP{0.3} & \mnistofficehomestdCR{0.8} & \mnistofficehomestdPA{0.9} & \mnistofficehomestdPC{1.3} & \mnistofficehomestdPR{0.7} & \mnistofficehomestdRA{0.4} & \mnistofficehomestdRC{1.1} & \mnistofficehomestdRP{0.6} & \mnistofficehomestdAvg{0.8} \\
MCD & \mnistofficehomestdMM{0.3} & \mnistofficehomestdAC{0.5} & \mnistofficehomestdAP{0.7} & \mnistofficehomestdAR{0.8} & \mnistofficehomestdCA{1.3} & \mnistofficehomestdCP{1.6} & \mnistofficehomestdCR{1.4} & \mnistofficehomestdPA{0.7} & \mnistofficehomestdPC{2.0} & \mnistofficehomestdPR{0.6} & \mnistofficehomestdRA{1.1} & \mnistofficehomestdRC{0.8} & \mnistofficehomestdRP{1.3} & \mnistofficehomestdAvg{1.1} \\
MMD & \mnistofficehomestdMM{0.4} & \mnistofficehomestdAC{0.9} & \mnistofficehomestdAP{0.4} & \mnistofficehomestdAR{0.5} & \mnistofficehomestdCA{0.9} & \mnistofficehomestdCP{1.2} & \mnistofficehomestdCR{0.9} & \mnistofficehomestdPA{0.7} & \mnistofficehomestdPC{1.3} & \mnistofficehomestdPR{0.8} & \mnistofficehomestdRA{0.5} & \textbf{\mnistofficehomestdRC{0.4}} & \mnistofficehomestdRP{1.4} & \mnistofficehomestdAvg{0.8} \\
MinEnt & \textbf{\mnistofficehomestdMM{0.1}} & \mnistofficehomestdAC{0.6} & \mnistofficehomestdAP{1.2} & \textbf{\mnistofficehomestdAR{0.3}} & \mnistofficehomestdCA{0.7} & \mnistofficehomestdCP{0.8} & \textbf{\mnistofficehomestdCR{0.4}} & \mnistofficehomestdPA{1.2} & \mnistofficehomestdPC{0.6} & \mnistofficehomestdPR{0.5} & \mnistofficehomestdRA{1.3} & \mnistofficehomestdRC{0.8} & \mnistofficehomestdRP{1.1} & \mnistofficehomestdAvg{0.8} \\
RTN & \mnistofficehomestdMM{0.7} & \mnistofficehomestdAC{0.4} & \mnistofficehomestdAP{0.9} & \mnistofficehomestdAR{0.6} & \mnistofficehomestdCA{0.9} & \mnistofficehomestdCP{0.6} & \mnistofficehomestdCR{1.0} & \mnistofficehomestdPA{1.4} & \mnistofficehomestdPC{1.2} & \mnistofficehomestdPR{0.7} & \mnistofficehomestdRA{0.6} & \mnistofficehomestdRC{1.3} & \mnistofficehomestdRP{0.9} & \mnistofficehomestdAvg{0.9} \\
STAR & \mnistofficehomestdMM{0.4} & \mnistofficehomestdAC{0.7} & \mnistofficehomestdAP{0.3} & \mnistofficehomestdAR{0.5} & \mnistofficehomestdCA{0.6} & \mnistofficehomestdCP{2.3} & \mnistofficehomestdCR{1.0} & \mnistofficehomestdPA{1.3} & \mnistofficehomestdPC{0.8} & \mnistofficehomestdPR{0.6} & \mnistofficehomestdRA{0.9} & \mnistofficehomestdRC{0.5} & \mnistofficehomestdRP{0.9} & \mnistofficehomestdAvg{0.9} \\
SWD & \mnistofficehomestdMM{1.9} & \mnistofficehomestdAC{1.1} & \mnistofficehomestdAP{0.5} & \mnistofficehomestdAR{0.9} & \mnistofficehomestdCA{1.1} & \mnistofficehomestdCP{0.3} & \mnistofficehomestdCR{0.8} & \mnistofficehomestdPA{0.9} & \mnistofficehomestdPC{1.0} & \mnistofficehomestdPR{0.6} & \mnistofficehomestdRA{1.4} & \mnistofficehomestdRC{1.7} & \mnistofficehomestdRP{0.8} & \mnistofficehomestdAvg{0.9} \\
SymNets & \mnistofficehomestdMM{24.7} & \mnistofficehomestdAC{1.4} & \mnistofficehomestdAP{1.1} & \mnistofficehomestdAR{1.1} & \mnistofficehomestdCA{1.2} & \mnistofficehomestdCP{3.7} & \mnistofficehomestdCR{1.2} & \mnistofficehomestdPA{0.7} & \mnistofficehomestdPC{2.1} & \mnistofficehomestdPR{0.9} & \mnistofficehomestdRA{0.5} & \mnistofficehomestdRC{1.1} & \mnistofficehomestdRP{1.5} & \mnistofficehomestdAvg{1.4} \\
VADA & \mnistofficehomestdMM{1.5} & \mnistofficehomestdAC{0.8} & \mnistofficehomestdAP{0.3} & \textbf{\mnistofficehomestdAR{0.3}} & \mnistofficehomestdCA{1.6} & \mnistofficehomestdCP{1.5} & \mnistofficehomestdCR{1.5} & \mnistofficehomestdPA{1.0} & \mnistofficehomestdPC{1.2} & \mnistofficehomestdPR{0.7} & \mnistofficehomestdRA{1.6} & \mnistofficehomestdRC{0.7} & \mnistofficehomestdRP{0.4} & \mnistofficehomestdAvg{1.0} \\
\bottomrule
\end{tabular}
\caption{Standard deviation on the MNIST $\rightarrow$ MNISTM (MM) and OfficeHome transfer tasks. The Avg column is the OfficeHome average.}
\label{supp_mnist_officehome_oracle_results_std}
\end{table*}

\begin{figure}
    \centering
    \includegraphics[width=\columnwidth]{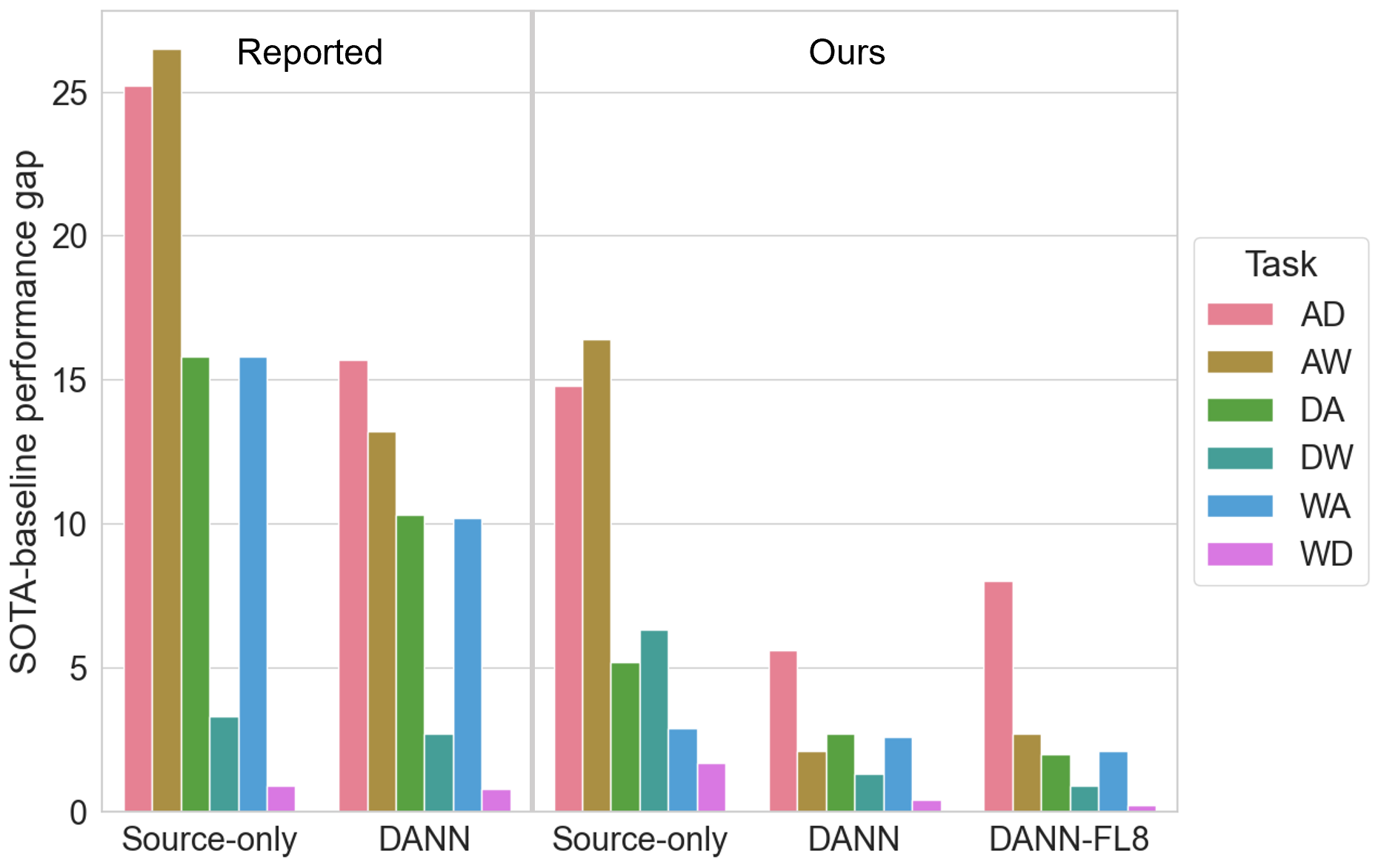}
    \caption{Performance gaps between SOTA and baseline algorithms (source-only and DANN) on Office31 tasks. The reported numbers are the average from 2021 papers.}
    \label{supp_reported_vs_our_diffs_office31}
\end{figure}
\begin{table*}
\centering
\begin{tabular}{lrrrrrrrrrrr}
\toprule
{Algorithm} & {MM} & {AD} & {AW} & {DA} & {DW} & {WA} & {WD} & {AP} & {CR} & {PA} & {RC} \\
\midrule
ADDA & 7.2 & 2.6 & 0.6 & 39.8 & 35.7 & 33.6 & 0.7 & 21.4 & 0.6 & 41.7 & - \\
AFN & 10.9 & 5.3 & 3.1 & 6.6 & 1.6 & 1.8 & 0.5 & 1.9 & 1.6 & 1.3 & 2.9 \\
ATDOC & 55.1 & 11.4 & 7.0 & 1.7 & 2.6 & 1.9 & 7.8 & - & 1.3 & 5.3 & - \\
BNM & 1.1 & 8.7 & 11.5 & 3.9 & 1.1 & 3.9 & 1.2 & 2.3 & 6.8 & 9.8 & 0.9 \\
BSP & 0.0 & 1.4 & 0.3 & 0.2 & 1.0 & 0.1 & 0.0 & 0.8 & 3.8 & 2.1 & 3.7 \\
CDAN & 22.5 & 13.5 & 7.1 & 9.5 & 10.2 & 2.5 & 5.7 & 0.9 & 6.7 & 9.5 & - \\
CORAL & 12.4 & 6.5 & 9.9 & 19.1 & 5.7 & 14.3 & 0.9 & 0.7 & 9.5 & 6.0 & 1.3 \\
DANN & 25.0 & 17.0 & 8.5 & 10.5 & 4.3 & 6.3 & 14.1 & 3.3 & 2.5 & 7.0 & 4.0 \\
DC & 3.7 & 10.4 & 6.6 & 6.9 & 5.7 & 4.5 & 5.7 & 3.3 & 0.7 & 7.0 & 0.6 \\
GVB & 17.6 & 3.0 & 8.3 & 16.9 & 9.6 & 12.9 & 9.5 & - & 5.6 & 13.5 & 1.8 \\
JMMD & 1.4 & 27.5 & 20.0 & 19.8 & 1.7 & 17.8 & 0.5 & 1.5 & 1.6 & 5.4 & 4.4 \\
MCC & 2.1 & 4.6 & 6.6 & 9.6 & 1.0 & 9.2 & 5.9 & 10.0 & 5.3 & 10.6 & 5.0 \\
MCD & 5.6 & 3.2 & 7.6 & 10.5 & 1.9 & 15.7 & 0.5 & - & 0.6 & 3.8 & - \\
MMD & 4.9 & 12.8 & 14.3 & 21.9 & 2.1 & 16.3 & 0.5 & 2.6 & 5.5 & 13.2 & 3.5 \\
RTN & 2.2 & 0.4 & 0.9 & 6.0 & 5.2 & 2.9 & 0.5 & 2.0 & 1.5 & 5.7 & 1.5 \\
SWD & 15.7 & 10.8 & 7.0 & 29.2 & 0.4 & 24.2 & 1.1 & 0.0 & 2.0 & 9.9 & - \\
SymNets & 0.0 & 0.9 & 20.2 & 10.3 & 4.5 & 9.6 & 3.6 & - & - & 16.2 & 25.8 \\
VADA & 9.7 & 3.7 & 5.0 & 0.5 & 1.6 & 8.0 & 0.9 & - & 7.0 & 2.8 & - \\
\bottomrule
\end{tabular}
\caption{Performance gap between oracle and IM, at 0.98 source thresholding.}
\label{supp_per_algorithm_diff_IM}
\end{table*}

\begin{table*}
\centering
\begin{tabular}{lrrrrrrrrrrr}
\toprule
{Algorithm} & {MM} & {AD} & {AW} & {DA} & {DW} & {WA} & {WD} & {AP} & {CR} & {PA} & {RC} \\
\midrule
ADDA & 0.3 & 80.2 & - & - & - & 2.4 & 94.9 & 1.1 & - & 2.4 & 1.7 \\
AFN & 44.5 & 4.5 & 0.6 & 7.5 & 6.4 & 27.1 & 0.6 & 1.4 & 1.6 & 7.0 & 3.9 \\
ATDOC & 52.7 & 16.8 & 9.2 & 6.2 & 6.8 & 5.4 & 10.9 & - & 0.4 & 2.9 & - \\
BNM & 43.0 & 7.3 & 2.5 & 0.9 & 4.4 & 3.4 & 0.8 & 3.1 & 5.3 & 1.3 & 8.5 \\
BSP & 5.8 & 1.3 & 3.8 & 18.6 & 12.4 & 35.2 & 2.3 & - & - & 1.4 & 0.8 \\
CDAN & 8.9 & 7.0 & - & - & - & 1.7 & 1.9 & 1.3 & - & 1.3 & 4.5 \\
CORAL & 24.5 & 8.8 & 9.2 & 25.6 & 7.4 & 25.1 & 8.3 & 3.1 & 11.1 & 6.5 & 2.6 \\
DANN & 20.0 & 16.5 & 6.0 & 0.9 & 2.9 & 2.0 & 3.4 & 0.6 & 3.0 & 3.0 & 2.2 \\
DC & 2.7 & 7.0 & 8.4 & 0.4 & 0.8 & 2.7 & 2.4 & 1.3 & - & 2.7 & 1.5 \\
GVB & 42.3 & 8.5 & 19.9 & 33.1 & 9.9 & 32.4 & 1.8 & - & 3.4 & 4.9 & 4.2 \\
JMMD & 37.2 & 5.9 & 1.7 & 4.1 & 1.7 & 16.7 & 5.4 & - & - & 3.8 & 1.4 \\
MCC & 1.6 & 3.6 & 3.8 & 4.3 & 2.6 & 0.0 & 2.0 & 5.6 & 5.2 & 6.1 & 1.7 \\
MCD & 22.6 & - & 1.1 & 26.9 & 1.1 & 5.0 & 0.0 & - & 8.1 & 2.4 & 1.6 \\
MMD & 20.4 & 8.1 & 3.7 & 8.7 & 5.4 & 42.2 & 1.7 & 2.2 & 8.6 & 3.4 & 1.2 \\
RTN & 22.3 & 5.2 & 4.2 & 1.4 & 5.2 & 67.0 & 2.2 & 1.4 & 8.0 & 2.2 & 2.4 \\
SWD & 41.0 & 7.9 & 2.9 & 0.6 & 1.6 & 29.7 & 1.3 & - & - & 0.9 & - \\
SymNets & 14.6 & 4.6 & 9.0 & 5.3 & 1.4 & 1.5 & 5.9 & - & - & 28.2 & - \\
VADA & 0.6 & 6.7 & 3.1 & - & 1.1 & 13.0 & 0.2 & 4.3 & 4.8 & 3.4 & - \\
\bottomrule
\end{tabular}
\caption{Performance gap between oracle and DEV, at 0.98 source thresholding.}
\label{supp_per_algorithm_diff_DEV}
\end{table*}
\begin{table*}
\centering
\begin{tabular}{lrrrrrrrrrrr}
\toprule
{Algorithm} & {MM} & {AD} & {AW} & {DA} & {DW} & {WA} & {WD} & {AP} & {CR} & {PA} & {RC} \\
\midrule
ADDA & 73.9 & 80.4 & 75.2 & 61.0 & 90.3 & 63.0 & 94.8 & 61.7 & 61.7 & 56.7 & 41.5 \\
AFN & 53.4 & 5.1 & 9.1 & 50.6 & 7.1 & 43.9 & 8.2 & 1.7 & 1.5 & 6.6 & 0.0 \\
ATDOC & 60.8 & 23.9 & 8.6 & 6.5 & 16.0 & 52.4 & 13.2 & 6.0 & 14.6 & 20.6 & - \\
BNM & 53.9 & 2.0 & 5.1 & 37.3 & 5.1 & 54.2 & 3.5 & 1.4 & 1.8 & 3.4 & 1.9 \\
BSP & 48.7 & 0.6 & 15.5 & 30.4 & 16.5 & 33.1 & 7.3 & - & 23.7 & 11.5 & 1.1 \\
CDAN & 47.5 & 8.3 & 4.3 & 10.4 & 8.8 & 8.0 & 3.1 & 2.7 & 15.4 & 7.3 & 0.5 \\
CORAL & 16.2 & 15.0 & 10.1 & 24.2 & 11.9 & 54.0 & 2.9 & 2.2 & 4.1 & 7.6 & 3.4 \\
DANN & 43.2 & 12.3 & 6.0 & 6.0 & 5.7 & 36.5 & 2.0 & 0.4 & 2.8 & 6.9 & 4.8 \\
DC & 47.1 & 8.9 & 4.4 & 13.2 & 7.8 & 27.7 & 0.8 & 1.2 & 4.9 & 14.8 & 1.9 \\
GVB & 60.8 & 34.6 & 19.6 & 9.8 & 6.0 & 34.2 & 16.4 & 2.0 & 12.2 & 19.6 & 2.8 \\
JMMD & 27.0 & 12.1 & 11.2 & 41.5 & 17.1 & 59.0 & 4.6 & 2.3 & 0.4 & 22.3 & 2.0 \\
MCC & 45.2 & 0.0 & 1.5 & 4.5 & 11.1 & 49.2 & 9.0 & 0.9 & 0.9 & 0.8 & 4.5 \\
MCD & 83.3 & 2.7 & 2.6 & 41.7 & 15.1 & 36.4 & 4.9 & - & 8.3 & 15.2 & 1.1 \\
MMD & 28.6 & 15.0 & 19.2 & 36.3 & 19.5 & 51.5 & 4.8 & - & 21.3 & 8.0 & 2.2 \\
RTN & 47.8 & 9.1 & 20.3 & 68.8 & 94.8 & 69.6 & 8.2 & - & 12.1 & 22.8 & 1.1 \\
SWD & 43.3 & 10.1 & 7.7 & 37.2 & 13.8 & 28.3 & 9.9 & 2.4 & 12.0 & 12.2 & - \\
SymNets & 77.4 & 13.1 & 80.3 & 13.2 & 89.2 & 54.0 & 84.5 & - & - & 50.8 & - \\
VADA & 79.0 & 2.3 & 5.5 & 1.5 & - & 7.3 & 0.0 & - & 2.4 & 9.2 & - \\
\bottomrule
\end{tabular}
\caption{Performance gap between oracle and SND, at 0.98 source thresholding.}
\label{supp_per_algorithm_diff_SND}
\end{table*}
\begin{table*}
\centering
\begin{tabular}{lrrrrrrrrrrr}
\toprule
{Algorithm} & {MM} & {AD} & {AW} & {DA} & {DW} & {WA} & {WD} & {AP} & {CR} & {PA} & {RC} \\
\midrule
ADDA & 10.6 & 4.8 & 6.7 & 11.5 & 12.3 & 16.4 & 20.3 & 9.9 & 4.6 & 7.0 & 1.9 \\
AFN & 3.4 & 3.5 & 3.2 & 1.3 & 8.7 & 44.6 & 4.8 & 1.9 & 5.2 & 3.1 & 2.5 \\
ATDOC & 10.9 & 4.5 & 4.9 & 0.4 & 5.1 & 2.1 & 4.0 & 5.4 & 0.0 & 8.2 & - \\
BNM & 4.0 & 4.7 & 8.9 & 3.5 & 7.9 & 4.3 & 8.0 & 1.5 & 11.5 & 6.9 & 5.5 \\
BSP & 0.2 & 0.8 & 1.7 & 0.6 & 9.6 & 40.8 & 8.3 & - & 4.2 & 0.9 & 1.1 \\
CDAN & 28.2 & 7.7 & 8.8 & 2.2 & 2.1 & 3.0 & 1.9 & 0.3 & 8.5 & 5.4 & 4.4 \\
CORAL & 16.2 & 7.2 & 6.1 & 1.0 & 9.8 & 0.5 & 5.1 & 0.5 & 2.2 & 1.5 & 0.9 \\
DANN & 29.2 & 9.6 & 3.1 & 1.6 & 7.0 & 4.3 & 3.9 & 2.0 & 8.0 & 6.8 & 7.0 \\
DC & 28.6 & 2.8 & 7.3 & 3.2 & 2.0 & 10.5 & 7.2 & 0.8 & 5.7 & 4.9 & 4.9 \\
GVB & 11.8 & 5.2 & 13.9 & 4.2 & 9.2 & 6.1 & 9.1 & 2.0 & 17.2 & 7.5 & 4.4 \\
JMMD & 25.7 & 5.4 & 6.8 & 3.0 & 17.4 & 2.0 & 8.8 & 2.7 & 2.3 & 5.3 & 2.6 \\
MCC & 5.3 & 6.9 & 12.6 & 3.3 & 12.0 & 5.2 & 9.6 & 5.4 & 7.3 & 7.1 & 4.5 \\
MCD & 29.1 & 3.8 & 3.1 & 0.9 & 3.3 & 0.9 & 1.4 & - & 4.2 & 3.6 & 1.1 \\
MMD & 7.9 & 3.5 & 6.3 & 1.8 & 7.8 & 39.5 & 6.7 & - & 2.6 & 4.6 & 6.2 \\
RTN & 0.0 & 2.9 & 9.2 & 16.6 & 5.0 & 2.6 & 4.0 & - & 5.6 & 5.4 & 0.1 \\
SWD & 24.7 & 2.4 & 9.3 & 0.0 & 3.6 & 9.2 & 1.9 & 2.7 & 1.4 & 3.5 & - \\
SYMNETS & 34.8 & 1.9 & 14.4 & 9.5 & 4.9 & 8.6 & 4.6 & - & - & 38.7 & - \\
VADA & 32.1 & 5.7 & 6.4 & 1.0 & - & 2.2 & 0.9 & - & 2.4 & 5.3 & - \\
\bottomrule
\end{tabular}
\caption{Performance gap between oracle and NegSND, at 0.98 source thresholding.}
\label{supp_per_algorithm_diff_NegSND}
\end{table*}

\end{document}